\documentclass{article}

\PassOptionsToPackage{numbers, compress}{natbib}

\usepackage[main, final]{neurips_2026}
\usepackage[utf8]{inputenc}
\usepackage[T1]{fontenc}
\usepackage{hyperref}
\usepackage{url}
\usepackage{booktabs}
\usepackage{amsfonts}
\usepackage{nicefrac}
\usepackage{microtype}

\usepackage{amsmath}
\usepackage{graphicx}
\usepackage{wrapfig}

\usepackage{multirow}
\usepackage[table]{xcolor}

\definecolor{headerbg}{RGB}{245, 245, 245}
\definecolor{classicalbg}{RGB}{245, 240, 235}
\definecolor{singlebg}{RGB}{245, 250, 245} 
\definecolor{multibg}{RGB}{250, 248, 240}  
\definecolor{pomobg}{RGB}{252, 245, 245} 
\definecolor{ourbg}{RGB}{225, 240, 255}  

\definecolor{green1}{RGB}{225, 240, 220} 
\definecolor{green2}{RGB}{195, 225, 185}   
\definecolor{green3}{RGB}{155, 205, 145} 

\definecolor{yellow1}{RGB}{252, 245, 210} 
\definecolor{yellow2}{RGB}{248, 230, 160}  
\definecolor{yellow3}{RGB}{240, 210, 110} 

\definecolor{blue1}{RGB}{225, 238, 250}
\definecolor{blue2}{RGB}{195, 220, 245}
\definecolor{blue3}{RGB}{155, 195, 235}

\definecolor{winNoNorm}{RGB}{225, 238, 250}
\definecolor{winBN}{RGB}{252, 235, 215}

\usepackage{caption}
\usepackage{tabularx}
\newcolumntype{C}{>{\centering\arraybackslash}X}

\title{MixRoute: Rethinking Single-Distribution Training for Generalizable Neural Routing}

\author{
Hang Yi$^{1}$, \hfil
Ziwei Huang$^{1}$, \hfil
Yining Ma$^{2}$, and \hfil
Zhiguang Cao$^{1}$ \\
$^{1}$Singapore Management University, $^{2}$Massachusetts Institute of Technology \\
\texttt{hang.yi.2024@phdcs.smu.edu.sg, ziweihuang@smu.edu.sg} \\
\texttt{yiningma@mit.edu, zgcao@smu.edu.sg}
}

\begin{document}

\maketitle

\begin{abstract}
Neural Combinatorial Optimization (NCO) is a promising paradigm for solving routing problems, yet its generalization across diverse data distributions remains a critical bottleneck. Existing methods typically tackle cross-distribution generalization through complex multi-distribution training or intricate meta-learning schemes, implicitly presuming that single-distribution training yields a weak zero-shot baseline. In this paper, we demonstrate that the potential of single-distribution training has been substantially underestimated and can be unlocked by a simple architectural inductive bias: Mix Normalization. To approach this, we first construct a comprehensive benchmark spanning three representative routing problems and 179 fine-grained datasets, capturing diverse shifts in node coordinates, customer demands, capacity and time windows. Building on this benchmark, we propose MixRoute, a neural framework that adaptively combines normalization statistics at different granularities to mitigate varying distribution shifts. Trained exclusively on uniform distributions, MixRoute generalizes effectively to a wide range of unseen distributions without any distribution-specific adaptation. Extensive experiments demonstrate that MixRoute consistently achieves state-of-the-art zero-shot generalization. Our work revisits normalization as a parsimonious alternative for zero-shot generalization against complex training-adaptation schemes. Code is available at \url{https://github.com/yihang0410/MixRoute}.

\end{abstract}

\section{Introduction}

The Vehicle Routing Problem (VRP) is a fundamental class of combinatorial optimization problems that seek cost-minimizing routes for serving geographically distributed customers under various operational constraints~\citep{vrp}. As a cornerstone of modern logistics, VRPs underpin a wide range of real-world applications, including transportation~\citep{transportation}, supply chain management~\citep{supplychain}, and ride-sharing~\citep{ridesharing}. Owing to their NP-hard nature~\citep{nphard}, obtaining exact solutions becomes computationally prohibitive as problem size grows. Classical heuristics, such as LKH3~\citep{lkh} and HGS~\citep{hgs}, can produce high-quality solutions but rely heavily on hand-crafted rules and domain expertise, making them difficult to adapt to new problem variants or operational constraints. To overcome these limitations, Neural Combinatorial Optimization (NCO) has emerged as a promising data-driven alternative~\citep{nco1, nco2}, leveraging deep neural networks to automatically learn solution policies from instance data. Among existing NCO paradigms, constructive methods~\citep{am,pomo,pointernetwork,radar} which sequentially build a solution by appending one node at a time have become the dominant approach due to their ability to learn solution policies end-to-end without relying on hand-crafted local-search operators.

Despite their empirical success, most NCO methods are trained and predominantly evaluated on instances synthesized from a single canonical distribution, typically with node coordinates sampled uniformly over the unit square \citep{pomo,am,pointernetwork,nco1,reld}. Even works targeting size generalization have largely retained this uniform-distribution setup on both training and testing sides \citep{lehd,bqnco,generalize,zhou2024instance,son2023meta}. However, empirical evidence shows that once the node coordinates depart from this idealized regime, the performance of these solvers degrades sharply \citep{bi2022learning,joshi2022learning,geisler2022generalization}. This indicates that what such models have learned is, to a large extent, statistical regularities specific to the training distribution rather than the problem-invariant combinatorial structure of routing itself. To address this cross-distribution generalization problem, a growing line of work has resorted to increasingly complex remedies, such as data augmentation across multiple distributions \citep{jiang2022learning}, meta-learning over diverse task families \citep{zhou2023towards, manchanda2022generalization, qiu2022dimes}, multi-teacher distillation \citep{bi2022learning}, and distribution-specific architectural components \citep{xiao2025improving}. A common premise in these approaches is that single-distribution training is fundamentally insufficient.

In this work, we revisit this premise and ask a question that has received little attention: \emph{what is the actual ceiling of single-distribution training?} Rather than further scaling up training data diversity, we look for inspiration outside of NCO. In computer vision, where cross-distribution generalization has been studied extensively, normalization layers have repeatedly emerged as a critical lever for explicitly controlling what statistics a model preserves or discards across distributions \citep{luo2019differentiable, nam2018batch, pan2018ibn}. This motivates us to ask whether normalization could serve as an equally important entry point for cross-distribution generalization in NCO. Our preliminary experiments confirm that different normalization schemes indeed exhibit different behaviors across routing distributions, with no single scheme being uniformly optimal across the diverse distributions. Motivated by this observation, we propose \textbf{MixRoute}, a neural routing framework that adaptively harnesses the complementary strengths of multiple normalizations through a Mix Normalization (MixNorm) module and demonstrates that with the right normalization design, single-distribution training alone can already achieve strong zero-shot cross-distribution generalization. Our contributions are summarized as follows:
\begin{itemize}
\item We introduce MixNorm to NCO for the first time and identify it as a simple yet effective lever for cross-distribution generalization, showing that the ceiling of single-distribution training is substantially higher than previously assumed.
\item We construct a comprehensive benchmark of 179 distribution shifts spanning diverse spatial and demand patterns, providing a unified testbed for systematically evaluating cross-distribution generalization in neural routing solvers.
\item We extensively validate MixRoute for different routing problems, backbone architectures, and normalization choices, demonstrating its consistent effectiveness.
\end{itemize}

\section{Related Work}

\textbf{Neural Solvers for VRPs.} NCO leverages deep neural networks to learn solution policies directly from problem instances, and existing neural solvers for VRPs broadly fall into two paradigms: 
\emph{construction} methods, which generate a complete solution from scratch, 
and \emph{improvement} methods, which iteratively refine an initial solution 
through learned local-search operators~\citep{wu2021learning, ma2021learning, hottung2020neural}. 
Within the construction paradigm, autoregressive (AR) approaches have become 
the dominant choice: starting from \citet{pointernetwork} and \citet{nco1}, 
\citet{am} introduced the Attention Model (AM), which was further advanced by 
POMO~\citep{pomo} and remains a canonical backbone for neural routing. 
Subsequent works have extended this line through heavier 
decoders~\citep{lehd}, bisimulation-based reformulations~\citep{bqnco}, 
instance-conditioned adaptation~\citep{zhou2024instance},
mixture-of-experts architectures~\citep{mvmoe}, and 
vision-assisted multi-modal encoders~\citep{gui2026vision}. 
Non-autoregressive (NAR) variants generate all edges in parallel via graph 
neural networks for faster inference~\citep{joshi2022learning}, but typically 
trail AR methods in solution quality. 
Complementary to learned policies, recent work also employs large language models to automatically generate solvers for complex VRP variants~\citep{eoh, zhang2026agentic}. 
We adopt the AR construction paradigm as our backbone, following the 
dominant practice in recent neural routing literature.

\textbf{Cross-Distribution Generalization in Neural VRP Solvers.} A growing body of work has investigated cross-distribution generalization 
in neural VRP solvers. \emph{Data-centric} approaches enrich the training 
distribution: \citet{jiang2022learning} formulate routing under a 
distributionally robust optimization objective, 
\citet{geisler2022generalization} introduce adversarial perturbations to 
expose model vulnerabilities, and \citet{wang2024asp} adopt curriculum-style 
instance generation guided by policy-space response oracles. 
\emph{Model-centric} approaches inject distributional priors into the 
solver itself through meta-learning over diverse task 
families~\citep{zhou2023towards, manchanda2022generalization}, 
multi-teacher knowledge distillation that consolidates expertise across 
distributions~\citep{bi2022learning}, or distribution-specific architectural 
components such as multiple decoders~\citep{xiao2025improving}. 
A common thread across these methods is a shared assumption that single-distribution training is inherently insufficient for cross-distribution generalization, which has driven the field toward increasingly complex solutions. Our work challenges this assumption from a different angle: we show that a single uniform training distribution can already unlock strong cross-distribution generalization ability.

\textbf{Mix Normalization.} Different normalization schemes, including Batch Normalization 
(BN)~\citep{ioffe2015batch}, Layer Normalization (LN)~\citep{ba2016layer}, 
and Instance Normalization (IN)~\citep{ulyanov2016instance}, differ in the 
granularity at which statistics are aggregated and exhibit complementary 
strengths across architectures and tasks. 
Switchable Normalization (SN)~\citep{luo2019differentiable} introduces the idea of learnable normalizer mixtures, motivated by the empirical observation that no single normalizer is universally optimal across layers or tasks. Subsequent works extend this idea along several directions, including sparse selection among candidates~\citep{shao2019ssn}, gated combinations restricted to specific normalizer pairs~\citep{nam2018batch, pan2018ibn}, and domain-aware mixtures designed to mitigate distributional shift~\citep{seo2020learning, fan2021adversarially}. However, these developments have remained almost exclusively within image-based settings. To the best of our knowledge, our work is the first to investigate mixture-based normalization in the context of NCO.
\section{Benchmark Datasets across Diverse Distributions}
\label{sec:benchmark}

\begin{figure}[t]
    \centering
    \includegraphics[width=0.9\linewidth]{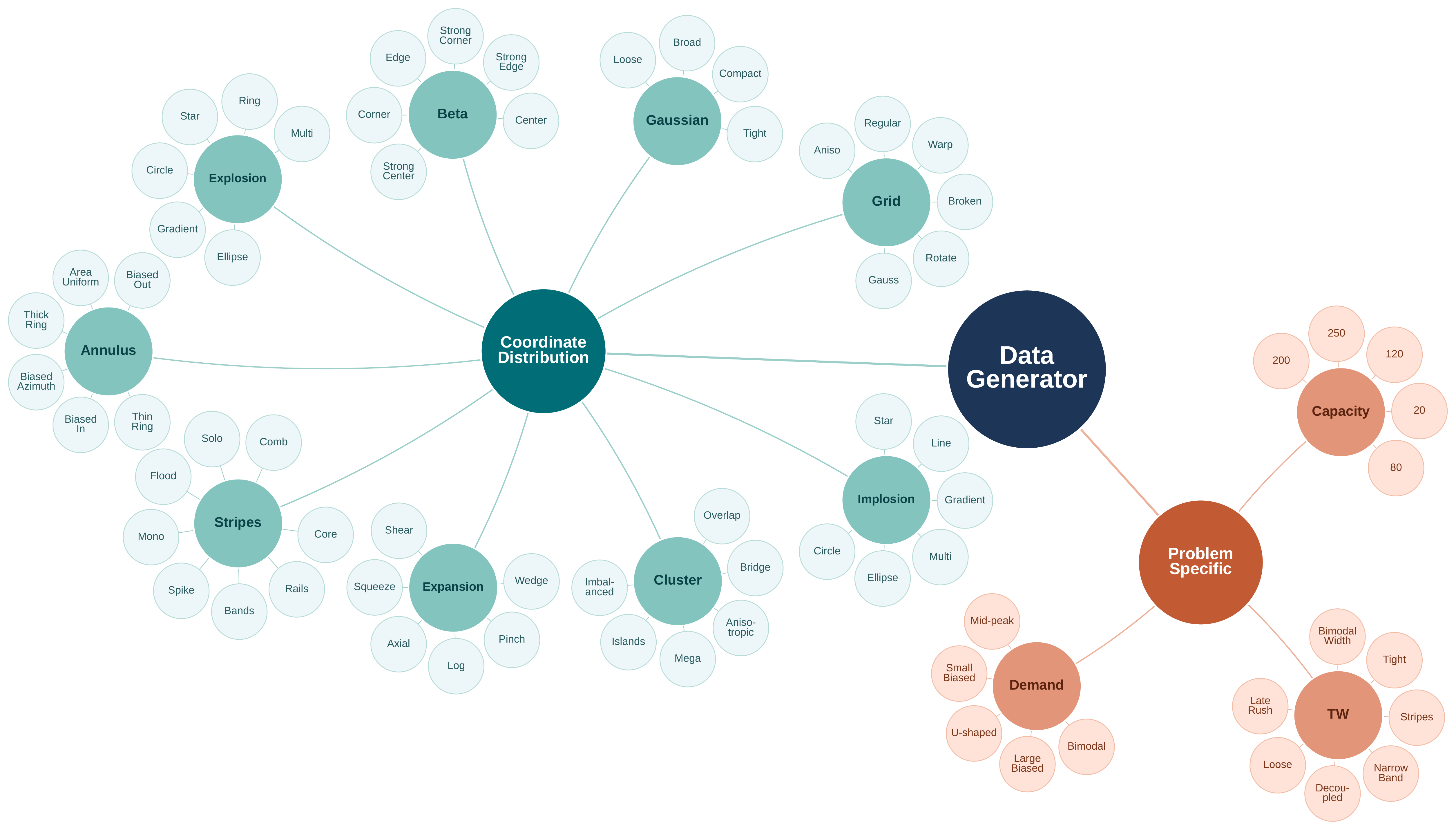}
    \caption{Hierarchical structure of our cross-distribution benchmark. Starting from a central data generator, the benchmark branches into two top-level axes: coordinate distributions and problem-specific shifts. Each family further expands into multiple parametric variants.}
    \label{fig:benchmark}
\end{figure}

To rigorously evaluate cross-distribution generalization, we construct a 
test suite that systematically varies the four structural attributes of 
a routing instance most likely to differ between training and deployment: node coordinates, customer demands, capacity and time window. All benchmark instances are used exclusively at test 
time; training is conducted on a single uniform distribution. Our full benchmark constitutes a structured taxonomy of distribution shifts, as visualized in Figure~\ref{fig:benchmark}.

\textbf{Coordinate distributions.}
Most prior neural VRP solvers train and evaluate exclusively on a single uniform distribution, while a handful of recent works extend evaluation to a small number of mutated patterns derived from \citet{bossek2019evolving}, commonly \textit{Explosion}, \textit{Implosion}, \textit{Expansion}, and \textit{Grid}, which serve as the de facto standard for cross-distribution evaluation~\citep{bi2022learning, zhou2023towards}. While these patterns introduce diversity, they remain a narrow slice of the broader instance space and are typically used in their canonical, fixed-parameter form only. We instead design nine structurally distinct coordinate families: five novel families, namely \textit{Annulus}, \textit{Beta}, \textit{Gaussian}, \textit{Stripe}, and \textit{Cluster}, together with the four traditional ones mentioned above. To stress-test generalization, we further generate parametric variants within each family by varying its control parameters, where each parameter setting yields a substantially different spatial pattern rather than a mere re-sampling of the same distribution.

\textbf{Demand distributions.}
Existing neural VRP solvers almost universally adopt a single uniform 
demand distribution~\citep{bi2022learning, zhou2023towards}, 
overlooking the substantial demand variability observed in real-world 
deployments. We design five demand families, all constrained to share the same numerical range $[1, D_{\max}]$ to isolate the effect of distributional \emph{shape} from that of \emph{scale}.

\textbf{Time window.} Prior neural works almost exclusively evaluate on a single uniform time-window distribution identical to the training distribution, which fails to reflect the highly structured temporal patterns that arise in real-world delivery. Service windows in production logistics are rarely independent uniform draws: they are shaped by morning and evening rush hours, fixed departure slots in scheduled fleets, and customer business hours that cluster into a few standard bands. A model trained only on uniformly distributed windows therefore implicitly relies on two strong priors: that every time of day is equally likely, and that the temporal pattern decouples from the spatial layout. To address this, we build a time-window distribution shift benchmark consisting of 7 variants that systematically perturb three orthogonal axes of the temporal structure: the width and tightness of the windows themselves, the temporal location and concentration of when the bulk of service must occur, and the spatial-temporal coupling between customer position and window timing.

\begin{wraptable}[8]{r}{0.35\linewidth}
    \vspace{-1.2\baselineskip}
    \centering
    \footnotesize
    \caption{Capacity regimes. Demands are sampled from $\{1,\ldots,9\}$ ($\bar{d}=5$).}
    \label{tab:capacity}
    \setlength{\tabcolsep}{3pt}
    \resizebox{\linewidth}{!}{%
    \begin{tabular}{lcccc}
        \toprule
        Regime & $Q$ & $d_{\min}/Q$ & $\bar{d}/Q$ & $d_{\max}/Q$ \\
        \midrule
        Tight  & 20  & 0.050 & 0.250 & 0.450 \\
        Medium & 50  & 0.020 & 0.100 & 0.180 \\
        Loose  & 250 & 0.004 & 0.020 & 0.036 \\
        \bottomrule
    \end{tabular}}
    \vspace{-0.8\baselineskip}
\end{wraptable}
\textbf{Capacity configurations.} The first three dimensions vary the \emph{distributional shape} of coordinates and demands; the fourth dimension instead perturbs the \emph{numerical regime} itself. The effective load on each route is jointly determined by demand and vehicle capacity. Because demands enter the model as the normalized ratio $d_i / Q$, scaling the capacity $Q$ is equivalent to rescaling all demands proportionally. Table~\ref{tab:capacity} provides three illustrative configurations, where varying $Q$ from $20$ to $250$ produces a more than tenfold change in the numerical range, despite identical demand sampling.

Table~\ref{tab:benchmark_comparison} compares our benchmark with representative cross-distribution benchmarks for neural routing solvers. Existing benchmarks fall into two largely disconnected categories: AMDKD~\citep{bi2022learning} and INViT~\citep{fang2024invit} vary only node coordinates on TSP/CVRP, while Luo et al.~\citep{hard-constraint} vary only constraint tightness on CVRP/CVRPTW. In contrast, our benchmark unifies both types of shift in a single, comprehensive testbed, spanning three problems (TSP, CVRP, CVRPTW), four shift dimensions (coordinates, demand, capacity, time window), and 179 fine-grained out-of-distribution test settings.

\begin{table}[!h]
\centering
\fontsize{6.5pt}{7.5pt}\selectfont
\caption{Comparison of cross-distribution generalization benchmarks for neural routing solvers. \# Train and \# Test denote the numbers of training and out-of-distribution test distributions reported in the original works.}
\label{tab:benchmark_comparison}
\vspace{6pt}
\renewcommand{\arraystretch}{1.0}
\setlength{\tabcolsep}{3pt}
\begin{tabularx}{\textwidth}{l|CC|CCC|C|CC|CCC|C}
\toprule
\rowcolor{headerbg}
\textbf{Benchmark} 
& \multicolumn{2}{c|}{\textbf{\# Distributions}}
& \multicolumn{3}{c|}{\textbf{Problem Coverage}} 
& \multicolumn{1}{c|}{\textbf{TSP}}
& \multicolumn{2}{c|}{\textbf{CVRP-specific}}
& \multicolumn{3}{c|}{\textbf{CVRPTW-specific}}
& \textbf{Open-} \\
\rowcolor{headerbg}
& \# Train & \# Test & TSP & CVRP & CVRPTW 
& Coord.\
& Demand & Capacity 
& Width & Timing & Coupling 
& \textbf{source} \\
\midrule
AMDKD~\citep{bi2022learning}            & 3        & 4 & \checkmark & \checkmark & $\times$   & \checkmark & $\times$   & $\times$   & $\times$   & $\times$ & $\times$ & \checkmark \\
INViT~\citep{fang2024invit}             & 1        & 4 & \checkmark & \checkmark & $\times$   & \checkmark & $\times$   & $\times$   & $\times$   & $\times$ & $\times$ & \checkmark \\
Luo et al.~\citep{hard-constraint}      & multiple & 7 & $\times$   & \checkmark & \checkmark & $\times$   & $\times$   & \checkmark & \checkmark & $\times$ & $\times$ & \checkmark \\
\midrule
\rowcolor{ourbg} \textbf{Ours}          & \textbf{1} & \textbf{179} & \checkmark & \checkmark & \checkmark & \checkmark & \checkmark & \checkmark & \checkmark & \checkmark & \checkmark & \checkmark \\
\bottomrule
\end{tabularx}
\end{table}
\section{Methodologies}

\label{sec:method}

\textbf{Notation.}
Let $\mathcal{X}$ denote the instance space of a routing problem and $\mathcal{D}$ denote a probability distribution over $\mathcal{X}$. A neural solver $\pi_\theta : \mathcal{X} \rightarrow \mathcal{Y}$ parameterized by $\theta$ maps an instance $x \in \mathcal{X}$ to a feasible solution. Training amounts to minimizing the expected cost $\mathcal{L}(\theta; \mathcal{D}) = \mathbb{E}_{x \sim \mathcal{D}}[\, c(\pi_\theta(x))\,]$, where $c(\cdot)$ denotes the problem-specific cost (e.g., tour length).

\textbf{Single-distribution training.}
Following the convention in early NCO works~\citep{am, pomo}, the model is trained on instances drawn from a single source distribution $\mathcal{D}_{\text{train}}$:
\begin{equation}
    \theta^* = \arg\min_\theta \; \mathbb{E}_{x \sim \mathcal{D}_{\text{train}}}\bigl[\, c(\pi_\theta(x)) \,\bigr],
    \label{eq:single-dist}
\end{equation}
where $\mathcal{D}_{\text{train}}$ is typically the Uniform distribution over the unit square $[0,1]^2$. The trained model is expected to generalize to a target distribution $\mathcal{D}_{\text{test}} \neq \mathcal{D}_{\text{train}}$ at deployment, without further adaptation.

\textbf{Multi-distribution training.}
To address the cross-distribution generalization issue, a growing line of work~\citep{jiang2022learning, bi2022learning, zhou2023towards} expands the training data to span $K \geq 2$ distinct source distributions $\{\mathcal{D}_k\}_{k=1}^{K}$. Concrete instantiations vary in how the $K$ distributions are leveraged, including weighted empirical risk minimization~\citep{jiang2022learning}, multi-teacher knowledge distillation~\citep{bi2022learning}, and meta-learning over distribution-conditioned tasks~\citep{zhou2023towards}, but they share a common premise: the support of the training distribution is enlarged beyond a single $\mathcal{D}_{\text{train}}$. A canonical realization is the mixture form
\begin{equation}
    \theta^* = \arg\min_\theta \; \sum_{k=1}^{K} w_k \, \mathbb{E}_{x \sim \mathcal{D}_k}\bigl[\, c(\pi_\theta(x)) \,\bigr], \quad \sum_{k=1}^{K} w_k = 1, \; w_k \geq 0,
    \label{eq:multi-dist}
\end{equation}
which subsumes Eq.~\eqref{eq:single-dist} as the special case $K=1$.

\textbf{Remark.}
The two regimes also differ sharply in their underlying assumption about what is needed for cross-distribution generalization. Existing work has overwhelmingly favored multi-distribution training, building increasingly elaborate machinery on top of it, such as multi-teacher distillation, meta-learning over diverse task families, and distribution-specific architectural components, under the implicit premise that single-distribution training is fundamentally insufficient. This trend has, perhaps inadvertently, led the community to underestimate what single-distribution training alone can achieve. In this work, we revisit this premise and ask how far cross-distribution generalization can be pushed without enlarging $\mathcal{D}_{\text{train}}$, showing that with the right design, the ceiling of single-distribution training is substantially higher than previously assumed.

\begin{figure*}[t]
    \centering
    \includegraphics[width=0.97\textwidth]{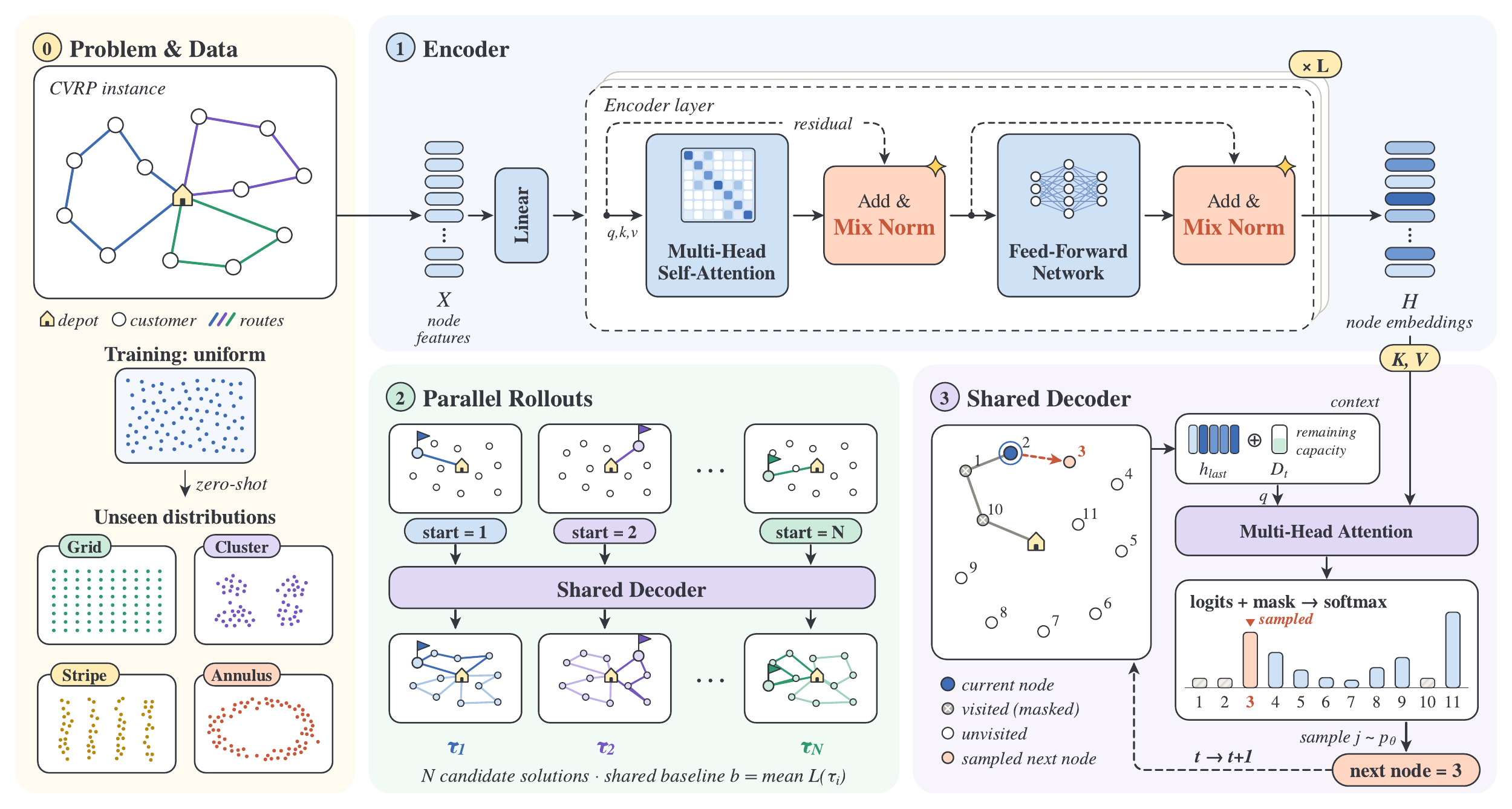}
    \vspace{2pt}
    \caption{Overview of the MixRoute framework. (1) The encoder consists of $L$ Transformer layers, each comprising multi-head self-attention and a feed-forward network, each followed by Add\,\&\,Mix\,Norm. (2) parallel rollouts: each node serves as a starting point, and all trajectories are decoded in parallel. (3) The autoregressive decoder constructs the solution one node at a time.}
    \label{fig:framework}
\end{figure*}

\textbf{Overview.} Figure~\ref{fig:framework} illustrates the overall framework of \textbf{MixRoute}. We follow the encoder-decoder paradigm of POMO~\citep{pomo}, with our key modification being the replacement of the original norm module with the proposed Mix Norm. Given a routing instance with $N$ nodes, MixRoute proceeds in three stages: (i) the encoder embeds all nodes into a shared representation through $L$ stacked Transformer layers, where each layer consists of a multi-head self-attention block and a feed-forward network, both followed by an Add\,\&\,Mix\,Norm module; (ii) parallel rollouts leverage the POMO multi-start strategy by treating each of the $N$ nodes as a starting point and decoding $N$ trajectories; (iii) the decoder autoregressively constructs each solution by attending to the encoded node representations and selecting the next node via a softmax over the unvisited candidates.

\textbf{Add \& Mix Norm Module.} Given input embeddings $x \in \mathbb{R}^{B \times N \times D}$ (batch size $B$, $N$ nodes, embedding dim $D$), Mix Norm produces three independently normalized representations using IN, LN, and BN, each equipped with its own learnable affine parameters:
\begin{equation}
\hat{x}_{\text{IN}} = \mathrm{IN}(x), \quad
\hat{x}_{\text{LN}} = \mathrm{LN}(x), \quad
\hat{x}_{\text{BN}} = \mathrm{BN}(x).
\end{equation}
To combine these representations, we introduce an input-adaptive gate that produces per-token mixing weights conditioned on both the token itself and a global instance context. Specifically, we compute the global context as the mean of the input across the node dimension, $c = \tfrac{1}{N}\sum_{n=1}^{N} x_n \in \mathbb{R}^{B \times 1 \times D}$,{\parfillskip=0pt\par}
\vspace{-\parskip}
\begin{wrapfigure}[22]{r}{0.44\textwidth}
    \vspace{-3pt}
    \centering
    \includegraphics[width=0.44\textwidth]{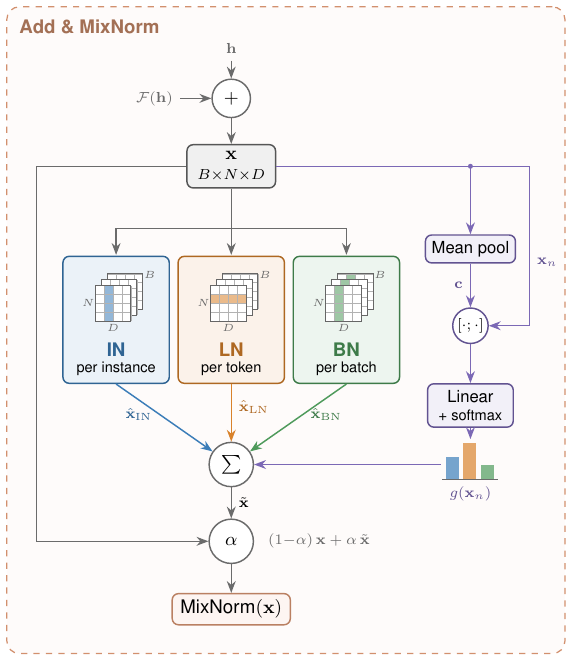}
    \vspace{-14pt}
    \caption{The Add \& MixNorm module.}
    \label{fig:mixnorm}
\end{wrapfigure}
\noindent and broadcast it back to each token. The gate then takes the concatenation $[x_n; c]$ as input and produces a $3$-way distribution over the normalizers:
\begin{equation}
g(x_n) = \mathrm{softmax}\big(W_g [x_n; c] + b_g\big) \in \Delta^{3},
\end{equation}
where $W_g \in \mathbb{R}^{3 \times 2D}$ and $b_g \in \mathbb{R}^{3}$ are learnable parameters and $\Delta^{3}$ denotes the probability simplex. The mixed representation is then a weighted combination of the three normalized outputs:
\begin{equation}
\tilde{x}_n = \sum_{k \in \{\text{IN},\text{LN},\text{BN}\}} g_k(x_n)\,\hat{x}_{k,n}.
\end{equation}
To stabilize training and preserve the original feature when normalization is unhelpful, we introduce a learnable scalar $\alpha \in [0, 1]$ that interpolates between the input and the mixed output:
\begin{equation}
\mathrm{MixNorm}(x) = (1 - \alpha)\, x + \alpha\, \tilde{x}.
\end{equation}
We initialize the gate bias as $b_g = [0, 0, 1]$, which biases the gate toward BN at the start of training. The gate then adapts these weights on its own; the resulting trajectories are reported in Appendix~\ref{sec:gate_analysis}.
\section{Experiments}

\subsection{Problem-Level Generality}
\label{sec:mixroute}

\textbf{Setup.}
We evaluate the zero-shot generalization capability of MixRoute on TSP, CVRP and CVRPTW without any fine-tuning or task-specific adaptation. The test benchmark covers 12 categories of distribution shift, comprising at least 54 datasets, all of size $100$. The shifts are organized along two dimensions: node coordinate distributions and problem-specific constraint shifts, as summarized in Table~\ref{tab:benchmark}. Each test set contains 128 instances, and the full specification of every sub-variant is detailed in Appendix~\ref{app:benchmark_details}. Crucially, none of these distributions are observed during training. 

\textbf{Training settings.}
We adopt the standard POMO~\citep{pomo} backbone, with $N=100$ starting nodes per instance. The encoder consists of $6$ Transformer layers with embedding dimension $128$, $8$ attention heads; logit clipping is applied with $C=10$. All models are trained on uniformly distributed instances with problem size $100$ using the Adam optimizer (learning rate $1 \times 10^{-4}$, weight decay $1 \times 10^{-6}$), with a learning rate scheduler decaying by $\gamma = 0.1$ at epochs $6401$ and $6451$. Training is performed for $6500$ epochs, each consisting of $10{,}000$ instances with a batch size of $64$. All experiments are conducted on a single NVIDIA RTX 3090 GPU with 24GB of memory.

\begin{table*}[!htbp]
\centering
\small
\caption{Composition of the cross-distribution benchmark across three routing problems (TSP, CVRP, CVRPTW), all of size 100. Vertical bars separate node coordinate shifts (left) from problem-specific shifts (right).}
\label{tab:benchmark}
\resizebox{\textwidth}{!}{%
\begin{tabular}{l|ccccccccc|ccc|c}
\toprule
& \multicolumn{9}{c|}{\textbf{Node coordinate distributions}} & \multicolumn{3}{c|}{\textbf{Problem-specific shifts}} & \\
\textbf{Task} & Annulus & Beta & Cluster & Expansion & Explosion & Gaussian & Grid & Implosion & Stripe & Demand & Capacity & TW & \textbf{Total} \\
\midrule
TSP    & 6 & 6 & 6 & 6 & 6 & 4 & 6 & 6 & 8 & --- & --- & --- & \textbf{54} \\
CVRP   & 6 & 6 & 6 & 6 & 6 & 4 & 6 & 6 & 8 & 5 & 5 & --- & \textbf{64} \\
CVRPTW & 6 & 6 & 6 & 6 & 6 & 4 & 6 & 6 & 8 & --- & ---& 7 & \textbf{61} \\
\bottomrule
\end{tabular}}
\end{table*}

\textbf{Baselines.}
We compare MixRoute against three distinct categories of baselines. For all neural baselines, we use their officially released pre-trained checkpoints.
\textit{(i) Classical solvers.} We include LKH3~\citep{lkh} (max trials = 100) and additionally HGS~\citep{hgs} (time limit = 10s) since LKH3 fails to produce feasible solutions on a considerable fraction of CVRPTW instances.
\textit{(ii) Single-distribution neural baselines.} This category covers neural solvers trained exclusively on uniformly distributed instances. We include POMO~\citep{pomo}, ELG~\citep{gao2024towards}, LEHD~\citep{lehd}, BQ~\citep{bqnco}, and INViT~\citep{fang2024invit}. Among these, POMO is the most direct baseline, as MixRoute is built directly on top of POMO.
\textit{(iii) Multi-distribution neural baselines.} To contextualize the strength of our single-source training protocol, we further compare against neural solvers that explicitly leverage distributionally diverse training data. We include AMDKD~\citep{bi2022learning} with its two reported variants AMDKD-AM and AMDKD-POMO, and LEHD-VCT+MEM~\citep{hard-constraint}. The first three are trained on multiple node coordinate distributions, while LEHD-VCT+MEM is instead trained on instances with varying capacity and time window values.

\textbf{Results.}
Table~\ref{tab:main_combined} reports zero-shot generalization results on the TSP, CVRP, and CVRPTW benchmarks, respectively. Across all three problems, MixRoute achieves the lowest average gap among neural methods, and consistently improves over POMO across almost all distribution shifts.

\begin{table*}[!tbp]
\centering
\small
\caption{Zero-shot generalization results across TSP, CVRP, CVRPTW. Gap (\%) is computed against LKH3, except for the CVRPTW where LKH3 produces no feasible solution and HGS is used instead. The Average column is the arithmetic mean over all distributions reported. Best results among neural methods are in \textbf{bold}; second best are \underline{underlined}. Time denotes the average inference time per distribution family.}
\label{tab:main_combined}
\setlength{\tabcolsep}{4pt}
\renewcommand{\arraystretch}{1.15}

\resizebox{\textwidth}{!}{%
\begin{tabular}{l|c|ccccccccc|c|c}
\toprule
\rowcolor{headerbg}
& \textbf{In-dist.} & \multicolumn{9}{c|}{\textbf{Out-of-distribution: Node coordinate shifts}} & & \\
\rowcolor{headerbg}
\textbf{Method (TSP)} & Uniform & Annulus & Beta & Cluster & Expansion & Explosion & Gaussian & Grid & Implosion & Stripe & \textbf{Avg} & \textbf{Time} \\
\midrule
\rowcolor{classicalbg} LKH3 & --- & --- & --- & --- & --- & --- & --- & --- & --- & --- & --- & 4.31s \\
\midrule
\multicolumn{13}{l}{\textit{Single-distribution training}} \\
\rowcolor{singlebg} ELG    & 0.66\% & 0.79\% & 1.03\% & 2.45\% & 0.79\% & 0.57\% & 1.03\% & 1.12\% & 0.96\% & 1.70\% & 1.11\% & 3.26s \\
\rowcolor{singlebg} LEHD   & \underline{0.57\%} & \underline{0.62\%} & 5.71\% & \underline{1.62\%} & \underline{0.64\%} & \underline{0.50\%} & \textbf{0.73\%} & \underline{0.94\%} & \textbf{0.69\%} & 5.03\% & 1.71\% & 3.82s \\
\rowcolor{singlebg} INViT  & 3.80\% & 4.30\% & 4.15\% & 5.90\% & 3.88\% & 3.85\% & 4.13\% & 3.51\% & 4.37\% & 5.36\% & 4.33\% & 13.99m \\
\midrule
\multicolumn{13}{l}{\textit{Multi-distribution training}} \\
\rowcolor{multibg} AMDKD-AM     & 3.84\% & 6.36\% & 11.91\% & 12.74\% & 5.21\% & 3.70\% & 4.75\% & 4.33\% & 4.54\% & 12.07\% & 6.95\% & 0.98s \\
\rowcolor{multibg} AMDKD-POMO   & 1.02\% & 1.49\% & \textbf{0.63\%} & \textbf{1.12\%} & 0.81\% & 0.84\% & 0.94\% & 1.43\% & 0.89\% & \underline{1.35\%} & \underline{1.05\%} & 0.97s \\
\midrule
\rowcolor{ourbg} POMO         & 0.88\% & 1.13\% & 1.86\% & 3.89\% & 1.30\% & 0.77\% & 1.55\% & 1.28\% & 1.30\% & 2.99\% & 1.70\% & 0.74s \\
\rowcolor{ourbg} \textbf{MixRoute} & \textbf{0.54\%} & \textbf{0.52\%} & \underline{0.69\%} & 2.12\% & \textbf{0.61\%} & \textbf{0.45\%} & \underline{0.91\%} & \textbf{0.93\%} & \underline{0.78\%} & \textbf{1.14\%} & \textbf{0.87\%} & 0.75s \\
\bottomrule
\end{tabular}}

\vspace{0.5em}

\resizebox{\textwidth}{!}{%
\begin{tabular}{l|c|ccccccccc|cc|c|c}
\toprule
\rowcolor{headerbg}
& \textbf{In-dist.} & \multicolumn{9}{c|}{\textbf{Out-of-distribution: Node coordinate shifts}} & \multicolumn{2}{c|}{\textbf{CVRP-specific shifts}} & & \\
\rowcolor{headerbg}
\textbf{Method (CVRP)} & Uniform & Annulus & Beta & Cluster & Expansion & Explosion & Gaussian & Grid & Implosion & Stripe & Demand & Capacity & \textbf{Avg} & \textbf{Time} \\
\midrule
\rowcolor{classicalbg} LKH3 & --- & --- & --- & --- & --- & --- & --- & --- & --- & --- & --- & --- & --- & 15.41s \\
\midrule
\multicolumn{15}{l}{\textit{Single-distribution training}} \\
\rowcolor{singlebg} ELG    & -0.38\% & 0.31\% & 2.10\% & 3.52\% & 0.85\% & 0.18\% & 2.16\% & 1.65\% & 0.29\% & 3.02\% & 1.94\% & 4.89\% & 1.71\% & 4.18s \\
\rowcolor{singlebg} LEHD   & 2.77\% & 0.61\% & 3.92\% & 3.96\% & 0.45\% & 0.33\% & 3.19\% & 2.60\% & 0.15\% & 2.92\% & 2.88\% & 3.97\% & 2.31\% & 4.44s \\
\rowcolor{singlebg} BQ     & 0.02\% & 1.00\% & 4.83\% & 2.90\% & 1.36\% & 0.80\% & 2.91\% & 2.67\% & 0.76\% & 3.02\% & 2.92\% & 5.18\% & 2.36\% & 2.41m \\
\rowcolor{singlebg} INViT  & 6.13\% & 6.65\% & 10.76\% & 10.60\% & 6.70\% & 7.15\% & 9.75\% & 7.60\% & 6.38\% & 10.45\% & 9.69\% & 11.73\% & 8.63\% & 15.01m \\
\midrule
\multicolumn{15}{l}{\textit{Multi-distribution training}} \\
\rowcolor{multibg} AMDKD-AM     & 2.15\% & 2.59\% & 5.22\% & 6.88\% & 3.34\% & 2.88\% & 5.97\% & 4.14\% & 3.23\% & 4.91\% & 4.78\% & 7.12\% & 4.43\% & 1.28s \\
\rowcolor{multibg} AMDKD-POMO   & -0.83\% & 1.13\% & \underline{1.83\%} & \underline{2.37\%} & \underline{0.13\%} & -0.01\% & 1.79\% & 1.64\% & \underline{-0.25\%} & \underline{1.77\%} & 1.87\% & 5.53\% & \underline{1.41\%} & 1.28s \\
\rowcolor{multibg} LEHD-VCT+MEM & 0.69\% & \underline{0.27\%} & 3.68\% & 3.11\% & 1.89\% & 1.81\% & 2.75\% & 2.13\% & 0.96\% & 2.66\% & 2.59\% & \textbf{1.68\%} & 2.02\% & 8.06s \\
\midrule
\rowcolor{ourbg} POMO         & \underline{-0.93\%} & 7.11\% & 2.95\% & 3.25\% & 0.65\% & \underline{-0.19\%} & \underline{1.72\%} & \underline{1.38\%} & -0.15\% & 8.30\% & \underline{1.49\%} & 8.34\% & 2.83\% & 2.03s \\
\rowcolor{ourbg} \textbf{MixRoute} & \textbf{-1.03\%} & \textbf{-0.87\%} & \textbf{1.11\%} & \textbf{1.82\%} & \textbf{-0.13\%} & \textbf{-0.28\%} & \textbf{1.36\%} & \textbf{1.18\%} & \textbf{-0.30\%} & \textbf{1.13\%} & \textbf{1.24\%} & \underline{3.36\%} & \textbf{0.72\%} & 2.30s \\
\bottomrule
\end{tabular}}

\vspace{0.5em}

\resizebox{\textwidth}{!}{%
\begin{tabular}{l|c|ccccccccc|ccc|c|c}
\toprule
\rowcolor{headerbg}
& \textbf{In-dist.} & \multicolumn{9}{c|}{\textbf{OOD: Node coordinate shifts}} & \multicolumn{3}{c|}{\textbf{Time Window shifts}} & & \\
\rowcolor{headerbg}
\textbf{Method (CVRPTW)} & Uniform & Annulus & Beta & Cluster & Expansion & Explosion & Gaussian & Grid & Implosion & Stripe & Width & Timing & Coupling & \textbf{Avg} & \textbf{Time} \\
\midrule
\rowcolor{classicalbg} HGS & --- & --- & --- & --- & --- & --- & --- & --- & --- & --- & --- & --- & --- & --- & 763.79s \\
\midrule
\rowcolor{multibg} LEHD-VCT+MEM & 19.39\% & 23.44\% & 22.60\% & 24.16\% & 22.15\% & 19.71\% & 23.44\% & 19.10\% & 19.15\% & 23.53\% & 12.82\% & \underline{-4.70\%} & 18.52\% & 18.72\% & 1.69s \\
\midrule
\rowcolor{ourbg} POMO              & \underline{3.57\%} & \underline{4.99\%} & \underline{5.37\%} & \underline{5.86\%} & \underline{4.72\%} & \underline{3.77\%} & \underline{4.16\%} & \underline{3.76\%} & \underline{4.01\%} & \underline{5.02\%} & \textbf{7.73\%} & 0.00\% & \underline{3.83\%} & \underline{4.37\%} & 3.90s \\
\rowcolor{ourbg} \textbf{MixRoute} & \textbf{3.19\%} & \textbf{4.01\%} & \textbf{4.42\%} & \textbf{2.45\%} & \textbf{3.61\%} & \textbf{3.56\%} & \textbf{2.20\%} & \textbf{3.47\%} & \textbf{3.68\%} & \textbf{3.80\%} & \underline{12.55\%} & \textbf{-15.98\%} & \textbf{3.50\%} & \textbf{2.65\%} & 4.02s \\
\bottomrule
\end{tabular}}

\end{table*}

\subsection{Architecture-Level Generality}
To verify the generality of our framework across different architectures, we apply \textsc{MixNorm} to MatNet~\citep{matnet} and use it to solve the Asymmetric CVRP (ACVRP). MixRoute retains this exact setup and only replaces the standard Add \& Norm sub-layer with our Add \& MixNorm module. It is worth noting that all type-A instances are post-processed with the triangle inequality constraint, which sufficiently homogenizes the resulting distance matrices and removes meaningful distributional shift along this axis. We accordingly restrict our evaluation to the demand axis.

As shown in Table~\ref{tab:matnet_acvrp}, MixRoute outperforms MatNet on all five demand distributions. The improvement also holds on the in-distribution setting. These gains incur negligible overhead in inference time ($0.65$s to $0.69$s), confirming that MixNorm transfers to a different backbone as a drop-in module.

\begin{table}[htbp]
\centering
\fontsize{6.5pt}{7.5pt}\selectfont
\caption{Generalization performance on ACVRP100 across different demand distributions. Gap (\%) is computed against LKH3. Best results among neural methods are in \textbf{bold}. Time is the total inference time over the test set.}
\label{tab:matnet_acvrp}
\vspace{6pt}
\renewcommand{\arraystretch}{0.9}
\setlength{\tabcolsep}{2.5pt}
\begin{tabularx}{\textwidth}{l|C|CCCCC|C|C}
\toprule
\rowcolor{headerbg}
& \textbf{In-dist.} & \multicolumn{5}{c|}{\textbf{Out-of-distribution: Demand distribution shifts}} & & \\
\rowcolor{headerbg}
\textbf{Method} & Uniform & Small-biased & Large-biased & U-shaped & Mid-peak & Bimodal & \textbf{Average Gap} & \textbf{Time} \\
\midrule
\rowcolor{classicalbg} LKH3 & --- & --- & --- & --- & --- & --- & --- & 1.63h \\
\midrule
\rowcolor{ourbg} MatNet & -1.49\%& 1.90\% & -3.92\%& -0.82\%& -1.24\%& -0.78\%& -1.06\%& 0.65s \\
\rowcolor{ourbg} \textbf{MixRoute} & \textbf{-2.02\%}& \textbf{1.06\%}& \textbf{-4.58\%}& \textbf{-1.12\%}& \textbf{-1.74\%}&\textbf{-1.35\%} & \textbf{-1.63\%} & 0.69s \\
\bottomrule
\end{tabularx}
\end{table}


\subsection{Normalization-Level Generality}
\label{sec:ablation_norms}

To probe this, we evaluate MixRoute on the CVRP benchmark under three different normalization combinations (IN+LN+NoNorm, IN+LN+BN, and IN+LN+RMSNorm~\citep{rms}), chosen to span normalizations of qualitatively different characters: identity (no normalization), batch-level statistics, and root-mean-square scaling without mean centering.

As shown in Table~\ref{tab:mixing}, this is precisely what we observe. All three MixRoute variants substantially outperform POMO, reducing the average gap from $3.17\%$ to between $0.88\%$ and $1.41\%$. The IN+LN+BN variant performs marginally best ($0.88\%$ Avg), but the differences among the three variants are small relative to the gap between any of them and POMO. This pattern indicates that \textsc{MixRoute}'s effectiveness is robust to the constituent choice. 

\begin{table*}[!htbp]
\centering
\small
\caption{Robustness of Mix Norm to the choice of constituent normalizations. We compare three MixRoute variants with different normalization combinations against the POMO baseline. Gap (\%) is computed against LKH3. Best results among neural methods are in \textbf{bold}; second best are \underline{underlined}. Shaded cells indicate improvement over POMO. Time denotes the average inference time per distribution family.}
\label{tab:mixing}
\setlength{\tabcolsep}{4pt}
\renewcommand{\arraystretch}{1.15}
\resizebox{\textwidth}{!}{%
\begin{tabular}{ll|ccccccccc|cc|c|c}
\toprule
\rowcolor{headerbg}
& & \multicolumn{9}{c|}{\textbf{Node coordinate distributions}} & \multicolumn{2}{c|}{\textbf{CVRP-specific}} & & \\
\rowcolor{headerbg}
\textbf{Method} & \textbf{Norms} & Annulus & Beta & Cluster & Expansion & Explosion & Gaussian & Grid & Implosion & Stripe & Demand & Capacity & \textbf{Avg} & \textbf{Time} \\
\midrule
LKH3 & --- & --- & --- & --- & --- & --- & --- & --- & --- & --- & --- & --- & --- & --- \\
\midrule
POMO & IN (original) & 7.11\% & 2.95\% & 3.25\% & 0.65\% & -0.19\% & 1.72\% & 1.38\% & -0.15\% & 8.30\% & 1.49\% & 8.34\% & 3.17\% & 2.03s \\
\midrule
MixRoute & IN+LN+NoNorm
& \cellcolor{classicalbg}\textbf{-0.91\%}
& \cellcolor{classicalbg}\textbf{1.01\%}
& \cellcolor{classicalbg}\underline{1.97\%}
& \cellcolor{classicalbg}\underline{-0.16\%}
& \cellcolor{classicalbg}\underline{-0.32\%}
& \cellcolor{classicalbg}\underline{1.52\%}
& \cellcolor{classicalbg}\underline{1.20\%}
& \cellcolor{classicalbg}\underline{-0.33\%}
& \cellcolor{classicalbg}\textbf{1.08\%}
& \cellcolor{classicalbg}\textbf{1.21\%}
& \cellcolor{classicalbg}\underline{4.22\%}
& \cellcolor{classicalbg}\underline{0.95\%}
& 2.18s \\
MixRoute & IN+LN+BN
& \cellcolor{singlebg}\underline{-0.87\%}
& \cellcolor{singlebg}\underline{1.11\%}
& \cellcolor{singlebg}\textbf{1.82\%}
& \cellcolor{singlebg}-0.13\%
& \cellcolor{singlebg}-0.28\%
& \cellcolor{singlebg}\textbf{1.36\%}
& \cellcolor{singlebg}\textbf{1.18\%}
& \cellcolor{singlebg}-0.30\%
& \cellcolor{singlebg}\underline{1.13\%}
& \cellcolor{singlebg}\underline{1.24\%}
& \cellcolor{singlebg}\textbf{3.36\%}
& \cellcolor{singlebg}\textbf{0.88\%}
& 2.30s \\
MixRoute & IN+LN+RMSNorm
& \cellcolor{blue1}-0.51\%
& \cellcolor{blue1}1.58\%
& \cellcolor{blue1}2.75\%
& \cellcolor{blue1}\textbf{-0.18\%}
& \cellcolor{blue1}\textbf{-0.47\%}
& 1.82\%
& 1.43\%
& \cellcolor{blue1}\textbf{-0.38\%}
& \cellcolor{blue1}1.92\%
& 1.73\%
& \cellcolor{blue1}5.79\%
& \cellcolor{blue1}1.41\%
& 2.21s \\
\bottomrule
\end{tabular}}
\end{table*}

\subsection{Generalization to Real World Benchmark}

\begin{figure}[htbp]
    \centering
\includegraphics[width=0.8\textwidth]{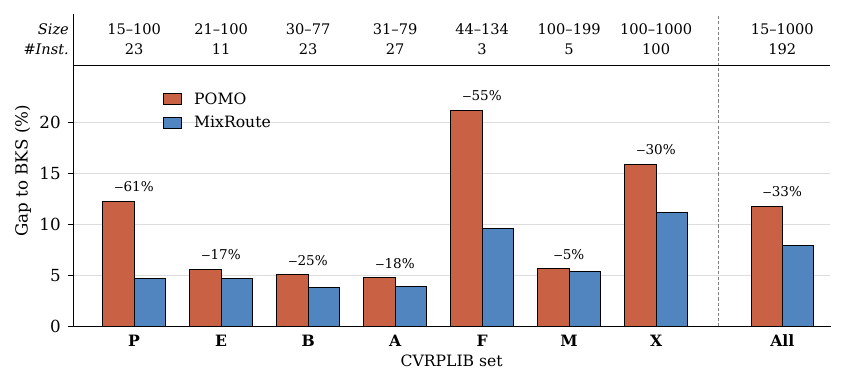}
    \vspace{2pt}
    \caption{Zero-shot generalization of POMO and MixRoute on seven CVRPLIB sets. Bars report the average optimality gap (\%) with respect to the best known solutions (BKS). The number above each pair of bars is the relative gap reduction of MixRoute over POMO.}
    \label{fig:cvrplib}
\end{figure}

We further evaluate MixRoute on CVRPLIB \citep{uchoa2017new}, a widely used real-world benchmark whose instances are drawn from practical routing applications. CVRPLIB instances exhibit heterogeneous customer layouts, demand patterns, and capacity constraints that are not generated from any parametric distribution, and their sizes range from 15 to 1000 nodes. We evaluate on seven sets using the same checkpoints as in Section~\ref{sec:mixroute} without fine-tuning. As shown in Figure~\ref{fig:cvrplib}, MixRoute improves over POMO on every set, reducing the overall gap from $11.80\%$ to $7.94\%$. 

\subsection{Analysis}

\subsubsection{What Matters is Mixing: Robustness Across Variants}
\label{sec:ablation_mixing}

We categorize mixing strategies into two types based on what is being mixed: \emph{output-level mixing}, which combines the outputs of multiple normalization layers via learnable weights, and \emph{statistic-level mixing}, which fuses the underlying normalization statistics before applying a single affine transformation. Full descriptions of each mixing scheme are provided in Appendix~\ref{app:mixing_type}.

For each variant, we further examine two combinations of constituent normalizations: \texttt{IN+LN+NoNorm} and \texttt{IN+LN+BN}. The first combination treats \texttt{NoNorm} as a constituent; the second replaces it with the standard Batch Normalization. Table~\ref{tab:ablation_mixing} shows that all mixing variants substantially outperform POMO (Avg 3.17\%), with the worst-performing variant (Switch Norm IN+LN, 1.68\%) still nearly halving POMO's gap. This confirms that the gain originates from the act of mixing itself. 

\begin{table*}[!htbp]
\centering
\small
\caption{Ablation on Mix Norm variants. Best results are in \textbf{bold}; second best are \underline{underlined}. Color depth indicates improvement magnitude over POMO: light ($<$1\%), medium (1--5\%), and dark ($\geq$5\%).}
\label{tab:ablation_mixing}
\setlength{\tabcolsep}{4pt}
\renewcommand{\arraystretch}{1.15}
\resizebox{\textwidth}{!}{%
\begin{tabular}{ll|ccccccccc|cc|c}
\toprule
\rowcolor{headerbg}
& & \multicolumn{9}{c|}{\textbf{Node coordinate distributions}} & \multicolumn{2}{c|}{\textbf{CVRP-specific}} & \\
\rowcolor{headerbg}
\textbf{Method} & \textbf{Norms} & Annulus & Beta & Cluster & Expansion & Explosion & Gaussian & Grid & Implosion & Stripe & Demand & Capacity & \textbf{Avg} \\
\midrule
LKH3 & --- & --- & --- & --- & --- & --- & --- & --- & --- & --- & --- & --- & --- \\
\midrule
POMO & IN (original) & 7.11\% & 2.95\% & 3.25\% & 0.65\% & -0.19\% & 1.72\% & 1.38\% & -0.15\% & 8.30\% & 1.49\% & 8.34\% & 3.17\% \\
\midrule
\multicolumn{14}{l}{\textit{Statistic mixing}} \\
Switch Norm & IN+LN
& \cellcolor{green3}-0.67\%
& \cellcolor{green2}1.76\%
& \cellcolor{green1}2.47\%
& \cellcolor{green1}0.02\%
& -0.20\%
& \cellcolor{green1}1.51\%
& \cellcolor{green1}1.24\%
& -0.17\%
& \cellcolor{green3}1.39\%
& \cellcolor{green1}1.41\%
& \cellcolor{green2}9.72\%
& \cellcolor{green2}1.68\% \\
Switch Norm & IN+LN+BN
& \cellcolor{green3}-0.68\%
& \cellcolor{green2}1.17\%
& \cellcolor{green2}\underline{1.90\%}
& \cellcolor{green1}-0.11\%
& \cellcolor{green1}-0.28\%
& \cellcolor{green1}\underline{1.45\%}
& \cellcolor{green1}\textbf{1.18\%}
& \cellcolor{green1}\underline{-0.30\%}
& \cellcolor{green3}1.57\%
& \cellcolor{green1}1.22\%
& \cellcolor{green2}8.98\%
& \cellcolor{green2}1.46\% \\
\midrule
\multicolumn{14}{l}{\textit{Output mixing}} \\
Simple Mixing & IN+LN+NoNorm
& \cellcolor{yellow3}-0.70\%
& \cellcolor{yellow2}1.41\%
& \cellcolor{yellow2}2.22\%
& \cellcolor{yellow1}-0.01\%
& \cellcolor{yellow1}-0.24\%
& \cellcolor{yellow1}\underline{1.45\%}
& \cellcolor{yellow1}1.26\%
& \cellcolor{yellow1}-0.24\%
& \cellcolor{yellow3}1.76\%
& \cellcolor{yellow1}1.32\%
& \cellcolor{yellow3}5.43\%
& \cellcolor{yellow2}1.24\% \\
Simple Mixing & IN+LN+BN
& \cellcolor{yellow3}-0.52\%
& \cellcolor{yellow1}1.96\%
& \cellcolor{yellow1}2.53\%
& \cellcolor{yellow1}\textbf{-0.27\%}
& \cellcolor{yellow1}\textbf{-0.47\%}
& \cellcolor{yellow1}1.60\%
& \cellcolor{yellow1}1.31\%
& \cellcolor{yellow1}\textbf{-0.45\%}
& \cellcolor{yellow3}1.84\%
& 1.49\%
& \cellcolor{yellow3}\underline{4.15\%}
& \cellcolor{yellow2}1.20\% \\
MixRoute (ours) & IN+LN+NoNorm
& \cellcolor{blue3}\textbf{-0.91\%}
& \cellcolor{blue2}\textbf{1.01\%}
& \cellcolor{blue2}1.97\%
& \cellcolor{blue1}\underline{-0.16\%}
& \cellcolor{blue1}\underline{-0.32\%}
& \cellcolor{blue1}1.52\%
& \cellcolor{blue1}\underline{1.20\%}
& \cellcolor{blue1}\underline{-0.33\%}
& \cellcolor{blue3}\textbf{1.08\%}
& \cellcolor{blue1}\textbf{1.21\%}
& \cellcolor{blue3}4.22\%
& \cellcolor{blue2}\underline{0.95\%} \\
MixRoute (ours) & IN+LN+BN
& \cellcolor{blue3}\underline{-0.87\%}
& \cellcolor{blue2}\underline{1.11\%}
& \cellcolor{blue2}\textbf{1.82\%}
& \cellcolor{blue1}-0.13\%
& \cellcolor{blue1}-0.28\%
& \cellcolor{blue1}\textbf{1.36\%}
& \cellcolor{blue1}\textbf{1.18\%}
& \cellcolor{blue1}\underline{-0.30\%}
& \cellcolor{blue3}\underline{1.13\%}
& \cellcolor{blue1}\underline{1.24\%}
& \cellcolor{blue3}\textbf{3.36\%}
& \cellcolor{blue2}\textbf{0.88\%} \\
\bottomrule
\end{tabular}}
\end{table*}

\subsection{Additional Results}

Due to space constraints, we defer a number of supplementary experiments and analyses to the appendix. Table~\ref{tab:appendix_index} summarizes their content and corresponding location for reference.

\begin{table}[htbp]
\centering
\small
\caption{Overview of additional results deferred to the appendix.}
\label{tab:appendix_index}
\vspace{6pt}
\renewcommand{\arraystretch}{1.1}
\begin{tabularx}{\textwidth}{llX}
\toprule
\rowcolor{headerbg}
\textbf{Topic} & \textbf{Section} & \textbf{Content} \\
\midrule
Multi-seed results & Appendix~\ref{app:seeds} & Seed variability of MixRoute. \\
Mechanism of MixNorm & Appendix~\ref{sec:analysis} & How MixNorm responds to distributional shift. \\
Behavior of the adaptive gate & Appendix~\ref{sec:gate_analysis} & Gate weights across layers and distributions. \\
Related normalizations & Appendix~\ref{sec:compare_existing} & Comparison with existing Mix Norm. \\
Normalizer choice & Appendix~\ref{app:pool} & Effect of pool size and composition. \\
\bottomrule
\end{tabularx}
\end{table}
\section{Conclusion}

We revisit cross-distribution generalization in neural routing solvers from a perspective that has received little attention: the role of normalization. While prior work has largely assumed single-distribution training to be fundamentally insufficient, pursuing increasingly elaborate multi-distribution machinery, we show that its ceiling is, in fact, substantially higher than previously assumed. By identifying normalization layers as a key source of distribution-specific entanglement and proposing MixNorm, a simple plug-in module combining complementary normalization schemes, we demonstrate that strong cross-distribution generalization is achievable from a single uniform training distribution alone. Extensive experiments across three routing problems, four shift dimensions, and 179 out-of-distribution settings confirm that MixRoute consistently improves diverse backbones, validating it as an architecture-agnostic and broadly applicable component. We hope this work encourages the community to reconsider single-distribution training not as a limitation to circumvent, but as a stress test that exposes the true inductive biases of a solver.

\section*{Acknowledgments}
This research is supported by the Ministry of Education, Singapore under its Academic Research Fund (AcRF) Tier 2 Grant Call (Proposal ID:  T2EP20225-0056). Any opinions, findings and conclusions or recommendations expressed in this material are those of the author(s) and do not reflect the views of the Ministry of Education, Singapore. This research is also supported by the Lee Kong Chian Fellowship awarded to CAO Zhiguang by Singapore Management University.

\bibliographystyle{plainnat}
\bibliography{neurips_references} 

\newpage
\appendix

\vbox{
\hrule height 4pt
\vskip 0.2in
\vskip -\parskip%
\centering
{\Large\bf MixRoute: Rethinking Single-Distribution Training
for Generalizable Neural Routing (Appendix)} 
\vskip 0.2in
\vskip -\parskip
\hrule height 1pt
}

\section{Preliminaries}

\subsection{Vehicle Routing Problems}
\label{sec:prelim_vrp}

We consider three representative VRP variants: the Traveling Salesman Problem (TSP), the Capacitated VRP (CVRP), and the Capacitated VRP with Time Windows (CVRPTW). All are defined on a complete graph $\mathcal{G} = (\mathcal{V}, \mathcal{E})$, where $\mathcal{V} = \{v_0, v_1, \ldots, v_N\}$ consists of a depot $v_0$ and $N$ customer nodes (in the case of TSP, $v_0$ is the starting node and there is no notion of a depot), and each edge is associated with a Euclidean travel cost $c_{ij}$. For CVRP and CVRPTW, a fleet of vehicles, each with capacity $Q$, is stationed at the depot.

\textbf{TSP.} A feasible solution $\boldsymbol{\tau}$ is a single Hamiltonian tour that visits every node exactly once and returns to the starting node, with the objective of minimizing the total tour length.

\textbf{CVRP.} Each customer $v_i$ has a demand $d_i \in (0, Q]$. A feasible solution $\boldsymbol{\tau}$ is a set of routes that partitions the customer set, where each route starts and ends at the depot, visits each assigned customer exactly once, and respects the capacity constraint. The objective is to minimize the total travel cost
\begin{equation}
    \mathcal{L}(\boldsymbol{\tau}) = 
    \sum_{(v_i, v_j) \in \boldsymbol{\tau}} c_{ij}.
\end{equation}

\textbf{CVRPTW.} CVRPTW extends CVRP by additionally associating each customer $v_i$ with a service time $s_i$ and a time window $[e_i, l_i]$, within which service must commence. Vehicles arriving before $e_i$ must wait until the window opens, while arrivals after $l_i$ render the route infeasible.

\subsection{Reinforcement Learning Formulation}
\label{sec:prelim_rl}

Following the standard practice in neural combinatorial 
optimization~\citep{am, pomo}, we formulate VRPs as a sequential decision 
process and learn a construction policy via reinforcement learning. 
Given an instance $\mathcal{I}$, the policy 
$\pi_{\boldsymbol{\theta}}$ parameterized by $\boldsymbol{\theta}$ 
constructs a solution $\boldsymbol{\tau}$ autoregressively:
\begin{equation}
    \pi_{\boldsymbol{\theta}}(\boldsymbol{\tau} \mid \mathcal{I}) 
    = \prod_{t=1}^{T} \pi_{\boldsymbol{\theta}}(a_t \mid s_t),
\end{equation}
where at each step $t$, the action $a_t$ selects the next node to visit 
based on the current state $s_t$ (i.e., the partial route, remaining 
capacity, and elapsed time). The reward is defined as the negative tour 
length, $R(\boldsymbol{\tau}) = -\mathcal{L}(\boldsymbol{\tau})$, so that 
maximizing expected reward is equivalent to minimizing total travel cost.

The policy is trained with the REINFORCE 
algorithm~\citep{williams1992simple} using a baseline $b(\mathcal{I})$ 
to reduce variance:
\begin{equation}
    \nabla_{\boldsymbol{\theta}} \mathcal{J}(\boldsymbol{\theta}) 
    = \mathbb{E}_{\boldsymbol{\tau} \sim \pi_{\boldsymbol{\theta}}}
    \!\left[ \big(R(\boldsymbol{\tau}) - b(\mathcal{I})\big) 
    \nabla_{\boldsymbol{\theta}} \log \pi_{\boldsymbol{\theta}}
    (\boldsymbol{\tau} \mid \mathcal{I}) \right].
\end{equation}
Following POMO~\citep{pomo}, we exploit the symmetry of routing 
solutions by sampling $M$ trajectories with different starting nodes 
from the same instance and using their mean reward as a shared baseline.

\section{Additional Experimental Details}
\label{app:additional_results}

This section provides supplementary tables that complement the main paper. Specifically, we include detailed results that could not fit within the main paper due to space constraints, organized to mirror the corresponding sections in the main text. Each table is referenced from the main paper at the relevant point of discussion.

\subsection{Comparison with Single Normalizations}
\label{sec:compare_single}

We further compare MixRoute against single-normalization baselines to isolate the gain of mixing from any particular constituent. We replace the Add\,\&\,Norm module in POMO with each individual normalization---NoNorm (an identity mapping with affine transformation), Instance Normalization (IN), Layer Normalization (LN), and Batch Normalization (BN)---and compare them against the two MixRoute variants \texttt{IN+LN+NoNorm} and \texttt{IN+LN+BN}. All models share the same architecture and training protocol; the only difference lies in the normalization layer. 

\begin{table*}[t]
\centering
\small
\caption{Comparison with single normalizations on the CVRP benchmark. We compare POMO equipped with each individual normalization (IN, LN, NoNorm, BN) against MixRoute variants. Best results among neural methods are in \textbf{bold}; second best are \underline{underlined}. \colorbox{winNoNorm}{Blue cells} indicate where MixRoute (IN+LN+NoNorm) outperforms all three single normalizations (IN, LN, NoNorm); \colorbox{winBN}{orange cells} indicate where MixRoute (IN+LN+BN) outperforms all three single normalizations (IN, LN, BN).}
\label{tab:compare_single}
\setlength{\tabcolsep}{4pt}
\renewcommand{\arraystretch}{1.15}
\resizebox{\textwidth}{!}{%
\begin{tabular}{ll|ccccccccc|cc|c}
\toprule
\rowcolor{headerbg}
& & \multicolumn{9}{c|}{\textbf{Node coordinate distributions}} & \multicolumn{2}{c|}{\textbf{CVRP-specific}} & \\
\rowcolor{headerbg}
\textbf{Method} & \textbf{Norms} & Annulus & Beta & Cluster & Expansion & Explosion & Gaussian & Grid & Implosion & Stripe & Demand & Capacity & \textbf{Avg} \\
\midrule
LKH3 & --- & --- & --- & --- & --- & --- & --- & --- & --- & --- & --- & --- & --- \\
\midrule
POMO & IN (original) & 7.11\% & 2.95\% & 3.25\% & 0.65\% & -0.19\% & 1.72\% & 1.38\% & -0.15\% & 8.30\% & 1.49\% & 8.34\% & 3.17\% \\
POMO & LN & -0.38\% & 4.09\% & 2.85\% & 0.03\% & -0.20\% & 2.14\% & 1.43\% & -0.28\% & 2.25\% & 1.59\% & 5.79\% & 1.76\% \\
\midrule
POMO & NoNorm & -0.61\% & 1.68\% & 3.24\% & -0.11\% & -0.27\% & 2.73\% & 1.34\% & \underline{-0.30\%} & 2.01\% & 1.47\% & 4.35\% & 1.41\% \\
\textbf{MixRoute} & IN+LN+NoNorm 
& \cellcolor{winNoNorm}\textbf{-0.91\%}
& \cellcolor{winNoNorm}\textbf{1.01\%}
& \cellcolor{winNoNorm}\underline{1.97\%}
& \cellcolor{winNoNorm}\textbf{-0.16\%}
& \cellcolor{winNoNorm}\textbf{-0.32\%}
& \cellcolor{winNoNorm}1.52\%
& \cellcolor{winNoNorm}1.20\%
& \cellcolor{winNoNorm}\textbf{-0.33\%}
& \cellcolor{winNoNorm}\textbf{1.08\%}
& \cellcolor{winNoNorm}\textbf{1.21\%}
& \cellcolor{winNoNorm}4.22\%
& \cellcolor{winNoNorm}\underline{0.95\%} \\
\midrule
POMO & BN & -0.76\% & 1.26\% & 2.29\% & -0.09\% & -0.20\% & \underline{1.46\%} & \textbf{1.16\%} & -0.22\% & 1.24\% & 1.28\% & \underline{4.19\%} & 1.06\% \\
\textbf{MixRoute} & IN+LN+BN 
& \cellcolor{winBN}\underline{-0.87\%}
& \cellcolor{winBN}\underline{1.11\%}
& \cellcolor{winBN}\textbf{1.82\%}
& \cellcolor{winBN}\underline{-0.13\%}
& \cellcolor{winBN}\underline{-0.28\%}
& \cellcolor{winBN}\textbf{1.36\%}
& \underline{1.18\%}
& \cellcolor{winBN}\underline{-0.30\%}
& \cellcolor{winBN}\underline{1.13\%}
& \cellcolor{winBN}\underline{1.24\%}
& \cellcolor{winBN}\textbf{3.36\%}
& \cellcolor{winBN}\textbf{0.88\%} \\
\bottomrule
\end{tabular}}
\end{table*}

\textbf{Results.}
Table~\ref{tab:compare_single} shows that no single normalization dominates across all distributions: each of IN, LN, NoNorm, and BN performs well on some categories but poorly on others. In contrast, the two MixRoute variants achieve the lowest average gaps.

\subsection{Mixing Types}
\label{app:mixing_type}
\textbf{Two types of mixing.}
We categorize mixing strategies into two types based on what is being mixed.

\textit{(i) Output mixing} performs each normalization independently and then linearly combines the outputs:
\begin{equation}
\text{OutputMix}(x) = \sum_{i=1}^{K} \alpha_i \cdot \text{Norm}_i(x),
\end{equation}
where each $\text{Norm}_i(\cdot)$ produces a fully normalized output, and $\{\alpha_i\}_{i=1}^{K}$ are mixing coefficients with $\sum_i \alpha_i = 1$. Output mixing preserves the complete transformation of every constituent normalization.

\textit{(ii) Statistic mixing}, in contrast, fuses the normalizations at a more fundamental level by combining their statistics (mean and variance) prior to a single normalization step:
\begin{equation}
\text{StatMix}(x) = \frac{x - \sum_{i=1}^{K} \alpha_i \, \mu_i}{\sqrt{\sum_{i=1}^{K} \alpha_i \, \sigma_i^2 + \epsilon}},
\end{equation}
where $\mu_i$ and $\sigma_i^2$ denote the mean and variance computed under the $i$-th normalization scheme. Statistic mixing produces a single normalized output whose statistics interpolate between the constituents.

\textbf{Mixing variants.}
Within these two paradigms, we evaluate three representative variants:

\begin{itemize}
    \item \textbf{MixRoute (ours)} --- output mixing with learnable, input-adaptive weights. The mixing coefficients $\alpha_i(x)$ are produced by a lightweight gating network conditioned on the input.
    
    \item \textbf{Simple Mixing} --- output mixing with fixed equal weights. We set $\alpha_i = 1/K$ for all $i$, removing all learnable parameters from the mixing module. 
    
    \item \textbf{Switch Norm}~\citep{luo2019differentiable} --- statistic mixing with learnable scalar weights. Originally proposed for image recognition, Switch Norm learns a single set of scalar coefficients (shared across the input) to combine the means and variances of multiple normalization schemes.
\end{itemize}

\section{Benchmark Details and Full Results}
\label{app:benchmark_details}

In this section, we provide a detailed description of our cross-distribution generalization benchmark. Given that fully reporting per-variant results across all three problems would consume a prohibitive amount of space, we focus the detailed presentation on the CVRP benchmark, which spans the broadest set of shift dimensions in our evaluation. Training \textsc{MixRoute} for 6500 epochs takes approximately 66.97 hours on TSP, 91.15 hours on CVRP, and 106.49 hours on CVRPTW, all measured on a single NVIDIA RTX 3090 GPU. Throughout this appendix, the Average Gap column of each family is computed by pooling all instances of that family.

\subsection{Annulus Distribution}
\label{app:annulus}

\begin{figure}[t]
    \centering
    \includegraphics[width=\textwidth]{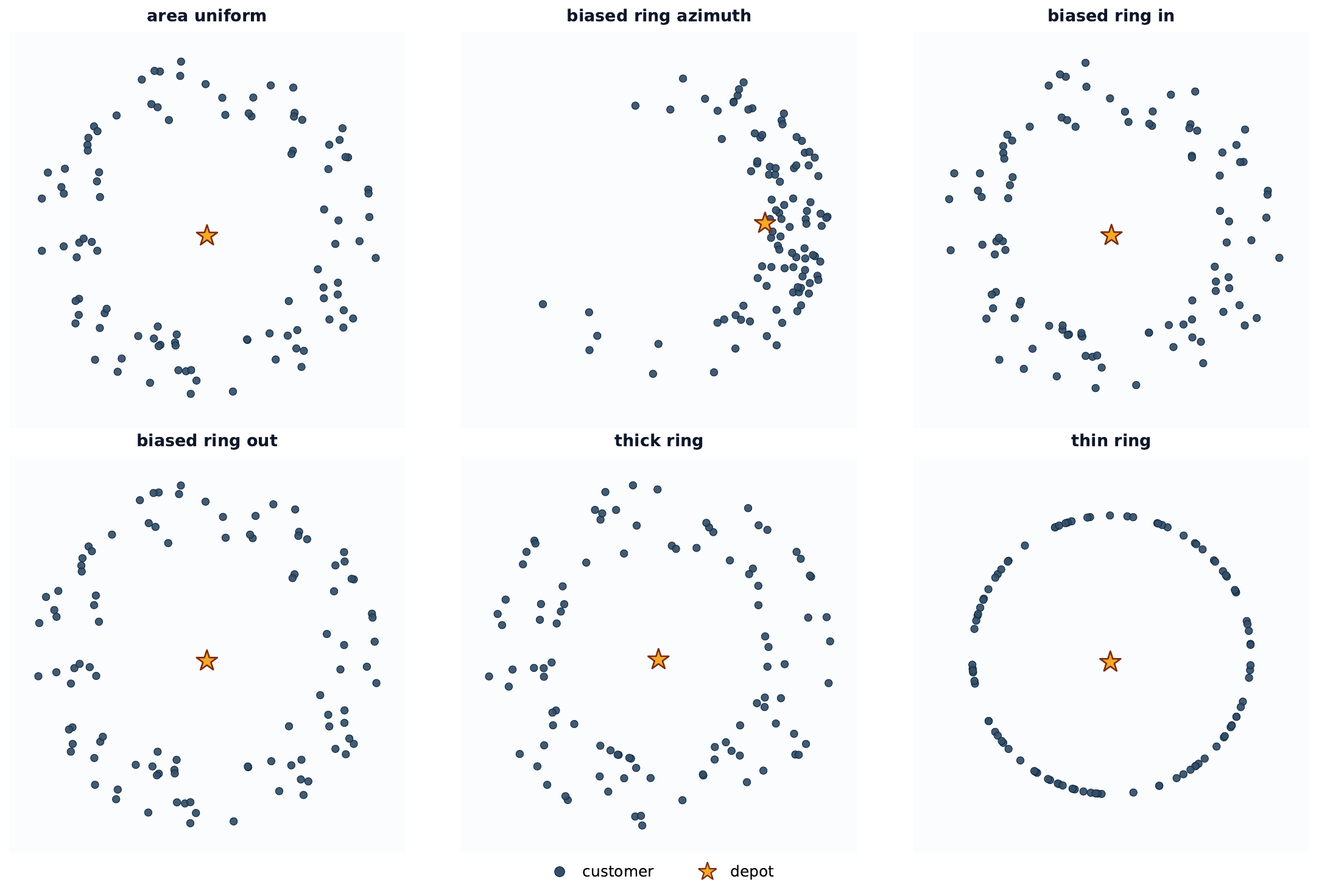}
    \caption{Visualization of the six Annulus distribution variants on CVRP100 instances. The orange star marks the depot at the center.}
    \label{fig:annulus_variants}
\end{figure}

The Annulus distribution samples node coordinates from an annular (ring-shaped) region surrounding a central depot, capturing real-world scenarios where customers are distributed around but not too close to a logistics hub---for example, suburban delivery zones surrounding a city center, or service areas around a transportation terminal. We construct six variants that vary in radial thickness, radial density, and angular density. Figure~\ref{fig:annulus_variants} visualizes one representative instance from each variant.

\textbf{Area-uniform.}
Coordinates are sampled uniformly over the annular region, serving as the unbiased baseline of this family.

\textbf{Biased ring azimuth.}
The angular coordinate is concentrated within a specific sector rather than spread uniformly, modeling one-sided coastal cities or service regions bounded by a natural barrier.

\textbf{Biased ring in.}
Radial sampling skewed toward the inner boundary, producing customer density that thins out with distance from the depot.

\textbf{Biased ring out.}
The inverse of \emph{Biased ring in}: customers concentrate near the outer perimeter, with sparser nodes near the depot.

\textbf{Thick ring.}
A wide annulus that interpolates toward a near-uniform disk distribution with only a small central exclusion zone.

\textbf{Thin ring.}
A narrow annulus concentrating all nodes along a thin circular band, approximating a one-dimensional ring.

\textbf{Results.} Table~\ref{tab:annulus_full} reports the per-variant performance on the Annulus family. \textsc{MixRoute} achieves the lowest gap on five of the six variants and an average gap of $-0.87\%$, surpassing LKH3 in aggregate. The most striking case is \emph{Thin ring}, where POMO catastrophically degrades to $43.25\%$ while \textsc{MixRoute} attains $-3.26\%$ under identical training conditions---a $46.5$-point improvement that highlights how normalization design alone can recover generalization on extreme one-dimensional geometries.

\begin{table*}[t]
\centering
\fontsize{6.5pt}{7.5pt}\selectfont
\caption{Generalization performance on CVRP100 across different Annulus distributions. Gap (\%) is computed against LKH3. Best results among neural methods are in \textbf{bold}; second best are \underline{underlined}.}
\label{tab:annulus_full}
\renewcommand{\arraystretch}{0.9}
\setlength{\tabcolsep}{2.5pt}
\begin{tabularx}{\textwidth}{l|CCCCCC|C|C}
\toprule
\rowcolor{headerbg}
& \multicolumn{6}{c|}{\textbf{Annulus distribution variants}} & & \\
\rowcolor{headerbg}
\textbf{Method} & Area-uniform & Thin ring & Thick ring & Biased-in & Biased-out & Biased-azimuth & \textbf{Average Gap} & \textbf{Time} \\
\midrule
\rowcolor{classicalbg} LKH3 & --- & --- & --- & --- & --- & --- & --- & 7.74s \\
\midrule
\multicolumn{9}{l}{\textit{Single-distribution training}} \\
\rowcolor{singlebg} ELG    & -0.10\% & \underline{-2.44\%} & -0.21\% & 0.08\% & -0.31\% & 6.40\% & 0.31\% & 4.35s\\
\rowcolor{singlebg} LEHD   & 0.51\% & -0.90\% & 0.20\% & 0.07\% & 0.16\% & 4.71\% & 0.61\% & 4.27s\\
\rowcolor{singlebg} BQ     & 0.92\% & -1.27\% & 1.69\% & 1.37\% & 1.33\% & \textbf{1.94\%} & 1.00\% & 2.30m\\
\rowcolor{singlebg} INViT  & 5.95\% & 3.42\% & 6.48\% & 6.08\% & 6.58\% & 12.96\% & 6.65\% & 15.20m\\
\midrule
\multicolumn{9}{l}{\textit{Multi-distribution training}} \\
\rowcolor{multibg} AMDKD-AM     & 2.76\% & -0.74\% & 2.40\% & 2.52\% & 2.28\% & 7.43\% & 2.59\% & 1.01s\\
\rowcolor{multibg} AMDKD-POMO   & -0.04\% & 5.06\% & 0.17\% & 0.41\% & -0.01\% & \underline{2.01\%} & 1.13\% & 1.11s\\
\rowcolor{multibg} LEHD-VCT+MEM & 0.12\% & -1.39\% & 0.04\% & -0.07\% & 0.02\% & 3.83\% & \underline{0.27\%} & 8.12s\\
\midrule
\rowcolor{ourbg} POMO              & \underline{-0.50\%} & 43.25\% & \underline{-0.50\%} & \underline{-0.28\%} & \underline{-0.63\%} & 5.89\% & 7.11\% & 2.06s\\
\rowcolor{ourbg} \textbf{MixRoute} & \textbf{-0.88\%} & \textbf{-3.26\%} & \textbf{-0.95\%} & \textbf{-0.74\%} & \textbf{-0.99\%} & 2.31\% & \textbf{-0.87\%} & 2.27s\\
\bottomrule
\end{tabularx}
\end{table*}

\subsection{Beta Distribution}
\label{app:beta}

\begin{figure}[t]
    \centering
    \includegraphics[width=\textwidth]{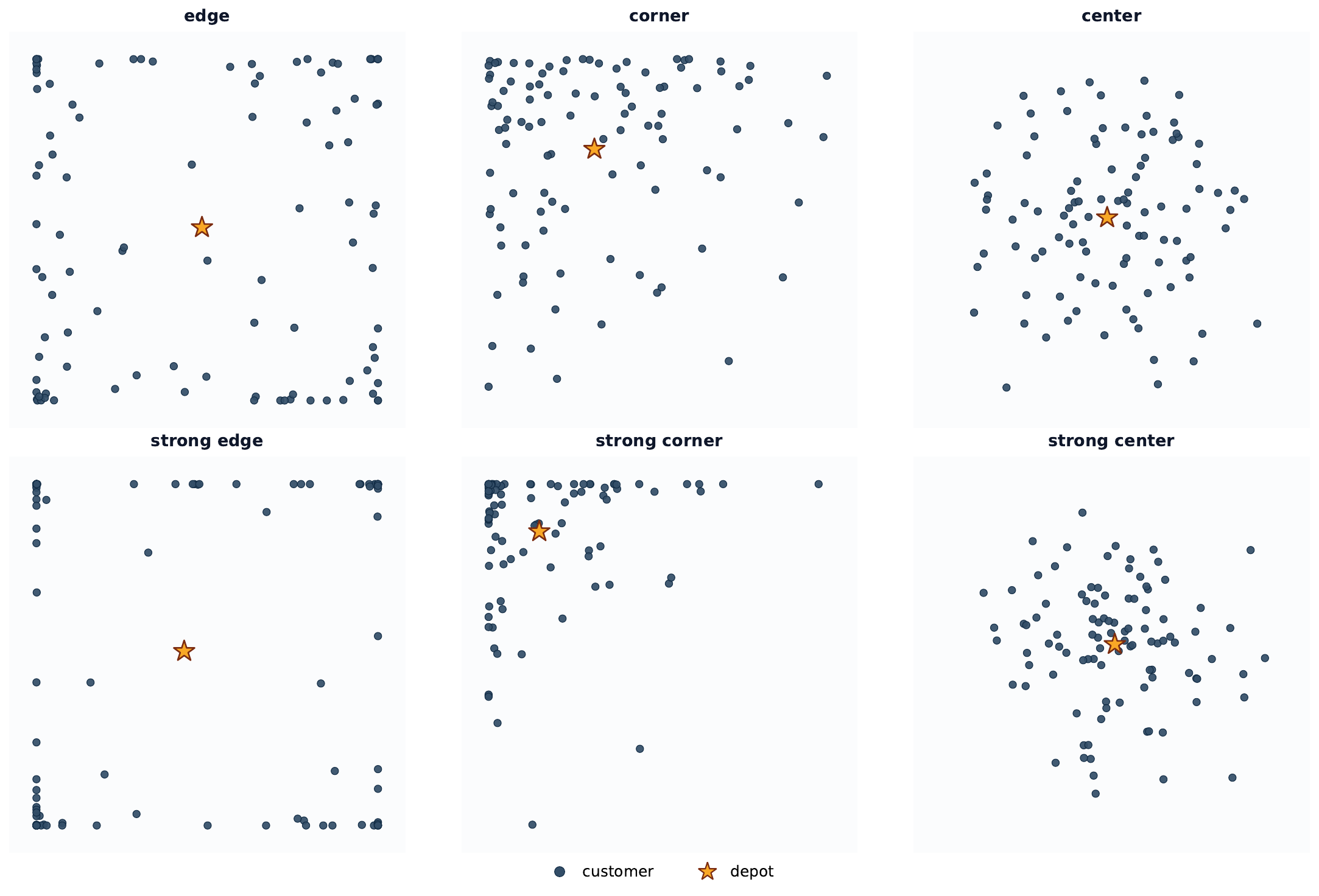}
    \caption{Visualization of the six Beta distribution variants on CVRP100 instances. The orange star marks the depot. Top row: standard intensity; bottom row: strong intensity.}
    \label{fig:beta_variants}
\end{figure}

The Beta distribution family generates node coordinates by independently sampling each axis from a Beta distribution $\text{Beta}(\alpha, \beta)$, with shape parameters tuned to concentrate density near the boundaries, the corners, or the center of the unit square---capturing real-world scenarios where customers are systematically biased toward city peripheries, specific regional sectors, or downtown cores. We construct six variants along two orthogonal axes: \emph{spatial bias type} (edge, corner, center) and \emph{bias intensity} (normal, strong). Figure~\ref{fig:beta_variants} visualizes one representative instance from each variant.

\textbf{Edge.}
A U-shaped marginal density that concentrates customers along the four edges of the unit square, modeling peripheral routes such as ring roads or coastal lines.

\textbf{Strong edge.}
A more extreme version of \emph{Edge}, with most nodes confined to a narrow band along the boundaries.

\textbf{Corner.}
Asymmetric Beta parameters push density toward one specific corner of the unit square, modeling deliveries localized to a specific geographic sector relative to the depot.

\textbf{Strong corner.}
A tighter concentration in the target corner, with only sparse outliers elsewhere.

\textbf{Center.}
A bell-shaped marginal density concentrating customers around the unit-square center, modeling downtown-style routing in a central business district.

\textbf{Strong center.}
Sharper concentration around the center, with very few outliers along the periphery.

\begin{table*}[t]
\centering
\fontsize{6.5pt}{7.5pt}\selectfont
\caption{Generalization performance on CVRP100 across different Beta distributions. Gap (\%) is computed against LKH3. Best results among neural methods are in \textbf{bold}; second best are \underline{underlined}.}
\label{tab:beta_full}
\renewcommand{\arraystretch}{0.9}
\setlength{\tabcolsep}{2.5pt}
\begin{tabularx}{\textwidth}{l|CCCCCC|C|C}
\toprule
\rowcolor{headerbg}
& \multicolumn{6}{c|}{\textbf{Beta distribution variants}} & & \\
\rowcolor{headerbg}
\textbf{Method} & Edge & Corner & Center & Strong edge & Strong corner & Strong center & \textbf{Average Gap} & \textbf{Time} \\
\midrule
\rowcolor{classicalbg} LKH3 & --- & --- & --- & --- & --- & --- & --- & 10.90s \\
\midrule
\multicolumn{9}{l}{\textit{Single-distribution training}} \\
\rowcolor{singlebg} ELG    & 1.21\% & 2.05\% & 1.87\% & \underline{1.28\%} & 5.59\% & 2.28\% & 2.10\% & 4.18s\\
\rowcolor{singlebg} LEHD   & 2.85\% & 3.17\% & 3.06\% & 5.94\% & 4.52\% & 3.45\% & 3.92\% & 4.20s\\
\rowcolor{singlebg} BQ     & 5.04\% & 3.25\% & 2.38\% & 8.87\% & 4.09\% & \underline{2.33\%} & 4.83\% & 2.30m\\
\rowcolor{singlebg} INViT  & 9.70\% & 11.46\% & 9.34\% & 10.44\% & 15.26\% & 9.78\% & 10.76\% & 15.2m\\
\midrule
\multicolumn{9}{l}{\textit{Multi-distribution training}} \\
\rowcolor{multibg} AMDKD-AM     & 3.82\% & 5.50\% & 4.70\% & 4.27\% & 9.47\% & 5.84\% & 5.22\% & 1.06s\\
\rowcolor{multibg} AMDKD-POMO   & 1.38\% & \underline{1.64\%} & 1.99\% & 2.14\% & \textbf{1.43\%} & 2.53\% & \underline{1.83\%} & 1.05s\\
\rowcolor{multibg} LEHD-VCT+MEM & 2.14\% & 2.93\% & 2.17\% & 6.64\% & 4.69\% & 2.61\% & 3.68\% & 8.13s\\
\midrule
\rowcolor{ourbg} POMO              & \underline{0.81\%} & 1.81\% & \underline{1.42\%} & 2.60\% & 12.29\% & 1.92\% & 2.95\% & 1.98s\\
\rowcolor{ourbg} \textbf{MixRoute} & \textbf{0.49\%} & \textbf{1.60\%} & \textbf{1.16\%} & \textbf{0.33\%} & \underline{2.74\%} & \textbf{1.44\%} & \textbf{1.11\%} & 2.23s\\
\bottomrule
\end{tabularx}
\end{table*}

\textbf{Results.} Table~\ref{tab:beta_full} reports the per-variant performance on the Beta family. \textsc{MixRoute} achieves the lowest gap on five of the six variants and an average gap of $1.11\%$, outperforming the second-best baseline (AMDKD-POMO, $1.83\%$). The exception is \emph{Strong corner}, where AMDKD-POMO ($1.43\%$) edges out \textsc{MixRoute} ($2.74\%$); we attribute this to AMDKD-POMO's exposure to corner-biased distributions during multi-distribution training.

\subsection{Cluster Distribution}
\label{app:cluster}

\begin{figure}[t]
    \centering
    \includegraphics[width=\textwidth]{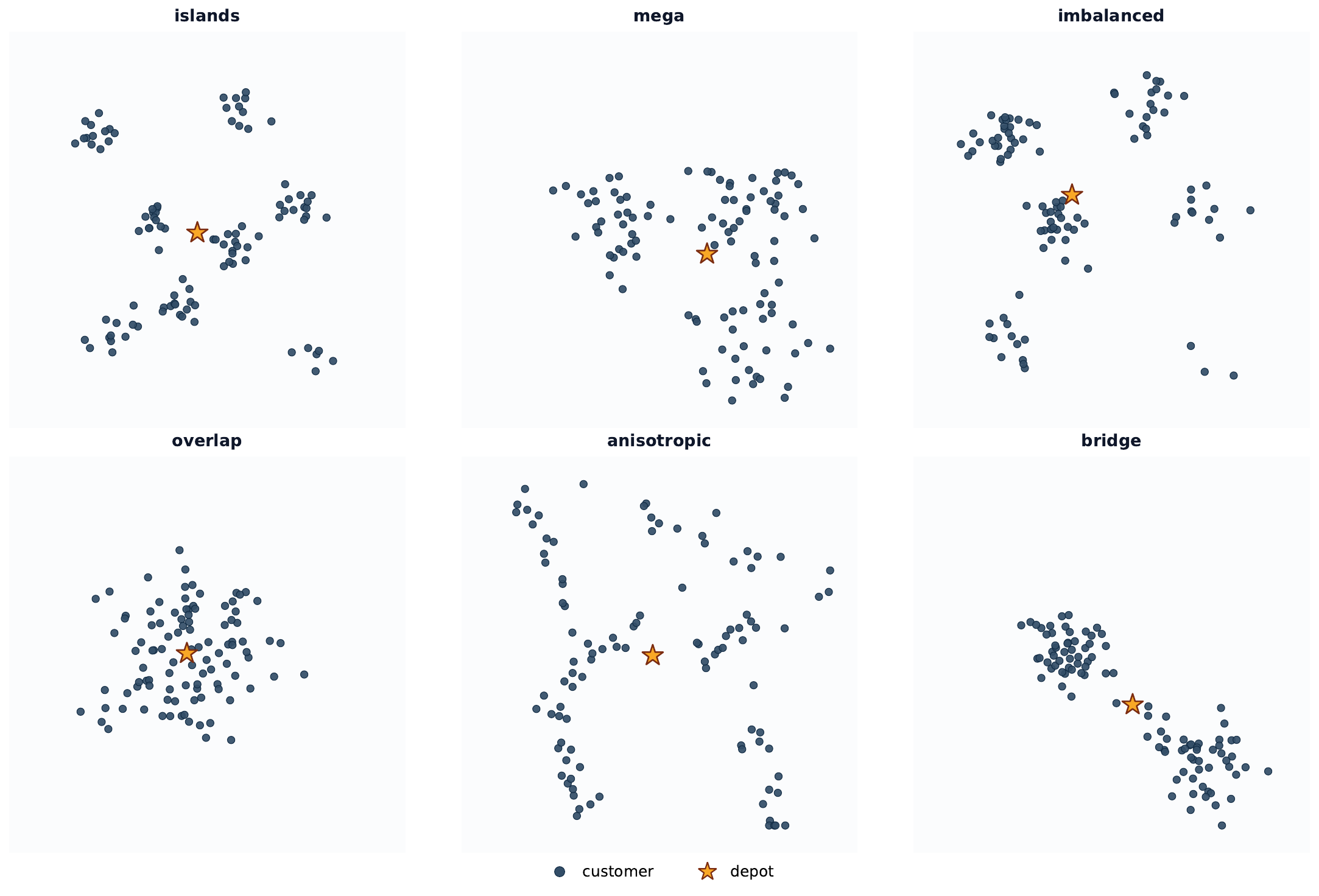}
    \caption{Visualization of the six Cluster distribution variants on CVRP100 instances. The orange star marks the depot. Top row: \emph{Islands}, \emph{Mega}, \emph{Imbalanced}; bottom row: \emph{Overlap}, \emph{Anisotropic}, \emph{Bridge}.}
    \label{fig:cluster_variants}
\end{figure}

The Cluster distribution family generates node coordinates from a mixture of Gaussian components, modeling routing scenarios where customers form spatial groupings such as residential neighborhoods, commercial districts, or industrial parks. We construct six variants that vary along distinct cluster-level dimensions: the number, size, balance, overlap, shape, and connectivity of clusters. Figure~\ref{fig:cluster_variants} visualizes one representative instance from each variant.

\textbf{Islands.}
A balanced multi-cluster pattern with $K=8$ tight clusters of equal population ($\sigma = 0.035$), resembling an archipelago of well-separated dense pockets.

\textbf{Mega.}
A few large, loose clusters ($K=3$, $\sigma = 0.09$) with diffuse edges, contrasting sharply with \emph{Islands} (few-and-large vs.\ many-and-small).

\textbf{Imbalanced.}
$K=7$ centers with weights drawn from $\text{Dirichlet}(\alpha=0.25)$, producing highly uneven populations---typically one dominant cluster contains $60$--$80\%$ of all customers.

\textbf{Overlap.}
$K=9$ centers with large noise ($\sigma = 0.12$) contracted toward the unit-square center, forcing nine latent clusters to entangle into a near-uniform point cloud.

\textbf{Anisotropic.}
$K=6$ elongated strands with anisotropic shape ($\sigma_x = 0.025$, $\sigma_y = 0.14$), each randomly rotated, modeling directionally biased clusters along varying axes.

\textbf{Bridge.}
$K=2$ well-separated dense clusters with $12\%$ of customers placed along the line connecting them, modeling bi-modal urban-rural patterns where customers thin out along a connecting highway.

\textbf{Results.} Table~\ref{tab:cluster_full} reports the per-variant performance on the Cluster family. \textsc{MixRoute} achieves the lowest gap on five of the six variants and an average gap of $1.82\%$, outperforming the second-best baseline (AMDKD-POMO, $2.37\%$). The exception is \emph{Imbalanced}, where AMDKD-POMO ($1.73\%$) edges out \textsc{MixRoute} ($2.44\%$); we attribute this to AMDKD-POMO's exposure to skewed cluster patterns during multi-distribution training.

\begin{table*}[t]
\centering
\fontsize{6.5pt}{7.5pt}\selectfont
\caption{Generalization performance on CVRP100 across different Cluster distributions. Gap (\%) is computed against LKH3. Best results among neural methods are in \textbf{bold}; second best are \underline{underlined}.}
\label{tab:cluster_full}
\renewcommand{\arraystretch}{0.9}
\setlength{\tabcolsep}{2.5pt}
\begin{tabularx}{\textwidth}{l|CCCCCC|C|C}
\toprule
\rowcolor{headerbg}
& \multicolumn{6}{c|}{\textbf{Cluster distribution variants}} & & \\
\rowcolor{headerbg}
\textbf{Method} & Islands & Mega & Imbalanced & Overlap & Anisotropic & Bridge & \textbf{Average Gap} & \textbf{Time} \\
\midrule
\rowcolor{classicalbg} LKH3 & --- & --- & --- & --- & --- & --- & --- & 13.56s \\
\midrule
\multicolumn{9}{l}{\textit{Single-distribution training}} \\
\rowcolor{singlebg} ELG    & 2.58\% & 2.89\% & 6.42\% & 3.38\% & 2.29\% & 4.59\% & 3.52\% & 4.35s\\
\rowcolor{singlebg} LEHD   & 3.59\% & 3.53\% & 4.55\% & 3.41\% & 3.68\% & 5.30\% & 3.96\% & 4.26s\\
\rowcolor{singlebg} BQ     & 3.67\% & 2.30\% & 3.08\% & 2.59\% & 3.20\% & 2.42\% & 2.90\% & 2.30m\\
\rowcolor{singlebg} INViT  & 11.79\% & 9.40\% & 10.24\% & 10.48\% & 11.99\% & 9.24\% & 10.60\% & 15.09m\\
\midrule
\multicolumn{9}{l}{\textit{Multi-distribution training}} \\
\rowcolor{multibg} AMDKD-AM     & 5.74\% & 6.08\% & 9.28\% & 7.77\% & 5.92\% & 7.57\% & 6.88\% & 1.17s\\
\rowcolor{multibg} AMDKD-POMO   & 2.65\% & 2.56\% & \textbf{1.73\%} & \underline{2.43\%} & 2.33\% & \underline{2.40\%} & \underline{2.37\%} & 1.01s\\
\rowcolor{multibg} LEHD-VCT+MEM & 3.01\% & 2.66\% & 3.40\% & 2.65\% & 2.89\% & 4.23\% & 3.11\% & 8.10s\\
\midrule
\rowcolor{ourbg} POMO              & \underline{2.03\%} & \underline{2.19\%} & 7.55\% & 2.71\% & \underline{1.99\%} & 4.35\% & 3.25\% & 2.00s\\
\rowcolor{ourbg} \textbf{MixRoute} & \textbf{1.62\%} & \textbf{1.49\%} & \underline{2.44\%} & \textbf{1.90\%} & \textbf{1.48\%} & \textbf{2.31\%} & \textbf{1.82\%} & 2.27s\\
\bottomrule
\end{tabularx}
\end{table*}

\subsection{Grid Distribution}
\label{app:grid}

\begin{figure}[t]
    \centering
    \includegraphics[width=\textwidth]{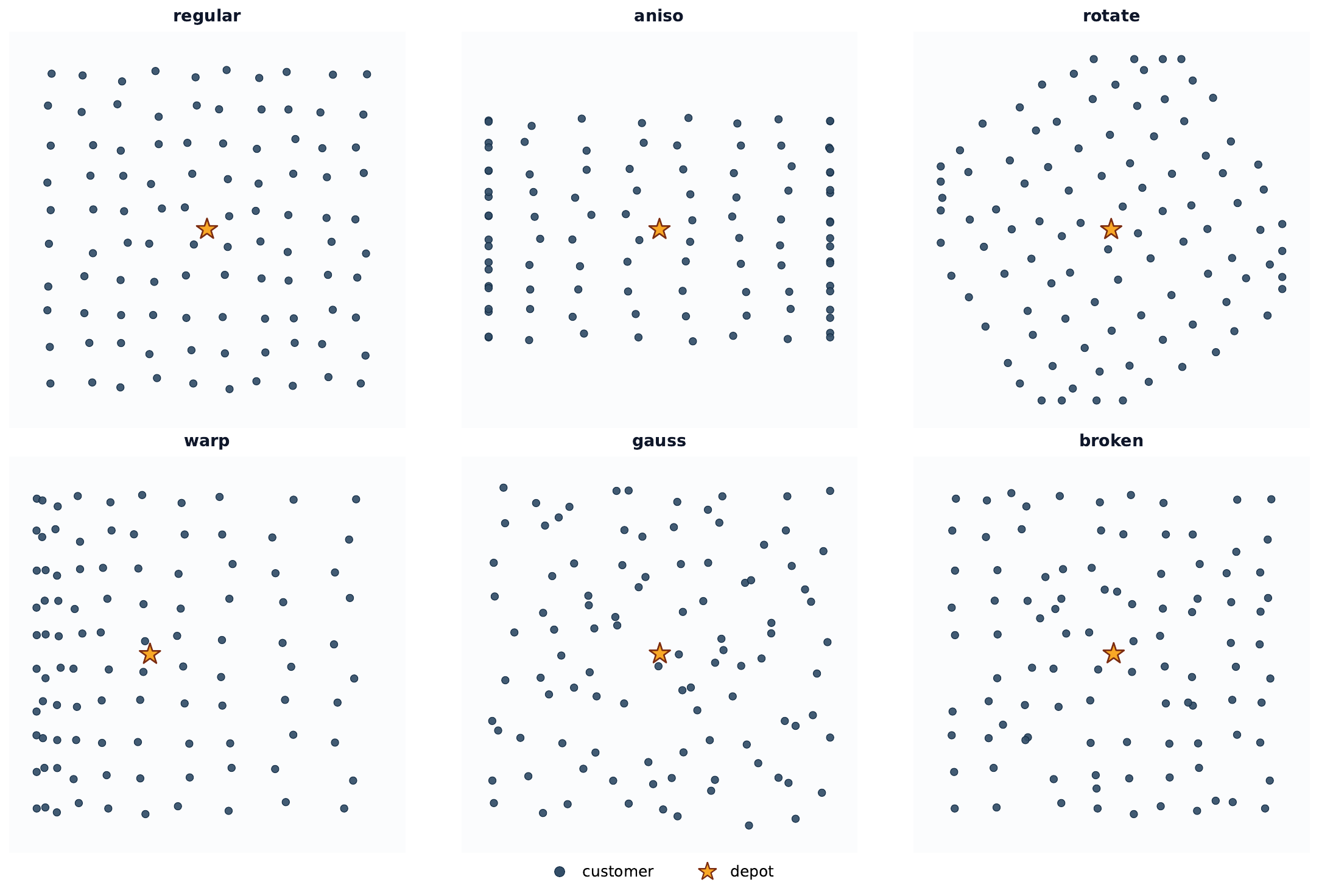}
    \caption{Visualization of the six Grid distribution variants on CVRP100 instances. The orange star marks the depot. \emph{Regular} serves as the structural baseline; the other five variants apply distinct perturbations on top.}
    \label{fig:grid_variants}
\end{figure}

The Grid distribution family generates node coordinates from a regular $10 \times 10$ lattice over the unit square, perturbed by small jitter ($\pm 0.02$). Compared to the random distributions in earlier sections, Grid represents a structurally regular regime that differs sharply from the random training distribution. We construct six variants that share the same underlying lattice but apply distinct geometric or stochastic perturbations. Figure~\ref{fig:grid_variants} visualizes one representative instance from each variant.

\textbf{Regular.}
The base $10 \times 10$ lattice with only default jitter, yielding an almost ideal grid that an ideal NCO solver should exploit via row/column sweeps.

\textbf{Aniso.}
The grid is rescaled around the center with $a_x = 1.5$ and $a_y = 0.7$, producing a horizontally stretched array with spacing ratio $\sim 2.14$, modeling coordinate-scale anisotropy.

\textbf{Rotate.}
The grid is rotated by $\theta \sim \text{Uniform}(0, \pi/2)$ around its center, with edge points clamped back to $[0, 1]^2$. Although VRP solutions are rotation-invariant, models taking absolute coordinates as input may not learn this invariance.

\textbf{Warp.}
The $x$-coordinate is compressed via $x \leftarrow x^2$, densifying the left side and sparsifying the right---reminiscent of a perspective projection.

\textbf{Gauss.}
Gaussian noise ($\sigma = 0.03$) is added on top of the jitter, making the lattice structure nearly invisible while the underlying generative process remains anchored at lattice centers.

\textbf{Broken.}
$15\%$ of lattice points are replaced with uniform samples, simulating semi-structured logistics networks where the solver must exploit grid regularity while accommodating outliers.

\begin{table*}[t]
\centering
\fontsize{6.5pt}{7.5pt}\selectfont
\caption{Generalization performance on CVRP100 across different Grid distributions. Gap (\%) is computed against LKH3. Best results among neural methods are in \textbf{bold}; second best are \underline{underlined}.}
\label{tab:grid_full}
\renewcommand{\arraystretch}{0.9}
\setlength{\tabcolsep}{2.5pt}
\begin{tabularx}{\textwidth}{l|CCCCCC|C|C}
\toprule
\rowcolor{headerbg}
& \multicolumn{6}{c|}{\textbf{Grid distribution variants}} & & \\
\rowcolor{headerbg}
\textbf{Method} & Regular & Aniso & Rotate & Warp & Gauss & Broken & \textbf{Average Gap} & \textbf{Time} \\
\midrule
\rowcolor{classicalbg} LKH3 & --- & --- & --- & --- & --- & --- & --- & 9.64s \\
\midrule
\multicolumn{9}{l}{\textit{Single-distribution training}} \\
\rowcolor{singlebg} ELG    & 2.01\% & \underline{0.89\%} & 1.74\% & 1.70\% & 1.77\% & 1.72\% & 1.65\% & 4.12s\\
\rowcolor{singlebg} LEHD   & 2.64\% & 1.95\% & 2.32\% & 2.86\% & 2.67\% & 3.08\% & 2.60\% & 4.26s\\
\rowcolor{singlebg} BQ     & 2.71\% & 2.26\% & 2.48\% & 2.75\% & 2.62\% & 3.20\% & 2.67\% & 2.30m\\
\rowcolor{singlebg} INViT  & 7.02\% & 7.50\% & 7.35\% & 7.81\% & 7.62\% & 8.31\% & 7.60\% & 15.16m\\
\midrule
\multicolumn{9}{l}{\textit{Multi-distribution training}} \\
\rowcolor{multibg} AMDKD-AM     & 3.92\% & 4.05\% & 4.24\% & 4.29\% & 4.02\% & 4.35\% & 4.14\% & 1.02s\\
\rowcolor{multibg} AMDKD-POMO   & 1.88\% & 1.17\% & 1.69\% & 1.52\% & 1.59\% & 1.98\% & 1.64\% & 1.00s\\
\rowcolor{multibg} LEHD-VCT+MEM & 2.08\% & 1.85\% & 2.19\% & 2.21\% & 2.05\% & 2.36\% & 2.13\% & 8.09s\\
\midrule
\rowcolor{ourbg} POMO              & \textbf{1.43\%} & 1.30\% & \underline{1.40\%} & \underline{1.28\%} & \underline{1.25\%} & \underline{1.64\%} & \underline{1.38\%} & 1.94s\\
\rowcolor{ourbg} \textbf{MixRoute} & \underline{1.45\%} & \textbf{0.49\%} & \textbf{1.36\%} & \textbf{1.25\%} & \textbf{1.05\%} & \textbf{1.42\%} & \textbf{1.18\%} & 2.25s\\
\bottomrule
\end{tabularx}
\end{table*}

\textbf{Results.} Table~\ref{tab:grid_full} reports the per-variant performance on the Grid family. \textsc{MixRoute} achieves the lowest gap on five of the six variants and an average gap of $1.18\%$, outperforming the second-best baseline (POMO, $1.38\%$). The exception is \emph{Regular}, where POMO ($1.43\%$) marginally edges out \textsc{MixRoute} ($1.45\%$); we view this as evidence that \textsc{MixRoute} does not sacrifice in-structure performance to gain robustness, since the two methods remain effectively tied on the structural baseline while \textsc{MixRoute} dominates across all five perturbed variants.

\subsection{Stripe Distribution}
\label{app:stripe}

\begin{figure}[t]
    \centering
    \includegraphics[width=\textwidth]{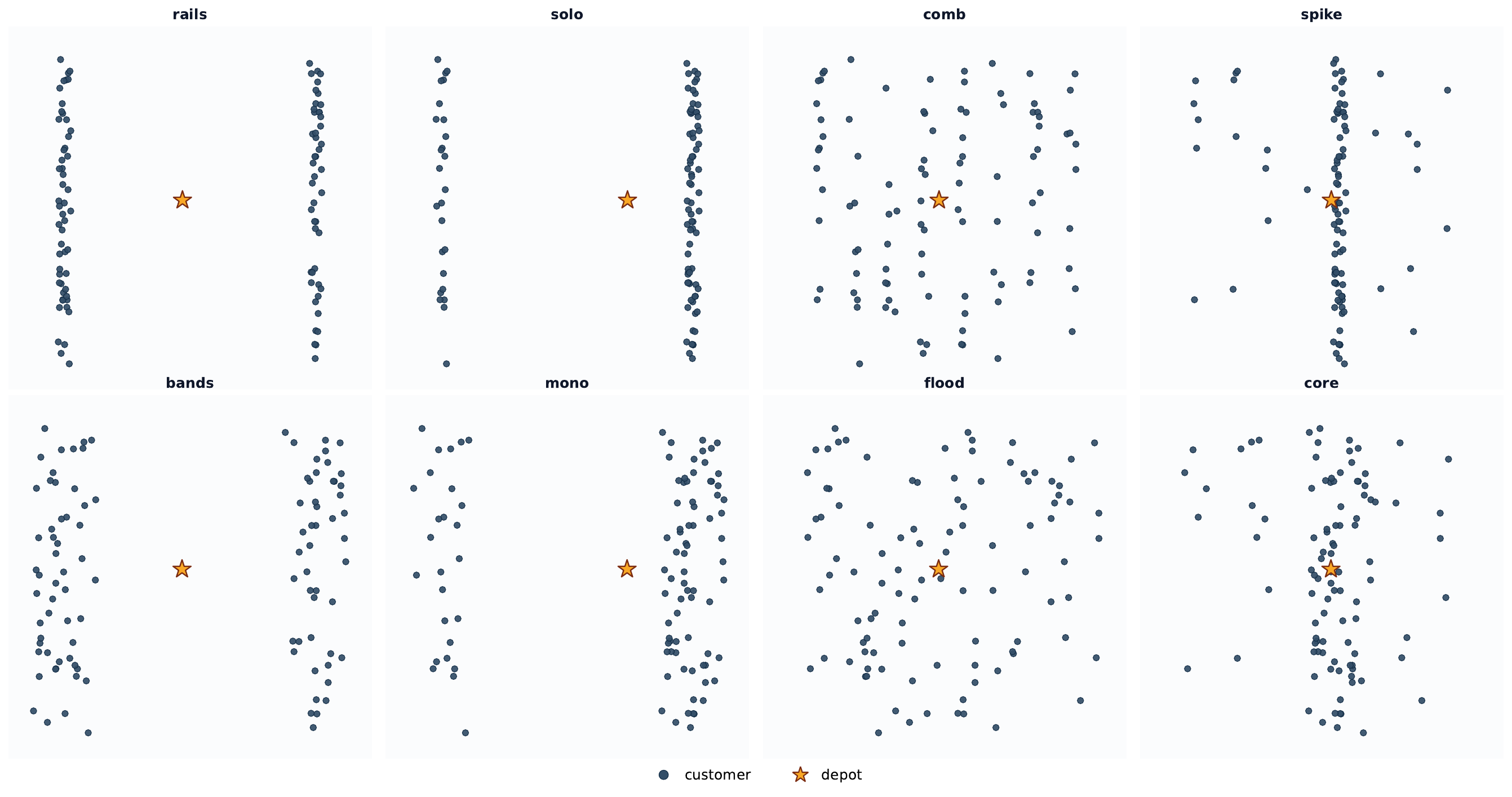}
    \caption{Visualization of the eight Stripe distribution variants on CVRP100 instances. The orange star marks the depot. The eight variants correspond to a $2 \times 2 \times 2$ design over band width, number of bands, and population balance.}
    \label{fig:stripe_variants}
\end{figure}

The Stripe distribution family generates node coordinates by sampling the $x$-coordinate from one of $K$ vertical bands of half-width $w$ centered at fixed positions across $[0.1, 0.9]$, with the $y$-coordinate drawn uniformly from $[0, 1]$. This family captures routing scenarios where customers are organized along structured longitudinal corridors---such as customers along major roads, parallel commercial avenues, or multi-lane delivery routes. We construct eight variants spanning a $2 \times 2 \times 2$ design: \emph{band width} (narrow $w=0.02$ vs.\ wide $w=0.10$), \emph{number of bands} ($K=2$ vs.\ $K=8$), and \emph{population balance} (uniform vs.\ biased). Figure~\ref{fig:stripe_variants} visualizes one representative instance from each variant.

\textbf{Rails.} \emph{(Narrow, $K=2$, uniform.)}
Two narrow vertical lines at $x \in \{0.1, 0.9\}$ each receiving roughly $50$ customers, leaving the central region empty.

\textbf{Solo.} \emph{(Narrow, $K=2$, biased.)}
The same two narrow lines as \emph{Rails}, but with population weights $(0.25, 0.75)$, stressing load balancing across asymmetric corridors.

\textbf{Comb.} \emph{(Narrow, $K=8$, uniform.)}
Eight narrow vertical lines evenly spaced across $[0.1, 0.9]$, each containing $\sim 12$--$13$ customers, resembling the teeth of a comb.

\textbf{Spike.} \emph{(Narrow, $K=8$, biased.)}
A comb-like layout where the central band absorbs $75\%$ of the population, creating a sharp ``spike'' protruding from an otherwise sparse comb.

\textbf{Bands.} \emph{(Wide, $K=2$, uniform.)}
The wide-band counterpart of \emph{Rails}: two wide bands ($w=0.10$) covering roughly $20\%$ of the $x$-range each, with noticeably more diffuse edges.

\textbf{Mono.} \emph{(Wide, $K=2$, biased.)}
The wide-band counterpart of \emph{Solo}: two bands with population weights $(0.25, 0.75)$, testing dominant-corridor identification under blurred boundaries.

\textbf{Flood.} \emph{(Wide, $K=8$, uniform.)}
Eight wide bands that heavily overlap due to $w$ exceeding half the inter-center spacing, producing a ``pseudo-uniform'' marginal whose surface appearance masks an underlying eight-corridor structure.

\textbf{Core.} \emph{(Wide, $K=8$, biased.)}
The wide-band counterpart of \emph{Spike}: eight wide bands with the central band carrying $75\%$ of the population, mirroring main-road plus side-street logistics patterns.

\begin{table*}[t]
\centering
\fontsize{6.5pt}{7.5pt}\selectfont
\caption{Generalization performance on CVRP100 across different Stripe distributions. Gap (\%) is computed against LKH3. Best results among neural methods are in \textbf{bold}; second best are \underline{underlined}.}
\label{tab:stripe_full}
\renewcommand{\arraystretch}{0.9}
\setlength{\tabcolsep}{2.5pt}
\begin{tabularx}{\textwidth}{l|CCCCCCCC|C|C}
\toprule
\rowcolor{headerbg}
& \multicolumn{8}{c|}{\textbf{Stripe distribution variants}} & & \\
\rowcolor{headerbg}
\textbf{Method} & Rails & Solo & Comb & Spike & Bands & Mono & Flood & Core & \textbf{Average Gap} & \textbf{Time} \\
\midrule
\rowcolor{classicalbg} LKH3 & --- & --- & --- & --- & --- & --- & --- & --- & --- & 16.39s \\
\midrule
\multicolumn{11}{l}{\textit{Single-distribution training}} \\
\rowcolor{singlebg} ELG    & 1.58\% & 2.48\% & 1.53\% & 3.40\% & 1.19\% & 2.18\% & 1.80\% & 2.60\% & 2.02\% & 5.57s\\
\rowcolor{singlebg} LEHD   & 1.98\% & 3.37\% & 2.86\% & 4.83\% & 1.44\% & 3.21\% & 2.94\% & 3.59\% & 2.92\% & 5.73\\
\rowcolor{singlebg} BQ     & 3.61\% & 4.93\% & 2.66\% & \textbf{2.03\%} & 1.92\% & 3.92\% & 3.03\% & \underline{1.92\%} & 3.02\% & 3.21m\\
\rowcolor{singlebg} INViT  & 11.64\% & 11.69\% & 9.31\% & 13.61\% & 8.11\% & 10.13\% & 8.93\% & 11.53\% & 10.45\% & 20.15m\\
\midrule
\multicolumn{11}{l}{\textit{Multi-distribution training}} \\
\rowcolor{multibg} AMDKD-AM     & 3.33\% & 3.80\% & 4.76\% & 9.33\% & 3.30\% & 4.16\% & 4.69\% & 7.62\% & 4.91\% & 1.42s\\
\rowcolor{multibg} AMDKD-POMO   & 1.69\% & \underline{1.95\%} & 1.57\% & 3.73\% & \underline{0.52\%} & \textbf{1.35\%} & 1.50\% & 2.65\% & \underline{1.77\%} & 1.44s\\
\rowcolor{multibg} LEHD-VCT+MEM & \underline{1.30\%} & 3.66\% & 2.53\% & 4.44\% & 0.97\% & 2.86\% & 2.59\% & 3.08\% & 2.56\% & 10.79s\\
\midrule
\rowcolor{ourbg} POMO              & 7.11\% & 50.55\% & \underline{1.20\%} & 4.01\% & 0.63\% & 2.02\% & \underline{1.36\%} & 2.21\% & 8.30\% & 2.66s\\
\rowcolor{ourbg} \textbf{MixRoute} & \textbf{0.21\%} & \textbf{1.57\%} & \textbf{1.00\%} & \underline{2.13\%} & \textbf{0.39\%} & \underline{1.46\%} & \textbf{1.13\%} & \textbf{1.68\%} & \textbf{1.13\%} & 3.04s\\
\bottomrule
\end{tabularx}
\end{table*}

\textbf{Results.} Table~\ref{tab:stripe_full} reports the per-variant performance on the Stripe family. \textsc{MixRoute} achieves the lowest gap on six of the eight variants and an average gap of $1.13\%$, outperforming the second-best baseline (AMDKD-POMO, $1.77\%$). The most striking case is \emph{Solo}, where POMO collapses to $50.55\%$---the largest failure in our benchmark---while \textsc{MixRoute} attains $1.57\%$ under identical training conditions, a $49$-point improvement that highlights how severely a fixed normalization can overfit to a single density assumption.

\subsection{Gaussian Distribution}
\label{app:gaussian}

\begin{figure}[t]
    \centering
    \includegraphics[width=\textwidth]{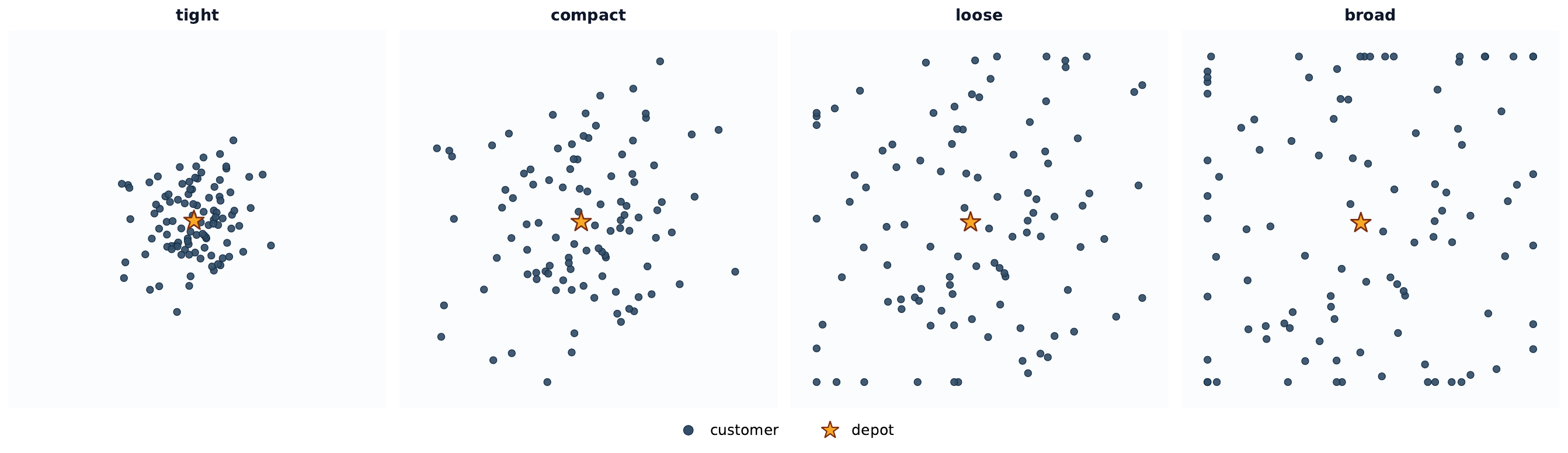}
    \caption{Visualization of the four Gaussian distribution variants on CVRP100 instances. The orange star marks the depot. From left to right, $\sigma$ increases from $0.1$ to $0.4$, traversing a spectrum from highly concentrated clusters to near-uniform with mild boundary clipping.}
    \label{fig:gaussian_variants}
\end{figure}

The Gaussian distribution family generates node coordinates by sampling each axis independently from a Gaussian distribution centered at $(0.5, 0.5)$ with standard deviation $\sigma$, then clipping to the unit square. The Gaussian family varies along a single continuous axis---the standard deviation $\sigma$---which controls how concentrated the customer population is around the depot, capturing routing scenarios at varying spatial scales: tightly localized urban deliveries, mid-sized city logistics, metropolitan networks, and country-scale distribution. We construct four variants with increasing $\sigma$. Figure~\ref{fig:gaussian_variants} visualizes one representative instance from each variant.

\textbf{Tight ($\sigma = 0.1$).}
A compact blob occupying roughly $30$--$40\%$ of the unit square with virtually no boundary clipping, mirroring urban deliveries within a single commercial district.

\textbf{Compact ($\sigma = 0.2$).}
A moderate cluster spanning $\pm 0.2$ around the center with mild boundary clipping, modeling mid-sized cities with a dense business district and sparser suburbs. This $\sigma$ matches the most common Gaussian setting in prior NCO literature.

\textbf{Loose ($\sigma = 0.3$).}
A broad cluster with the $3\sigma$ region nearly filling the unit square; the central density peak remains identifiable while the periphery approaches uniform-like density---a metropolitan-scale pattern with both dense urban core and non-negligible suburban customers.

\textbf{Broad ($\sigma = 0.4$).}
A near-uniform appearance with two latent signatures: a mildly elevated central peak, and ``walls'' of customers along the unit-square boundary due to substantial clipping. This serves as a ``pseudo-uniform'' diagnostic for whether models genuinely handle uniform structure or have overfit to a strict i.i.d.\ assumption.

\begin{table*}[t]
\centering
\fontsize{6.5pt}{7.5pt}\selectfont
\caption{Generalization performance on CVRP100 across different Gaussian distributions. Gap (\%) is computed against LKH3. Best results among neural methods are in \textbf{bold}; second best are \underline{underlined}.}
\label{tab:gaussian_full}
\renewcommand{\arraystretch}{0.9}
\setlength{\tabcolsep}{2.5pt}
\begin{tabularx}{\textwidth}{l|CCCC|C|C}
\toprule
\rowcolor{headerbg}
& \multicolumn{4}{c|}{\textbf{Gaussian distribution variants}} & & \\
\rowcolor{headerbg}
\textbf{Method} & Tight ($\sigma=0.1$) & Compact ($\sigma=0.2$) & Loose ($\sigma=0.3$) & Broad ($\sigma=0.4$) & \textbf{Average Gap} & \textbf{Time} \\
\midrule
\rowcolor{classicalbg} LKH3 & --- & --- & --- & --- & --- & 7.73s \\
\midrule
\multicolumn{7}{l}{\textit{Single-distribution training}} \\
\rowcolor{singlebg} ELG    & 4.35\% & 2.29\% & 1.80\% & 1.61\% & 2.16\% & 2.74s\\
\rowcolor{singlebg} LEHD   & 4.39\% & 3.07\% & 3.20\% & 2.84\% & 3.19\% & 2.80s\\
\rowcolor{singlebg} BQ     & 3.55\% & 2.56\% & 2.87\% & 2.98\% & 2.91\% & 1.40m\\
\rowcolor{singlebg} INViT  & 9.90\% & 9.78\% & 9.64\% & 9.77\% & 9.75\% & 10.14m\\
\midrule
\multicolumn{7}{l}{\textit{Multi-distribution training}} \\
\rowcolor{multibg} AMDKD-AM     & 12.00\% & 5.91\% & 5.10\% & 4.65\% & 5.97\% & 0.73s\\
\rowcolor{multibg} AMDKD-POMO   & \textbf{2.25\%} & 2.32\% & 1.62\% & 1.42\% & 1.79\% & 0.72s\\
\rowcolor{multibg} LEHD-VCT+MEM & 3.37\% & 2.58\% & 2.67\% & 2.72\% & 2.75\% & 5.41s\\
\midrule
\rowcolor{ourbg} POMO              & 3.86\% & \underline{1.87\%} & \underline{1.38\%} & \underline{1.16\%} & \underline{1.72\%} & 1.31s\\
\rowcolor{ourbg} \textbf{MixRoute} & \underline{2.29\%} & \textbf{1.66\%} & \textbf{1.22\%} & \textbf{0.93\%} & \textbf{1.36\%} & 1.50s\\
\bottomrule
\end{tabularx}
\end{table*}

\textbf{Results.} Table~\ref{tab:gaussian_full} reports the per-variant performance on the Gaussian family. \textsc{MixRoute} achieves the lowest gap on three of the four variants and an average gap of $1.36\%$, outperforming the second-best baseline (POMO, $1.72\%$). The exception is \emph{Tight} ($\sigma=0.1$), where AMDKD-POMO ($2.25\%$) marginally edges out \textsc{MixRoute} ($2.29\%$); we attribute this to AMDKD-POMO's exposure to highly concentrated cluster patterns during multi-distribution training.

\subsection{Expansion Distribution}
\label{app:expansion}

\begin{figure}[t]
    \centering
    \includegraphics[width=\textwidth]{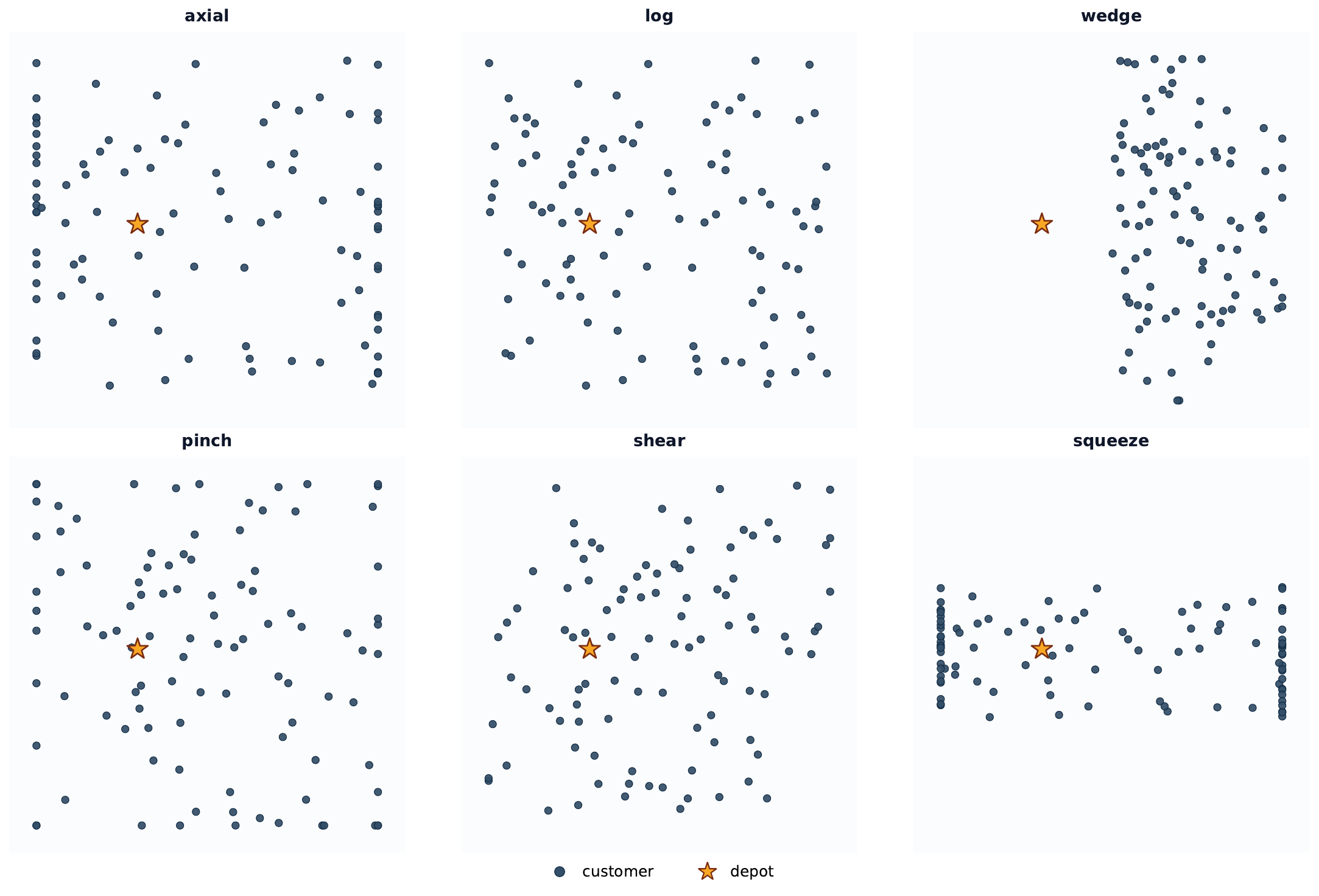}
    \caption{Visualization of the six Expansion distribution variants on CVRP100 instances. The orange star marks the depot. Each variant applies a distinct spatial deformation to a uniformly sampled point set.}
    \label{fig:expansion_variants}
\end{figure}

The Expansion distribution family generates node coordinates by first sampling uniformly from $[0, 1]^2$ and then applying a spatial deformation that distorts the geometric layout while preserving the unit-square support. Unlike density-shift families, Expansion varies the underlying \emph{coordinate geometry} itself, producing instances whose marginal density may remain near-uniform but whose spatial structure deviates substantially. We construct six variants, each corresponding to a distinct mode of real-world urban geometry. Figure~\ref{fig:expansion_variants} visualizes one representative instance from each variant.

\textbf{Axial.}
Horizontally stretched by a factor of $2$ with edge clipping, yielding an elongated configuration with sparse interior, mirroring coastal cities like Chile or Norway compressed along a major road or river.

\textbf{Log.}
The $x$-coordinate is nonlinearly stretched via a logarithmic-style mapping, producing a denser concentration around the center, corresponding to the classical \emph{monocentric city} pattern with a dominant CBD.

\textbf{Wedge.}
The angular coordinate is folded so that all customers lie on one side of the depot, capturing single-sided service regions bounded by a natural barrier such as a coastline, mountain range, or international border.

\textbf{Pinch.}
A radial outward push produces a center-evacuated configuration with customers accumulating near the corners, modeling \emph{donut-shaped urban patterns} where the depot is centrally placed but customers cluster in suburban belts.

\textbf{Shear.}
An affine shear deforms the square into a parallelogram while preserving local density, capturing non-orthogonal road networks such as Manhattan's $29^\circ$-tilted grid.

\textbf{Squeeze.}
Horizontally stretched by $2.5\times$ and vertically compressed by $0.4\times$, yielding a $6{:}1$ aspect-ratio strip, modeling linear corridor cities along major transit infrastructure such as the Tokaido Shinkansen or California's Highway 1.

\begin{table*}[t]
\centering
\fontsize{6.5pt}{7.5pt}\selectfont
\caption{Generalization performance on CVRP100 across different Expansion distributions. Gap (\%) is computed against LKH3. Best results among neural methods are in \textbf{bold}; second best are \underline{underlined}.}
\label{tab:expansion_full}
\renewcommand{\arraystretch}{0.9}
\setlength{\tabcolsep}{2.5pt}
\begin{tabularx}{\textwidth}{l|CCCCCC|C|C}
\toprule
\rowcolor{headerbg}
& \multicolumn{6}{c|}{\textbf{Expansion distribution variants}} & & \\
\rowcolor{headerbg}
\textbf{Method} & Axial & Log & Wedge & Pinch & Shear & Squeeze & \textbf{Average Gap} & \textbf{Time} \\
\midrule
\rowcolor{classicalbg} LKH3 & --- & --- & --- & --- & --- & --- & --- & 23.14s \\
\midrule
\multicolumn{9}{l}{\textit{Single-distribution training}} \\
\rowcolor{singlebg} ELG    & 0.59\% & 0.16\% & 1.35\% & -0.22\% & -0.28\% & 4.10\% & 0.85\% & 4.17s\\
\rowcolor{singlebg} LEHD   & 0.47\% & 0.14\% & 0.29\% & 0.02\% & \underline{-0.57\%} & 2.67\% & 0.45\% & 4.25s\\
\rowcolor{singlebg} BQ     & 2.05\% & 1.02\% & 0.91\% & 0.64\% & 0.11\% & 3.74\% & 1.36\% & 2.30m\\
\rowcolor{singlebg} INViT  & 6.86\% & 6.46\% & 5.88\% & 6.67\% & 5.86\% & 8.67\% & 6.70\% & 15.12m\\
\midrule
\multicolumn{9}{l}{\textit{Multi-distribution training}} \\
\rowcolor{multibg} AMDKD-AM     & 3.74\% & 2.75\% & 3.55\% & 2.46\% & 2.45\% & 5.47\% & 3.34\% & 1.13s\\
\rowcolor{multibg} AMDKD-POMO   & 0.37\% & -0.08\% & \textbf{-0.08\%} & -0.31\% & -0.52\% & \underline{1.61\%} & \underline{0.13\%} & 1.00s\\
\rowcolor{multibg} LEHD-VCT+MEM & 2.73\% & 1.43\% & 1.27\% & 1.65\% & 0.82\% & 3.61\% & 1.89\% & 8.12s\\
\midrule
\rowcolor{ourbg} POMO              & \underline{0.12\%} & \underline{-0.23\%} & 2.83\% & \underline{-0.54\%} & -0.56\% & 2.88\% & 0.65\% & 1.99s\\
\rowcolor{ourbg} \textbf{MixRoute} & \textbf{-0.10\%} & \textbf{-0.40\%} & \underline{-0.01\%} & \textbf{-0.75\%} & \textbf{-0.86\%} & \textbf{1.60\%} & \textbf{-0.13\%} & 2.25s\\
\bottomrule
\end{tabularx}
\end{table*}

\textbf{Results.} Table~\ref{tab:expansion_full} reports the per-variant performance on the Expansion family. \textsc{MixRoute} achieves the lowest gap on five of the six variants and an average gap of $-0.13\%$, surpassing LKH3 in aggregate and outperforming the second-best baseline (AMDKD-POMO, $0.13\%$). The exception is \emph{Wedge}, where AMDKD-POMO ($-0.08\%$) marginally edges out \textsc{MixRoute} ($-0.01\%$); we attribute this to AMDKD-POMO's exposure to one-sided spatial patterns during multi-distribution training.

\subsection{Implosion Distribution}
\label{app:implosion}

\begin{figure}[t]
    \centering
    \includegraphics[width=\textwidth]{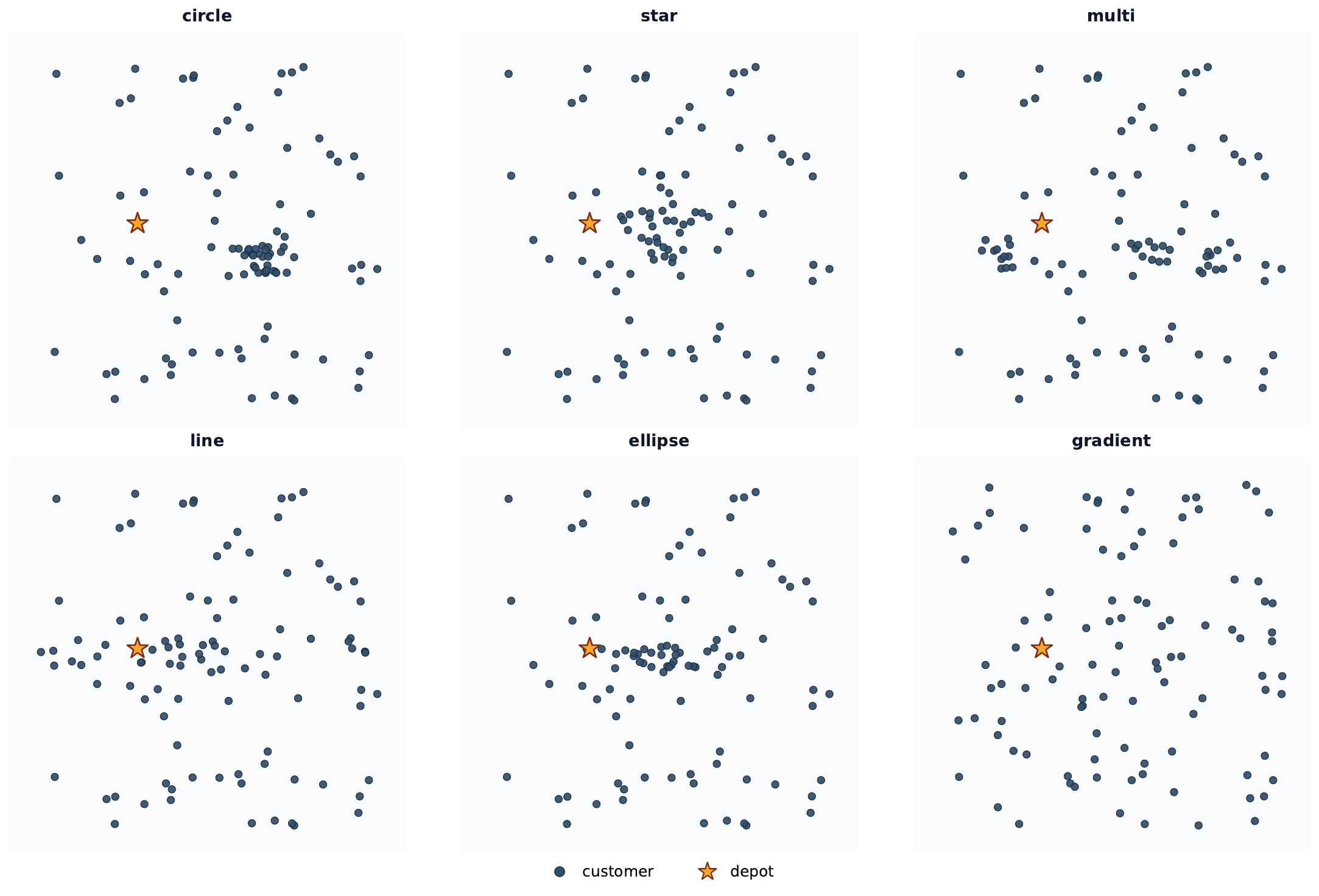}
    \caption{Visualization of the six Implosion distribution variants on CVRP100 instances. The orange star marks the depot.}
    \label{fig:implosion_variants}
\end{figure}

The Implosion distribution family combines a uniform background with a high-density hotspot, with $30\%$ of customers drawn from a structured concentrated region and the rest uniformly distributed. This models partial-concentration patterns---a dense commercial core coexisting with scattered surrounding customers---while varying the hotspot's geometry across six variants. Figure~\ref{fig:implosion_variants} visualizes one representative instance from each variant.

\textbf{Circle.}
A single Gaussian hotspot at a randomly chosen center, modeling a single-center city with a prominent commercial district surrounded by sparser residential customers.

\textbf{Star.}
A non-convex five-pointed star hotspot with sharp protrusions, modeling commercial cores carved into jagged shapes by rivers, railways, or highways.

\textbf{Multi.}
Three non-overlapping hotspots, modeling polycentric metropolitan regions such as Tokyo--Yokohama--Kawasaki or the Pearl River Delta.

\textbf{Line.}
A horizontal narrow band hotspot, modeling linear-corridor patterns where customers concentrate along a subway line, riverside strip, or coastal boulevard.

\textbf{Ellipse.}
An anisotropic Gaussian hotspot ($\sigma_x = 0.10$, $\sigma_y = 0.03$), modeling directional commercial cores elongated along a main avenue or waterfront.

\textbf{Gradient.}
A smooth radial density with no discrete hotspot boundary, modeling continuous monocentric gradients where real-world population densities decay smoothly from city to suburb.

\begin{table*}[t]
\centering
\fontsize{6.5pt}{7.5pt}\selectfont
\caption{Generalization performance on CVRP100 across different Implosion distributions. Gap (\%) is computed against LKH3. Best results among neural methods are in \textbf{bold}; second best are \underline{underlined}. Tied best results are both bolded.}
\label{tab:implosion_full}
\renewcommand{\arraystretch}{0.9}
\setlength{\tabcolsep}{2.5pt}
\begin{tabularx}{\textwidth}{l|CCCCCC|C|C}
\toprule
\rowcolor{headerbg}
& \multicolumn{6}{c|}{\textbf{Implosion distribution variants}} & & \\
\rowcolor{headerbg}
\textbf{Method} & Circle & Star & Multi & Line & Ellipse & Gradient & \textbf{Average Gap} & \textbf{Time} \\
\midrule
\rowcolor{classicalbg} LKH3 & --- & --- & --- & --- & --- & --- & --- & 22.19s \\
\midrule
\multicolumn{9}{l}{\textit{Single-distribution training}} \\
\rowcolor{singlebg} ELG    & 0.46\% & 0.16\% & 0.31\% & 0.40\% & 0.41\% & 0.01\% & 0.29\% & 4.12s\\
\rowcolor{singlebg} LEHD   & 0.25\% & -0.12\% & 0.20\% & 0.38\% & 0.24\% & -0.03\% & 0.15\% & 4.25s\\
\rowcolor{singlebg} BQ     & 0.85\% & 0.40\% & 1.11\% & 0.84\% & 0.74\% & 0.63\% & 0.76\% & 2.30m\\
\rowcolor{singlebg} INViT  & 7.29\% & 7.29\% & 6.92\% & 7.03\% & 8.21\% & 6.24\% & 7.15\% & 15.12m\\
\midrule
\multicolumn{9}{l}{\textit{Multi-distribution training}} \\
\rowcolor{multibg} AMDKD-AM     & 3.48\% & 3.42\% & 3.21\% & 3.14\% & 3.50\% & 2.65\% & 3.23\% & 1.02s\\
\rowcolor{multibg} AMDKD-POMO   & \textbf{-0.22\%} & \textbf{-0.52\%} & -0.09\% & -0.05\% & \textbf{-0.34\%} & -0.30\% & \underline{-0.25\%} & 1.05s\\
\rowcolor{multibg} LEHD-VCT+MEM & 1.36\% & 0.70\% & 1.18\% & 1.07\% & 0.64\% & 0.79\% & 0.96\% & 8.12s\\
\midrule
\rowcolor{ourbg} POMO              & \underline{0.08\%} & -0.16\% & \underline{-0.16\%} & \textbf{-0.14\%} & 0.00\% & \underline{-0.47\%} & -0.15\% & 2.02s\\
\rowcolor{ourbg} \textbf{MixRoute} & \textbf{-0.22\%} & \underline{-0.41\%} & \textbf{-0.24\%} & \underline{-0.13\%} & \underline{-0.16\%} & \textbf{-0.63\%} & \textbf{-0.30\%} & 2.24s\\
\bottomrule
\end{tabularx}
\end{table*}

\textbf{Results.} Table~\ref{tab:implosion_full} reports the per-variant performance on the Implosion family. \textsc{MixRoute} achieves the lowest gap on three variants and ties on \emph{Circle}, with an average gap of $-0.30\%$ that surpasses LKH3 in aggregate and outperforms the second-best baseline (AMDKD-POMO, $-0.25\%$). The exceptions are \emph{Star} and \emph{Ellipse}, where AMDKD-POMO marginally edges out \textsc{MixRoute}; we attribute this to its exposure to anisotropic and irregular hotspots during multi-distribution training.

\subsection{Explosion Distribution}
\label{app:explosion}

\begin{figure}[t]
    \centering
    \includegraphics[width=\textwidth]{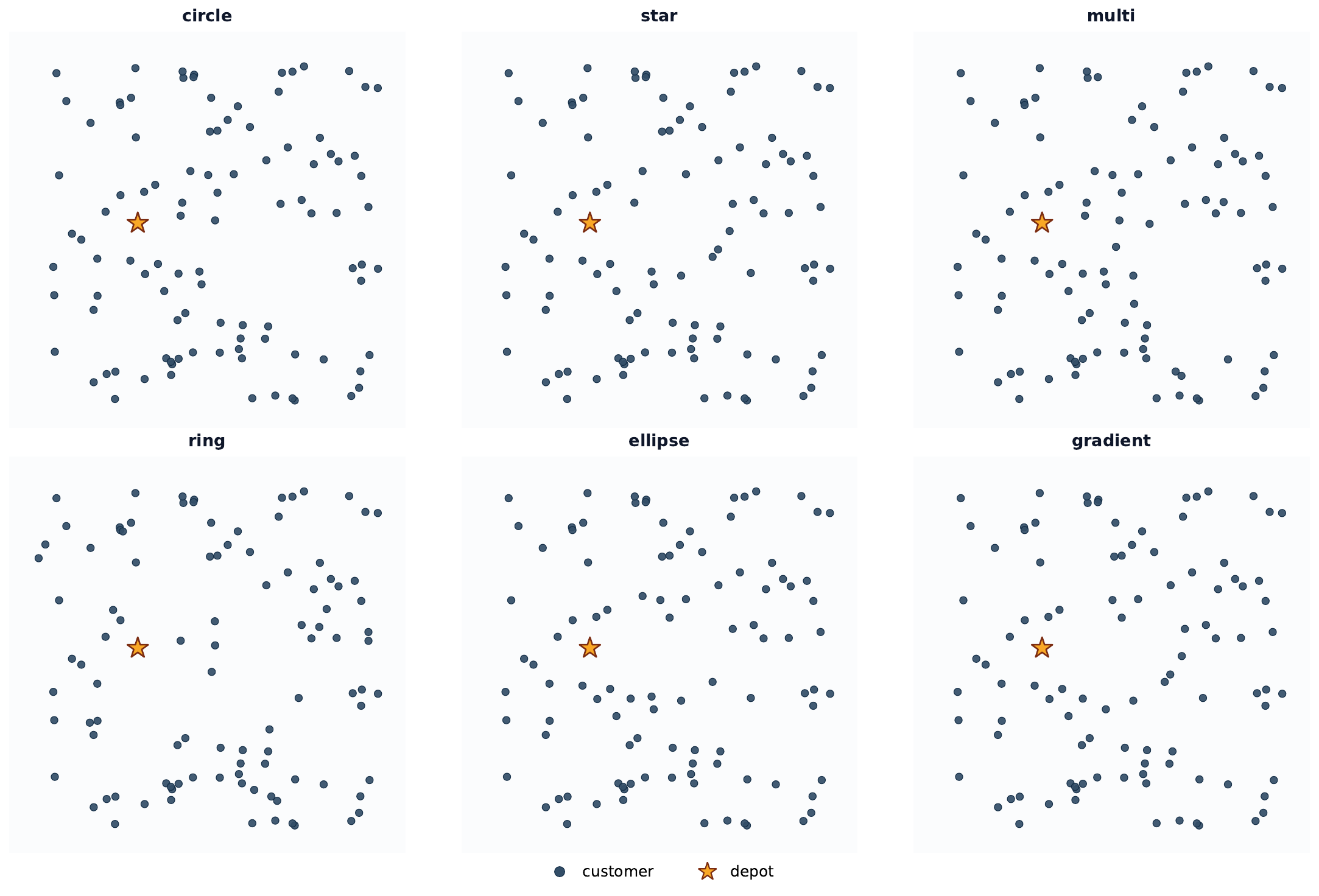}
    \caption{Visualization of the six Explosion distribution variants on CVRP100 instances. The orange star marks the depot.}
    \label{fig:explosion_variants}
\end{figure}

The Explosion distribution family is the structural counterpart of Implosion: instead of injecting a hotspot, Explosion \emph{removes} a structured low-density region (a ``hole'') from an otherwise uniform sample. This captures scenarios where certain areas are inaccessible or systematically depopulated---parks, lakes, military bases, airports, or protected zones. We construct variants whose hole geometry varies across distinct types. Figure~\ref{fig:explosion_variants} visualizes one representative instance from each variant.

\textbf{Circle.}
A single circular hole at a randomly chosen center, modeling a single inaccessible zone such as a park, lake, military base, or airport.

\textbf{Star.}
A non-convex five-pointed star hole, modeling forbidden regions with jagged outlines carved by rivers, railways, or airport perimeters.

\textbf{Multi.}
Three non-overlapping circular holes, modeling distributed forbidden zones such as several parks or scattered restricted areas, requiring inter-hole navigation.

\textbf{Ring.}
An annular exclusion zone separating an inner core from an outer periphery, modeling \emph{moat-and-greenbelt} patterns such as Beijing's second-ring belt or airport runway exclusions, where the optimal path must loop around the entire ring.

\textbf{Ellipse.}
An anisotropic elliptical hole with $2.5{:}1$ aspect ratio, modeling directionally elongated forbidden regions such as runways, canals, or highway median strips.

\textbf{Gradient.}
A soft hole with no hard boundary, where the rejection probability decays smoothly from the center, modeling soft low-attraction zones such as high-risk districts or fringe areas.

\begin{table*}[t]
\centering
\fontsize{6.5pt}{7.5pt}\selectfont
\caption{Generalization performance on CVRP100 across different Explosion distributions. Gap (\%) is computed against LKH3. Best results among neural methods are in \textbf{bold}; second best are \underline{underlined}.}
\label{tab:explosion_full}
\renewcommand{\arraystretch}{0.9}
\setlength{\tabcolsep}{2.5pt}
\begin{tabularx}{\textwidth}{l|CCCCCC|C|C}
\toprule
\rowcolor{headerbg}
& \multicolumn{6}{c|}{\textbf{Explosion distribution variants}} & & \\
\rowcolor{headerbg}
\textbf{Method} & Circle & Star & Multi & Ring & Ellipse & Gradient & \textbf{Average Gap} & \textbf{Time} \\
\midrule
\rowcolor{classicalbg} LKH3 & --- & --- & --- & --- & --- & --- & --- & 23.41s \\
\midrule
\multicolumn{9}{l}{\textit{Single-distribution training}} \\
\rowcolor{singlebg} ELG    & 0.26\% & 0.19\% & 0.07\% & 0.12\% & 0.22\% & 0.25\% & 0.18\% & 4.11s\\
\rowcolor{singlebg} LEHD   & 0.51\% & 0.26\% & 0.08\% & 0.35\% & 0.35\% & 0.41\% & 0.33\% & 4.26s\\
\rowcolor{singlebg} BQ     & 0.65\% & 0.69\% & 0.78\% & 1.01\% & 0.86\% & 0.81\% & 0.80\% & 2.30m\\
\rowcolor{singlebg} INViT  & 6.41\% & 6.47\% & 6.25\% & 6.21\% & 6.49\% & 6.45\% & 6.38\% & 15.14m\\
\midrule
\multicolumn{9}{l}{\textit{Multi-distribution training}} \\
\rowcolor{multibg} AMDKD-AM     & 3.01\% & 2.83\% & 2.68\% & 2.89\% & 3.04\% & 2.80\% & 2.88\% & 1.08s\\
\rowcolor{multibg} AMDKD-POMO   & -0.01\% & -0.05\% & -0.17\% & 0.02\% & 0.17\% & -0.02\% & -0.01\% & 1.05s\\
\rowcolor{multibg} LEHD-VCT+MEM & 2.21\% & 1.84\% & 1.23\% & 2.36\% & 1.45\% & 1.76\% & 1.81\% & 8.14s\\
\midrule
\rowcolor{ourbg} POMO              & \underline{-0.13\%} & \textbf{-0.30\%} & \underline{-0.32\%} & \underline{-0.14\%} & \underline{-0.10\%} & \underline{-0.18\%} & \underline{-0.19\%} & 1.99s\\
\rowcolor{ourbg} \textbf{MixRoute} & \textbf{-0.27\%} & \underline{-0.26\%} & \textbf{-0.47\%} & \textbf{-0.23\%} & \textbf{-0.21\%} & \textbf{-0.23\%} & \textbf{-0.28\%} & 2.25s\\
\bottomrule
\end{tabularx}
\end{table*}

\textbf{Results.} Table~\ref{tab:explosion_full} reports the per-variant performance on the Explosion family. \textsc{MixRoute} achieves the lowest gap on five of the six variants and an average gap of $-0.28\%$, surpassing LKH3 in aggregate and outperforming the second-best baseline (POMO, $-0.19\%$). The exception is \emph{Star}, where POMO ($-0.30\%$) marginally edges out \textsc{MixRoute} ($-0.26\%$); given the small magnitude, we view this as effectively tied performance.

\subsection{Demand Distribution}
\label{app:demand}

\begin{figure}[t]
    \centering
    \includegraphics[width=\textwidth]{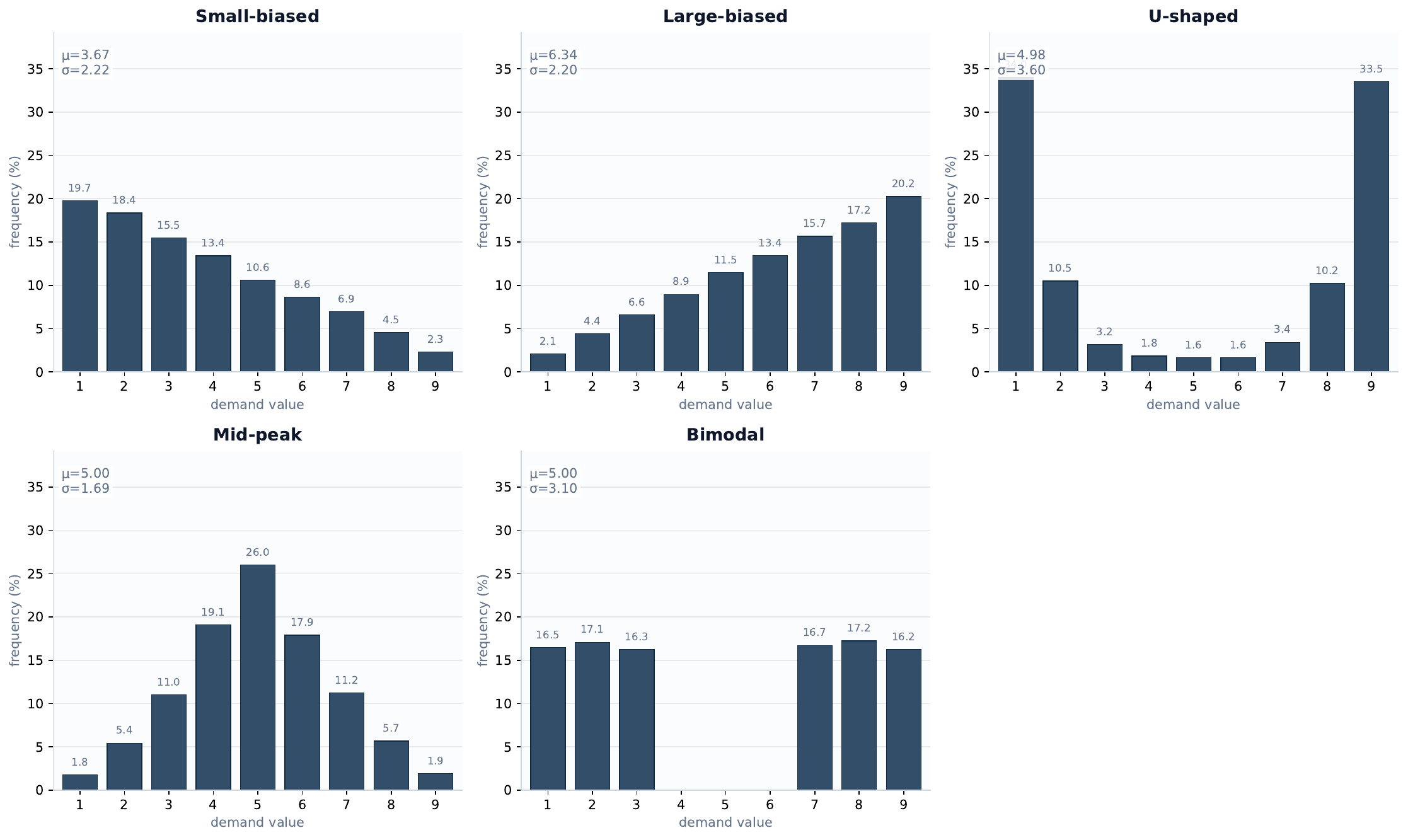}
    \caption{The five Demand distribution variants. Each panel shows the probability mass function over demand values $\{1, 2, \ldots, 9\}$. From left to right: \emph{Small-biased}, \emph{Large-biased}, \emph{U-shaped}, \emph{Mid-peak}, and \emph{Bimodal}.}
    \label{fig:demand_variants}
\end{figure}

The Demand distribution family departs from preceding families by varying \emph{customer demand values} rather than spatial coordinates. Standard CVRP samples demands uniformly from $\{1, 2, \ldots, 9\}$, but real-world ordering patterns exhibit diverse structures: retail last-mile is dominated by small orders, B2B by large orders, and mixed-channel warehouses by bimodal patterns. We construct five demand variants, each motivated by a distinct real-world ordering pattern. Figure~\ref{fig:demand_variants} visualizes the demand probability mass function for each variant.

\textbf{Small-biased.}
Probability mass decreases monotonically from $1$ to $9$ ($\mu \approx 3.67$), modeling retail last-mile patterns dominated by small orders such as food delivery or e-commerce final-mile.

\textbf{Large-biased.}
The mirror of \emph{Small-biased}, with mass increasing monotonically from $1$ to $9$ ($\mu \approx 6.34$), modeling B2B and bulk distribution such as wholesale supply chain or trunk-line logistics.

\textbf{U-shaped.}
Mass concentrates at the two extremes ($\sim 34\%$ each at $1$ and $9$, with intermediate values sparse but non-zero), modeling mixed-channel warehouses serving B2C retail and B2B wholesale customers from the same depot.

\textbf{Mid-peak.}
A symmetric bell-shape centered at demand $5$ with the smallest variance ($\sigma \approx 1.69$), modeling standardized procurement such as case-packed FMCG or routine institutional orders.

\textbf{Bimodal.}
A discontinuous distribution with mass at $\{1, 2, 3\}$ and $\{7, 8, 9\}$ but exactly zero at $\{4, 5, 6\}$, modeling dichotomous order regimes such as tiered packaging or retail-vs-wholesale partitions. The absence of intermediate values is a structural property that no smooth training distribution exposes the model to.

\begin{table*}[t]
\centering
\fontsize{6.5pt}{7.5pt}\selectfont
\caption{Generalization performance on CVRP100 across different Demand distributions. Gap (\%) is computed against LKH3. Best results among neural methods are in \textbf{bold}; second best are \underline{underlined}.}
\label{tab:demand_full}
\renewcommand{\arraystretch}{0.9}
\setlength{\tabcolsep}{2.5pt}
\begin{tabularx}{\textwidth}{l|CCCCC|C|C}
\toprule
\rowcolor{headerbg}
& \multicolumn{5}{c|}{\textbf{Demand distribution variants}} & & \\
\rowcolor{headerbg}
\textbf{Method} & Small-biased & Large-biased & U-shaped & Mid-peak & Bimodal & \textbf{Average Gap} & \textbf{Time} \\
\midrule
\rowcolor{classicalbg} LKH3 & --- & --- & --- & --- & --- & --- & 10.48s \\
\midrule
\multicolumn{8}{l}{\textit{Single-distribution training}} \\
\rowcolor{singlebg} ELG    & 2.21\% & 1.82\% & 2.23\% & 1.60\% & 1.89\% & 1.94\% & 4.48s\\
\rowcolor{singlebg} LEHD   & 2.67\% & 2.96\% & 3.08\% & 2.61\% & 3.05\% & 2.88\% & 4.22s\\
\rowcolor{singlebg} BQ     & 2.25\% & 3.45\% & 3.29\% & 2.28\% & 3.16\% & 2.92\% & 2.30m\\
\rowcolor{singlebg} INViT  & 9.60\% & 10.60\% & 10.00\% & 8.76\% & 9.35\% & 9.69\% & 15.21m\\
\midrule
\multicolumn{8}{l}{\textit{Multi-distribution training}} \\
\rowcolor{multibg} AMDKD-AM     & 5.08\% & 4.34\% & 4.96\% & 4.63\% & 5.00\% & 4.78\% & 2.35s\\
\rowcolor{multibg} AMDKD-POMO   & 2.20\% & 1.83\% & 2.00\% & 1.48\% & 1.90\% & 1.87\% & 1.01s\\
\rowcolor{multibg} LEHD-VCT+MEM & 1.92\% & 2.84\% & 2.98\% & 2.44\% & 2.98\% & 2.66\% & 8.88s\\
\midrule
\rowcolor{ourbg} POMO              & \underline{1.77\%} & \underline{1.36\%} & \underline{1.69\%} & \underline{1.17\%} & \underline{1.53\%} & \underline{1.49\%} & 1.97s\\
\rowcolor{ourbg} \textbf{MixRoute} & \textbf{1.25\%} & \textbf{1.27\%} & \textbf{1.43\%} & \textbf{0.96\%} & \textbf{1.29\%} & \textbf{1.24\%} & 2.25s\\
\bottomrule
\end{tabularx}
\end{table*}

\textbf{Results.} Table~\ref{tab:demand_full} reports the per-variant performance on the Demand family. \textsc{MixRoute} achieves the lowest gap on every variant, with an average gap of $1.24\%$ that outperforms the second-best baseline (POMO, $1.49\%$). Notably, the second-best method on this family is POMO rather than any multi-distribution baseline, suggesting that exposure to additional spatial distributions transfers poorly to demand-distribution shifts.

\subsection{Capacity Configuration}
\label{app:capacity}

The Capacity dimension perturbs the vehicle capacity $Q$ while keeping node coordinates and customer demands fixed at the standard uniform setting. Since demands enter the model only through the normalized ratio $d_i / Q$, scaling $Q$ is equivalent to rescaling all demands proportionally, providing a controlled probe of how solvers handle changes in the \emph{numerical regime} of the input.

We evaluate five capacity settings, $Q \in \{20, 80, 120, 200, 250\}$, inducing a more than tenfold variation in the per-customer load $\bar{d}/Q$ (from $0.250$ at $Q=20$ to $0.020$ at $Q=250$). All five settings lie outside the standard CVRP training capacity ($Q=50$) and thus constitute out-of-distribution evaluation. Table~\ref{tab:capacity_full} reports the per-setting performance.

\begin{table*}[t]
\centering
\fontsize{6.5pt}{7.5pt}\selectfont
\caption{Generalization performance on CVRP100 across different Capacity settings. Gap (\%) is computed against LKH3. Best results among neural methods are in \textbf{bold}; second best are \underline{underlined}.}
\label{tab:capacity_full}
\renewcommand{\arraystretch}{0.9}
\setlength{\tabcolsep}{2.5pt}
\begin{tabularx}{\textwidth}{l|CCCCC|C|C}
\toprule
\rowcolor{headerbg}
& \multicolumn{5}{c|}{\textbf{Capacity setting}} & & \\
\rowcolor{headerbg}
\textbf{Method} & $Q=20$ & $Q=80$ & $Q=120$ & $Q=200$ & $Q=250$ & \textbf{Average Gap} & \textbf{Time} \\
\midrule
\rowcolor{classicalbg} LKH3 & --- & --- & --- & --- & --- & --- & 24.30s \\
\midrule
\multicolumn{8}{l}{\textit{Single-distribution training}} \\
\rowcolor{singlebg} ELG    & 2.79\% & 2.66\% & 5.57\% & 8.42\% & 11.67\% & 4.89\% & 3.44s\\
\rowcolor{singlebg} LEHD   & 4.37\% & 3.23\% & 3.25\% & \underline{2.80\%} & \underline{5.60\%} & 3.97\% & 3.55s\\
\rowcolor{singlebg} BQ     & 7.66\% & \underline{1.61\%} & \underline{2.57\%} & 2.97\% & 6.24\% & 5.18\% & 2.51m\\
\rowcolor{singlebg} INViT  & 12.10\% & 8.02\% & 10.97\% & 13.60\% & 14.60\% & 11.73\% & 12.30m\\
\midrule
\multicolumn{8}{l}{\textit{Multi-distribution training}} \\
\rowcolor{multibg} AMDKD-AM     & 4.22\% & 5.50\% & 8.37\% & 12.52\% & 13.41\% & 7.12\% & 0.88s\\
\rowcolor{multibg} AMDKD-POMO   & 4.12\% & 2.45\% & 5.62\% & 8.99\% & 11.69\% & 5.53\% & 0.88s\\
\rowcolor{multibg} \textbf{LEHD-VCT+MEM} & \textbf{1.97\%} & 2.75\% & \textbf{1.58\%} & \textbf{0.11\%} & \textbf{0.82\%} & \textbf{1.68\%} & 6.79s\\
\midrule
\rowcolor{ourbg} POMO              & 3.88\% & 1.98\% & 5.86\% & 18.04\% & 27.46\% & 8.34\% & 1.76s\\
\rowcolor{ourbg} \textbf{MixRoute} & \underline{2.47\%} & \textbf{1.31\%} & 2.66\% & 5.13\% & 8.75\% & \underline{3.36\%} & 1.86s\\
\bottomrule
\end{tabularx}
\end{table*}

\subsection{Time Window Shifts}
\label{app:timewindow}

Among the three constraint dimensions in CVRPTW, time windows are arguably the most multi-faceted: a time-window assignment involves not only \emph{how wide} the window is, but also \emph{when} during the operating day it falls, and \emph{how} it correlates with the customer's spatial position relative to the depot. Standard CVRPTW benchmarks conflate these three aspects, making it difficult to attribute failures to any specific axis. We decompose time-window shifts into three orthogonal dimensions---\emph{Width}, \emph{Timing}, and \emph{Coupling}---each isolating one aspect while holding the others at training-distribution values.

\textbf{Width: varying the time-window length.}
This dimension perturbs $\texttt{tw\_length}$ while leaving the starting-time distribution unchanged:
\begin{itemize}
    \item \emph{tw\_loose}: $\texttt{tw\_length} \sim \text{Uniform}([0.5, 1.0])$, with windows wide enough to render time-window constraints nearly vacuous.
    \item \emph{tw\_tight}: $\texttt{tw\_length} \sim \text{Uniform}([0.05, 0.10])$, simulating precise appointments that allow virtually no scheduling flexibility.
    \item \emph{tw\_bimodal}: a $50/50$ mixture of tight and loose, modeling scenarios where some customers require precise scheduling (e.g., VIP deliveries) while others tolerate flexible service.
\end{itemize}

\textbf{Timing: varying when service occurs during the day.}
This dimension fixes the window length but perturbs $\texttt{tw\_start}$, changing \emph{when} during the operating horizon the customer load is concentrated:
\begin{itemize}
    \item \emph{tw\_late\_rush}: $\texttt{tw\_start} \sim \text{Uniform}([0.7 \cdot \text{MAX}, \text{MAX}])$, simulating an evening rush with morning idleness.
    \item \emph{tw\_narrow\_band}: $\texttt{tw\_start} \sim \mathcal{N}(\text{MAX}/2, 0.2)$, forcing parallel dispatch within a single mid-day window.
    \item \emph{tw\_stripes\_2}: two narrow bands at $0.25 \cdot \text{MAX}$ and $0.75 \cdot \text{MAX}$ with $\sigma = 0.05$, modeling morning and afternoon dispatch waves.
\end{itemize}
The standard training generator induces a mild positive correlation between $\texttt{tw\_start}$ and the depot-to-customer distance, which all three variants break in different ways.

\textbf{Coupling: varying the spatial--temporal correlation.}
This dimension fixes both window length and starting-time distribution but severs the implicit dependency between $\texttt{tw\_start}$ and the depot-to-customer distance:
\begin{itemize}
    \item \emph{tw\_decoupled}: $\texttt{tw\_start} \sim \text{Uniform}([\text{lo}, \text{hi}])$, sampled independently of $d_{0i}$.
\end{itemize}
The standard training distribution generates $\texttt{tw\_start}$ via a formula that scales with $d_{0i}$, inducing a strong ``far implies late'' prior that models may overfit to as a shortcut. \emph{tw\_decoupled} severs this prior, testing whether the model has learned genuine spatial--temporal reasoning rather than a distance-based heuristic.

\section{Multi-seed results}
\label{app:seeds}

We retrained MixRoute four times with different seeds and report the mean $\pm$ standard deviation across runs in Table~\ref{tab:seeds}, where run~1 corresponds to the
model reported in the main paper. We report a representative
subset of six distributions, including Implosion, Explosion, and Demand, on
which the margin of MixRoute over POMO is among the smallest in our benchmark.

\begin{table}[h]
\centering
\caption{Results of MixRoute across four independent training runs on CVRP100.
Gap (\%) is computed against LKH3. The mean $\pm$ standard deviation is
computed over the four runs.}
\label{tab:seeds}
\resizebox{\linewidth}{!}{%
\begin{tabular}{l|cccccc}
\toprule
\rowcolor{headerbg}
\textbf{Method (CVRP)} & Annulus & Expansion & Implosion & Explosion & Demand & Capacity \\
\midrule
\rowcolor{singlebg} POMO & 7.11\% & 0.65\% & -0.15\% & -0.19\% & 1.49\% & 8.34\% \\
\midrule
\rowcolor{multibg} MixRoute (run 1) & -0.87\% & -0.13\% & -0.30\% & -0.28\% & 1.24\% & 3.36\% \\
\rowcolor{multibg} MixRoute (run 2) & -0.59\% & -0.25\% & -0.50\% & -0.48\% & 1.36\% & 4.05\% \\
\rowcolor{multibg} MixRoute (run 3) & -0.56\% & -0.31\% & -0.48\% & -0.51\% & 1.30\% & 3.79\% \\
\rowcolor{multibg} MixRoute (run 4) & -0.57\% & -0.30\% & -0.51\% & -0.55\% & 1.33\% & 3.40\% \\
\midrule
\rowcolor{ourbg} \textbf{MixRoute (mean $\pm$ std)}
 & \textbf{-0.65$\pm$0.15\%} & \textbf{-0.25$\pm$0.08\%}
 & \textbf{-0.45$\pm$0.10\%} & \textbf{-0.46$\pm$0.12\%}
 & \textbf{1.31$\pm$0.05\%} & \textbf{3.65$\pm$0.33\%} \\
\bottomrule
\end{tabular}}
\end{table}

The variation across runs is small: the standard deviation is at most 0.15\%
on five of the six distributions, and 0.33\% on Capacity, where the margin over
POMO (4.69\%) remains more than an order of magnitude larger. More importantly,
the variation does not affect the conclusion, as each of the four runs
outperforms POMO on all six distributions.

\section{Mechanism of MixNorm}
\label{sec:analysis}

To understand how MixNorm responds to distributional shift, we measure how far the encoder representation moves when the input distribution changes. We define the \emph{mean shift} at each encoder layer as the $L_2$ distance between the mean node embedding under a given OOD distribution and that under the in-distribution setting, and group OOD distributions by the improvement of MixRoute over POMO.

\begin{table}[htbp]
\centering
\footnotesize
\renewcommand{\arraystretch}{1.1}
\caption{Mean embedding shift across encoder layers, grouped by the improvement of MixRoute over POMO. $\Delta$ is the difference in the layer-averaged shift between the two models; the last column reports the average gap reduction within each group.}
\label{tab:analysis_shift}
\begin{tabularx}{\textwidth}{l l *{6}{>{\centering\arraybackslash}X} | >{\centering\arraybackslash}X >{\centering\arraybackslash}X}
\toprule
\rowcolor{headerbg}
\textbf{Group} & \textbf{Model} & L0 & L1 & L2 & L3 & L4 & L5 & $\boldsymbol{\Delta}$ & \textbf{Improv.} \\
\midrule
\rowcolor{ourbg} $>$5\% & POMO & 0.08 & 0.11 & 0.11 & 0.10 & 0.09 & 0.06 & & \\
\rowcolor{ourbg} & MixRoute & 4.72 & 4.94 & 3.48 & 4.57 & 3.44 & 2.31 & 3.82 & +22.6\% \\
\midrule
\rowcolor{multibg} 1--5\% & POMO & 0.04 & 0.05 & 0.04 & 0.03 & 0.02 & 0.01 & & \\
\rowcolor{multibg} & MixRoute & 1.35 & 1.45 & 1.01 & 1.15 & 0.91 & 0.67 & 1.06 & +3.7\% \\
\midrule
\rowcolor{singlebg} $<$1\% & POMO & 0.03 & 0.03 & 0.03 & 0.02 & 0.02 & 0.02 & & \\
\rowcolor{singlebg} & MixRoute & 0.21 & 0.97 & 0.94 & 0.89 & 0.72 & 0.67 & 0.70 & +0.2\% \\
\bottomrule
\end{tabularx}
\end{table}

As shown in Table~\ref{tab:analysis_shift}, the encoder representation of POMO barely moves under distributional shift, whereas that of MixRoute responds substantially, and the magnitude of the response grows with the improvement over POMO. Extending the analysis to 16 additional distributions, the embedding shift of MixRoute is positively associated with its improvement over POMO (Pearson $r = 0.651$, Spearman $\rho = 0.685$, $n = 16$). We emphasize that this association is descriptive rather than causal: a larger mean shift does not by itself lead to better performance. We next turn to the gate itself, whose behavior is analyzed in Appendix~\ref{sec:gate_analysis}.

\section{Behavior of the Adaptive Gate}
\label{sec:gate_analysis}

We examine the learned gate weights from four perspectives: across encoder layers, over the course of training, across distribution families, and in response to specific instance-level characteristics. Recall that MixNorm is applied after both the multi-head self-attention (MHA) and the feed-forward (FFN) sublayers of each encoder layer.

\paragraph{Layer-wise patterns.}
Table~\ref{tab:gate_layer} reports the mean gate weights at each layer. The two insertion points exhibit distinct patterns. After MHA, BN receives the largest weight at every depth, while the contribution of IN is concentrated in the shallowest layer. After FFN, BN dominates only at L0, and LN becomes the largest component in all subsequent layers.

\begin{table}[htbp]
\centering
\footnotesize
\renewcommand{\arraystretch}{1.05}
\setlength{\tabcolsep}{6pt}
\caption{Mean gate weights of each normalizer across encoder layers.}
\label{tab:gate_layer}
\begin{tabular}{l|ccc|ccc}
\toprule
& \multicolumn{3}{c|}{\textbf{Post-MHA}} & \multicolumn{3}{c}{\textbf{Post-FFN}} \\
\textbf{Layer} & IN & LN & BN & IN & LN & BN \\
\midrule
L0 & 0.43 & 0.01 & 0.57 & 0.03 & 0.26 & 0.71 \\
L1 & 0.19 & 0.40 & 0.41 & 0.30 & 0.47 & 0.23 \\
L2 & 0.28 & 0.15 & 0.58 & 0.00 & 0.99 & 0.01 \\
L3 & 0.20 & 0.13 & 0.67 & 0.00 & 0.70 & 0.30 \\
L4 & 0.03 & 0.05 & 0.91 & 0.05 & 0.64 & 0.32 \\
L5 & 0.05 & 0.19 & 0.77 & 0.38 & 0.53 & 0.09 \\
\bottomrule
\end{tabular}
\end{table}

\paragraph{Training dynamics.}
Table~\ref{tab:gate_epoch} tracks the gate weights of the post-MHA MixNorm at layer L5 over four checkpoints. The weights evolve steadily throughout training, with BN gradually absorbing weight from IN.

\begin{table}[htbp]
\centering
\footnotesize
\renewcommand{\arraystretch}{1.05}
\setlength{\tabcolsep}{8pt}
\caption{Gate weights of the post-MHA MixNorm at layer L5 over training.}
\label{tab:gate_epoch}
\begin{tabular}{l|ccc}
\toprule
\textbf{Epoch} & IN & LN & BN \\
\midrule
500  & 0.21 & 0.23 & 0.56 \\
2000 & 0.14 & 0.21 & 0.65 \\
4000 & 0.11 & 0.18 & 0.71 \\
6500 & 0.05 & 0.19 & 0.77 \\
\bottomrule
\end{tabular}
\end{table}

\paragraph{Distribution-level adaptation.}
To test whether the gate adapts to distributional structure, we compare the gate weights under Uniform, Annulus (thin ring), and Stripe (spike) instances using Welch's $t$-test for all three pairwise comparisons, with Bonferroni correction ($\alpha = 0.05/3 \approx 0.0167$). As shown in Table~\ref{tab:gate_ttest}, the gate weights differ significantly between distribution families in the vast majority of comparisons, indicating that the gate adapts its weighting over normalizers to the input distribution.

\begin{table}[htbp]
\centering
\scriptsize
\renewcommand{\arraystretch}{1.02}
\setlength{\tabcolsep}{3pt}
\caption{$p$-values of Welch's $t$-test on gate weights between distribution families (U: Uniform, A: Annulus, S: Stripe). Values in parentheses are not significant after Bonferroni correction ($\alpha=0.0167$).}
\label{tab:gate_ttest}
\begin{tabular}{l|ccc|ccc|ccc}
\toprule
& \multicolumn{3}{c|}{\textbf{IN}} & \multicolumn{3}{c|}{\textbf{LN}} & \multicolumn{3}{c}{\textbf{BN}} \\
\textbf{Layer} & U vs A & U vs S & A vs S & U vs A & U vs S & A vs S & U vs A & U vs S & A vs S \\
\midrule
L0-attn & (6e-02) & 4e-37 & 4e-70 & 3e-33 & 4e-25 & 2e-89 & (5e-02) & 1e-31 & 3e-44 \\
L0-ffn & 7e-07 & 6e-29 & 1e-39 & 2e-50 & 8e-90 & 5e-195 & 4e-79 & 1e-105 & 2e-181 \\
L1-attn & (7e-02) & 1e-07 & 2e-155 & 5e-19 & 8e-10 & 2e-149 & 3e-06 & 9e-04 & 2e-13 \\
L1-ffn & 2e-55 & 1e-92 & 6e-189 & 3e-05 & 6e-04 & (2e-02) & 4e-61 & 4e-65 & 7e-162 \\
L2-attn & 2e-04 & 4e-58 & 2e-109 & 3e-70 & 3e-58 & 7e-119 & 1e-40 & 3e-13 & 2e-58 \\
L2-ffn & 1e-69 & 3e-55 & 3e-16 & 3e-20 & 5e-29 & 3e-41 & 1e-20 & 8e-29 & 3e-41 \\
L3-attn & 2e-28 & 1e-45 & 1e-102 & 2e-114 & 2e-55 & 6e-137 & 2e-73 & 5e-58 & 1e-136 \\
L3-ffn & 3e-39 & 1e-41 & (9e-02) & 4e-121 & 9e-28 & 1e-196 & 8e-132 & 1e-22 & 5e-197 \\
L4-attn & 3e-43 & 3e-19 & 6e-84 & 8e-33 & 4e-42 & 9e-04 & 9e-68 & (8e-02) & 2e-81 \\
L4-ffn & 1e-115 & 8e-87 & 6e-12 & 2e-53 & 1e-64 & 3e-06 & 2e-07 & 9e-25 & 2e-18 \\
L5-attn & 3e-55 & 2e-59 & (3e-02) & 3e-29 & 7e-41 & 4e-98 & 2e-47 & 5e-13 & 3e-98 \\
L5-ffn & 8e-28 & 8e-29 & (2e-01) & 8e-17 & 7e-42 & 4e-98 & 4e-142 & 2e-06 & 9e-200 \\
\bottomrule
\end{tabular}
\end{table}

\paragraph{Instance-level adaptation.}
Beyond distinguishing distribution families, we examine how the gate responds to two instance-level characteristics that vary continuously within a family: the spatial dispersion of nodes (Gaussian family) and the tightness of the capacity constraint (Capacity family). Tables~\ref{tab:gate_gaussian} and~\ref{tab:gate_capacity} report the mean and standard deviation of the gate weights at layer L5, expressed as ratios relative to the corresponding in-distribution baseline. Both reveal systematic trends.

\begin{table}[htbp]
\centering
\footnotesize
\renewcommand{\arraystretch}{1.05}
\setlength{\tabcolsep}{5pt}
\caption{Gate weight statistics at layer L5 under the Gaussian shift, as ratios relative to the Uniform baseline. From \emph{tight} to \emph{broad}, nodes become progressively more dispersed.}
\label{tab:gate_gaussian}
\begin{tabular}{l|cccc|cccc}
\toprule
& \multicolumn{4}{c|}{\textbf{Post-MHA}} & \multicolumn{4}{c}{\textbf{Post-FFN}} \\
\textbf{Statistic} & Tight & Compact & Loose & Broad & Tight & Compact & Loose & Broad \\
\midrule
IN mean & 1.11 & 1.10 & 1.08 & 1.04 & 1.10 & 1.09 & 1.06 & 1.03 \\
LN mean & 1.05 & 0.95 & 0.91 & 0.87 & 0.92 & 0.93 & 0.95 & 0.96 \\
BN mean & 0.98 & 1.00 & 1.01 & 1.02 & 0.94 & 0.97 & 1.01 & 1.07 \\
\midrule
IN std & 0.76 & 0.78 & 0.74 & 0.73 & 0.66 & 0.61 & 0.64 & 0.62 \\
LN std & 0.48 & 0.42 & 0.42 & 0.40 & 0.75 & 0.67 & 0.67 & 0.63 \\
BN std & 0.54 & 0.50 & 0.52 & 0.50 & 0.67 & 0.70 & 0.79 & 0.82 \\
\bottomrule
\end{tabular}
\end{table}

\emph{Spatial dispersion.} Under the Gaussian shift (Table~\ref{tab:gate_gaussian}), the standard deviation of the gate weights is uniformly lower than under Uniform, indicating a more consistent weighting across tokens. Moreover, as the nodes become progressively more dispersed, the mean weight of IN declines steadily while that of BN rises, a trend that holds at both insertion points.

\begin{table}[htbp]
\centering
\footnotesize
\renewcommand{\arraystretch}{1.05}
\setlength{\tabcolsep}{4pt}
\caption{Gate weight statistics at layer L5 under the Capacity shift, as ratios relative to the $Q=50$ baseline. Larger $Q$ corresponds to a looser capacity constraint.}
\label{tab:gate_capacity}
\begin{tabular}{l|ccccc|ccccc}
\toprule
& \multicolumn{5}{c|}{\textbf{Post-MHA}} & \multicolumn{5}{c}{\textbf{Post-FFN}} \\
\textbf{Statistic} & 20 & 80 & 120 & 200 & 250 & 20 & 80 & 120 & 200 & 250 \\
\midrule
IN mean & 0.53 & 1.39 & 1.71 & 1.90 & 1.93 & 0.84 & 0.99 & 0.95 & 0.90 & 0.88 \\
LN mean & 1.17 & 1.01 & 1.03 & 1.04 & 1.05 & 1.13 & 0.99 & 1.01 & 1.02 & 1.03 \\
BN mean & 1.01 & 0.95 & 0.91 & 0.89 & 0.88 & 1.07 & 1.08 & 1.21 & 1.37 & 1.43 \\
\midrule
IN std & 0.49 & 1.36 & 1.47 & 1.60 & 1.65 & 1.31 & 0.89 & 0.97 & 1.17 & 1.23 \\
LN std & 0.81 & 1.07 & 1.13 & 1.13 & 1.11 & 1.48 & 0.93 & 0.95 & 1.08 & 1.14 \\
BN std & 0.83 & 1.12 & 1.22 & 1.31 & 1.33 & 1.01 & 1.06 & 1.19 & 1.37 & 1.42 \\
\bottomrule
\end{tabular}
\end{table}

\emph{Capacity tightness.} The effect of capacity diverges between the two insertion points (Table~\ref{tab:gate_capacity}). After MHA, the mean weight of IN rises sharply as capacity loosens, nearly doubling at $Q=250$, while that of BN declines. After FFN, the trend reverses: BN gains substantial weight as capacity loosens, reaching $1.43\times$ its baseline at $Q=250$. Together, these results suggest that the gate responds not only to which distribution family an instance belongs to, but also to continuous variations in its structure.

\section{Comparison with Existing Mixing Normalizations}
\label{sec:compare_existing}

To position MixRoute relative to existing learning-to-normalize approaches, we compare against Switchable Normalization~\citep{luo2019differentiable}---the most widely adopted mixing normalization in the literature---together with a stronger variant of SN that we design as an additional baseline. We adopt \texttt{IN+LN+BN} as the constituent set, following the standard configuration in the original SN paper.

\textbf{Switchable Normalization (SN).}
SN performs \emph{statistic mixing}: it computes the mean and variance of each constituent normalizer, takes a learnable weighted average of these statistics, and applies a single shared affine transformation $(\gamma, \beta)$ to the result:
\begin{equation}
\text{SN}(x) = \gamma \cdot \frac{x - \sum_{i} w_i \mu_i}{\sqrt{\sum_{i} w_i' \sigma_i^2 + \epsilon}} + \beta,
\end{equation}
where $w_i$ and $w_i'$ are softmax-normalized scalar weights learned end-to-end, and $(\gamma, \beta)$ are shared across all constituent normalizers.

\textbf{SN-affine (our designed baseline).}
To stress-test whether SN's statistic-mixing paradigm can be pushed further, we design a stronger variant that we refer to as \textbf{SN-affine}, in which each constituent normalizer is additionally assigned its own affine parameters $(\gamma_i, \beta_i)$. The effective affine applied to the mixed-statistic output is itself a weighted combination using the same mixing weights:
\begin{equation}
\text{SN-affine}(x) = \Big(\sum_{i} w_i \gamma_i\Big) \cdot \frac{x - \sum_{i} w_i \mu_i}{\sqrt{\sum_{i} w_i' \sigma_i^2 + \epsilon}} + \sum_{i} w_i \beta_i.
\end{equation}

\begin{table*}[!htbp]
\centering
\small
\caption{Comparison with existing mixing normalizations on the CVRP benchmark. All variants use \texttt{IN+LN+BN} as the constituent set. Best results among neural methods are in \textbf{bold}.}
\label{tab:compare_mixing_norms}
\setlength{\tabcolsep}{4pt}
\renewcommand{\arraystretch}{1.15}
\resizebox{\textwidth}{!}{%
\begin{tabular}{l|ccccccccc|cc|c}
\toprule
\rowcolor{headerbg}
& \multicolumn{9}{c|}{\textbf{Node coordinate distributions}} & \multicolumn{2}{c|}{\textbf{CVRP-specific}} & \\
\rowcolor{headerbg}
\textbf{Method} & Annulus & Beta & Cluster & Expansion & Explosion & Gaussian & Grid & Implosion & Stripe & Demand & Capacity & \textbf{Avg} \\
\midrule
\rowcolor{classicalbg} LKH3 & --- & --- & --- & --- & --- & --- & --- & --- & --- & --- & --- & --- \\
\midrule
\rowcolor{singlebg} SN & -0.68\% & 1.17\% & 1.90\% & -0.11\% & -0.28\% & 1.45\% & 1.18\% & \textbf{-0.30\%} & 1.57\% & 1.22\% & 8.98\% & 1.46\% \\
\rowcolor{singlebg} SN-affine & -0.86\% & 1.12\% & 1.83\% & -0.07\% & -0.28\% & 1.40\% & \textbf{1.12\%} & -0.26\% & \textbf{1.06\%} & 1.31\% & 8.01\% & 1.31\% \\
\midrule
\rowcolor{ourbg} \textbf{MixRoute (ours)} & \textbf{-0.87\%} & \textbf{1.11\%} & \textbf{1.82\%} & \textbf{-0.13\%} & \textbf{-0.28\%} & \textbf{1.36\%} & 1.18\% & \textbf{-0.30\%} & 1.13\% & \textbf{1.24\%} & \textbf{3.36\%} & \textbf{0.88\%} \\
\bottomrule
\end{tabular}}
\end{table*}

Table~\ref{tab:compare_mixing_norms} shows that MixRoute achieves the lowest average gap (0.88\%), outperforming both SN (1.46\%) and SN-affine (1.31\%). Notably, although SN-affine pushes SN's design to its limit by additionally switching the affine parameters, it still falls short of MixRoute by a clear margin. 

\section{Normalizers Choice}
\label{app:pool}

The constituent set of MixNorm follows Switchable Normalization~\citep{luo2019differentiable}, which combines statistics estimated instance-wise, layer-wise, and minibatch-wise through IN, LN, and BN. This appendix examines how the choice of constituent set affects performance, training dynamics, and inference cost. All experiments follow the CVRP setting of Section~\ref{sec:mixroute}, and we report a representative subset of five distributions.

We first vary the number of constituent normalizers while keeping IN+LN as the base pair. As shown in Table~\ref{tab:pool_size}, expanding the pool from two to three normalizers yields a clear gain, reducing the average gap from $1.93\%$ to $0.77\%$. Adding a fourth component (No Norm) brings no further benefit.

\begin{table}[htbp]
\centering
\small
\caption{Effect of pool size on the CVRP benchmark. Gap (\%) is computed against LKH3 and averaged over the five distributions reported here. Best results are in \textbf{bold}.}
\label{tab:pool_size}
\setlength{\tabcolsep}{6pt}
\renewcommand{\arraystretch}{1.15}
\begin{tabular}{l|ccccc|c}
\toprule
\rowcolor{headerbg}
\textbf{Method} & Uniform & Annulus & Stripe & Demand & Capacity & \textbf{Avg} \\
\midrule
\rowcolor{singlebg} POMO & -0.93\% & 7.11\% & 8.30\% & 1.49\% & 8.34\% & 4.86\% \\
\midrule
\rowcolor{ourbg} MixRoute-IN+LN & -0.23\% & 0.36\% & 2.77\% & 1.58\% & 5.16\% & 1.93\% \\
\rowcolor{ourbg} \textbf{MixRoute-IN+LN+BN} & -1.03\% & \textbf{-0.87\%} & \textbf{1.13\%} & \textbf{1.24\%} & 3.36\% & \textbf{0.77\%} \\
\rowcolor{ourbg} MixRoute-IN+LN+BN+No Norm & \textbf{-1.04\%} & -0.53\% & 1.29\% & \textbf{1.18\%} & \textbf{3.18\%} & 0.82\% \\
\bottomrule
\end{tabular}
\end{table}

To test whether the benefit depends on a particular combination, we further evaluate pools that exclude the IN+LN base pair. As shown in Table~\ref{tab:pool_composition}, these variants still substantially outperform POMO, reducing the average gap from $4.86\%$ to $1.12\%$ and $1.07\%$ respectively. This indicates that the gain stems from adaptively mixing complementary normalizers rather than from any specific combination, although IN+LN+BN remains the best overall.

\begin{table}[htbp]
\centering
\small
\caption{Effect of pool composition on the CVRP benchmark. Gap (\%) is computed against LKH3. Best results are in \textbf{bold}.}
\label{tab:pool_composition}
\setlength{\tabcolsep}{6pt}
\renewcommand{\arraystretch}{1.15}
\begin{tabular}{l|ccccc|c}
\toprule
\rowcolor{headerbg}
\textbf{Method} & Uniform & Annulus & Stripe & Demand & Capacity & \textbf{Avg} \\
\midrule
\rowcolor{singlebg} POMO & -0.93\% & 7.11\% & 8.30\% & 1.49\% & 8.34\% & 4.86\% \\
\midrule
\rowcolor{ourbg} \textbf{MixRoute-IN+LN+BN} & \textbf{-1.03\%} & -0.87\% & 1.13\% & \textbf{1.24\%} & \textbf{3.36\%} & \textbf{0.77\%} \\
\rowcolor{ourbg} MixRoute-IN+BN+No Norm & -0.98\% & \textbf{-0.88\%} & \textbf{1.03\%} & 1.19\% & 5.24\% & 1.12\% \\
\rowcolor{ourbg} MixRoute-LN+BN+No Norm & -1.01\% & -0.68\% & 1.81\% & 1.52\% & 3.71\% & 1.07\% \\
\bottomrule
\end{tabular}
\end{table}

\section{Limitations}

While \textsc{MixRoute} achieves strong cross-distribution generalization across diverse routing problems and backbones, a limitation is worth noting. Our \textsc{MixNorm} module is integrated within the encoder layers' Add \& Norm sub-blocks, which makes its applicability tied to architectures with sufficiently deep encoder stacks. Modern lightweight or decoder-only neural routing solvers~\citep{bqnco, lehd}, which place most of the modeling capacity in the decoder and rely on shallow or even single-layer encoders, offer fewer normalization sites for \textsc{MixNorm} to take effect, potentially limiting the gain that mixing-based normalization can deliver. Extending \textsc{MixNorm} to the decoder side, or designing decoder-friendly mixing schemes, is a natural direction for future work. 

\section{Broader Impacts}

\textsc{MixRoute} aims to improve the cross-distribution generalization of neural routing solvers, which carries both positive and negative societal implications. On the positive side, more robust routing solvers can be deployed across diverse real-world conditions without retraining, directly translating to shorter vehicle routes, reduced fuel consumption, and lower carbon emissions in logistics operations. Our single-distribution training paradigm also requires substantially less compute than multi-distribution alternatives, modestly reducing the environmental footprint of training itself. On the potential negative side, increased automation in routing and dispatch may reduce demand for manual planning labor in the logistics sector, and routing optimization is inherently dual-use, with possible applications beyond civilian logistics. As our work is methodological in nature and evaluated on synthetic benchmarks, we do not foresee a direct path to malicious use, but we encourage practitioners to weigh these implications when integrating such methods into deployed systems.

\section{Asset and Dataset}

Table~\ref{tab:license} lists all third-party codebases and datasets used in our work, along with their licenses and usage.

\begin{table*}[!htbp]
\centering
\small
\setlength{\tabcolsep}{6pt}
\caption{Used assets, licenses, and their usage.}
\label{tab:license}
\begin{tabular}{@{}lllc@{}}
\toprule
\textbf{Type} & \textbf{Asset} & \textbf{License} & \textbf{Usage} \\
\midrule
\multirow{10}{*}{Code}
  & LKH3~\citep{lkh}                      & BSD 3-Clause License   & Evaluation (reference solver) \\
  & HGS~\citep{hgs}           & MIT License            & Evaluation (reference solver, CVRPTW) \\
  & POMO~\citep{pomo}                     & MIT License            & Backbone \& Baseline \\
  & MatNet~\citep{matnet}                 & MIT License            & Backbone (ACVRP) \\
  & ELG~\citep{gao2024towards}            & MIT License            & Baseline \\
  & LEHD~\citep{lehd}                     & MIT License            & Baseline \\
  & BQ-NCO~\citep{bqnco}                  & Apache-2.0 License     & Baseline \\
  & INViT~\citep{fang2024invit}           & MIT License            & Baseline \\
  & AMDKD~\citep{bi2022learning}          & MIT License            & Baseline \\
  & LEHD-VCT+MEM~\citep{hard-constraint}  & MIT License            & Baseline \\
\midrule
Dataset
  & CVRPLIB~\citep{uchoa2017new}         & Publicly available for research use & Real-world evaluation (Sec.~5.4) \\
\bottomrule
\end{tabular}
\end{table*}






\newpage
\section*{NeurIPS Paper Checklist}

\begin{enumerate}

\item {\bf Claims}
    \item[] Question: Do the main claims made in the abstract and introduction accurately reflect the paper's contributions and scope?
    \item[] Answer: \answerYes{} 
    \item[] Justification: The abstract and introduction state the paper's contributions, which are supported by the experiments and analyses presented in the main text.
    \item[] Guidelines:
    \begin{itemize}
        \item The answer \answerNA{} means that the abstract and introduction do not include the claims made in the paper.
        \item The abstract and/or introduction should clearly state the claims made, including the contributions made in the paper and important assumptions and limitations. A \answerNo{} or \answerNA{} answer to this question will not be perceived well by the reviewers. 
        \item The claims made should match theoretical and experimental results, and reflect how much the results can be expected to generalize to other settings. 
        \item It is fine to include aspirational goals as motivation as long as it is clear that these goals are not attained by the paper. 
    \end{itemize}

\item {\bf Limitations}
    \item[] Question: Does the paper discuss the limitations of the work performed by the authors?
    \item[] Answer: \answerYes{}
    \item[] Justification: We discussed limitations in Appendix.
    \item[] Guidelines:
    \begin{itemize}
        \item The answer \answerNA{} means that the paper has no limitation while the answer \answerNo{} means that the paper has limitations, but those are not discussed in the paper. 
        \item The authors are encouraged to create a separate ``Limitations'' section in their paper.
        \item The paper should point out any strong assumptions and how robust the results are to violations of these assumptions (e.g., independence assumptions, noiseless settings, model well-specification, asymptotic approximations only holding locally). The authors should reflect on how these assumptions might be violated in practice and what the implications would be.
        \item The authors should reflect on the scope of the claims made, e.g., if the approach was only tested on a few datasets or with a few runs. In general, empirical results often depend on implicit assumptions, which should be articulated.
        \item The authors should reflect on the factors that influence the performance of the approach. For example, a facial recognition algorithm may perform poorly when image resolution is low or images are taken in low lighting. Or a speech-to-text system might not be used reliably to provide closed captions for online lectures because it fails to handle technical jargon.
        \item The authors should discuss the computational efficiency of the proposed algorithms and how they scale with dataset size.
        \item If applicable, the authors should discuss possible limitations of their approach to address problems of privacy and fairness.
        \item While the authors might fear that complete honesty about limitations might be used by reviewers as grounds for rejection, a worse outcome might be that reviewers discover limitations that aren't acknowledged in the paper. The authors should use their best judgment and recognize that individual actions in favor of transparency play an important role in developing norms that preserve the integrity of the community. Reviewers will be specifically instructed to not penalize honesty concerning limitations.
    \end{itemize}

\item {\bf Theory assumptions and proofs}
    \item[] Question: For each theoretical result, does the paper provide the full set of assumptions and a complete (and correct) proof?
    \item[] Answer: \answerNA{} 
    \item[] Justification: 
    \item[] Guidelines:
    \begin{itemize}
        \item The answer \answerNA{} means that the paper does not include theoretical results. 
        \item All the theorems, formulas, and proofs in the paper should be numbered and cross-referenced.
        \item All assumptions should be clearly stated or referenced in the statement of any theorems.
        \item The proofs can either appear in the main paper or the supplemental material, but if they appear in the supplemental material, the authors are encouraged to provide a short proof sketch to provide intuition. 
        \item Inversely, any informal proof provided in the core of the paper should be complemented by formal proofs provided in appendix or supplemental material.
        \item Theorems and Lemmas that the proof relies upon should be properly referenced. 
    \end{itemize}

    \item {\bf Experimental result reproducibility}
    \item[] Question: Does the paper fully disclose all the information needed to reproduce the main experimental results of the paper to the extent that it affects the main claims and/or conclusions of the paper (regardless of whether the code and data are provided or not)?
    \item[] Answer: \answerYes{} 
    \item[] Justification: The paper provides full details on the architecture, training hyperparameters, and optimizer configuration.
    \item[] Guidelines:
    \begin{itemize}
        \item The answer \answerNA{} means that the paper does not include experiments.
        \item If the paper includes experiments, a \answerNo{} answer to this question will not be perceived well by the reviewers: Making the paper reproducible is important, regardless of whether the code and data are provided or not.
        \item If the contribution is a dataset and\slash or model, the authors should describe the steps taken to make their results reproducible or verifiable. 
        \item Depending on the contribution, reproducibility can be accomplished in various ways. For example, if the contribution is a novel architecture, describing the architecture fully might suffice, or if the contribution is a specific model and empirical evaluation, it may be necessary to either make it possible for others to replicate the model with the same dataset, or provide access to the model. In general. releasing code and data is often one good way to accomplish this, but reproducibility can also be provided via detailed instructions for how to replicate the results, access to a hosted model (e.g., in the case of a large language model), releasing of a model checkpoint, or other means that are appropriate to the research performed.
        \item While NeurIPS does not require releasing code, the conference does require all submissions to provide some reasonable avenue for reproducibility, which may depend on the nature of the contribution. For example
        \begin{enumerate}
            \item If the contribution is primarily a new algorithm, the paper should make it clear how to reproduce that algorithm.
            \item If the contribution is primarily a new model architecture, the paper should describe the architecture clearly and fully.
            \item If the contribution is a new model (e.g., a large language model), then there should either be a way to access this model for reproducing the results or a way to reproduce the model (e.g., with an open-source dataset or instructions for how to construct the dataset).
            \item We recognize that reproducibility may be tricky in some cases, in which case authors are welcome to describe the particular way they provide for reproducibility. In the case of closed-source models, it may be that access to the model is limited in some way (e.g., to registered users), but it should be possible for other researchers to have some path to reproducing or verifying the results.
        \end{enumerate}
    \end{itemize}

\item {\bf Open access to data and code}
    \item[] Question: Does the paper provide open access to the data and code, with sufficient instructions to faithfully reproduce the main experimental results, as described in supplemental material?
    \item[] Answer: \answerYes{} 
    \item[] Justification: Code, benchmark generators, and pre-trained checkpoints are released at github.
    \item[] Guidelines:
    \begin{itemize}
        \item The answer \answerNA{} means that paper does not include experiments requiring code.
        \item Please see the NeurIPS code and data submission guidelines (\url{https://neurips.cc/public/guides/CodeSubmissionPolicy}) for more details.
        \item While we encourage the release of code and data, we understand that this might not be possible, so \answerNo{} is an acceptable answer. Papers cannot be rejected simply for not including code, unless this is central to the contribution (e.g., for a new open-source benchmark).
        \item The instructions should contain the exact command and environment needed to run to reproduce the results. See the NeurIPS code and data submission guidelines (\url{https://neurips.cc/public/guides/CodeSubmissionPolicy}) for more details.
        \item The authors should provide instructions on data access and preparation, including how to access the raw data, preprocessed data, intermediate data, and generated data, etc.
        \item The authors should provide scripts to reproduce all experimental results for the new proposed method and baselines. If only a subset of experiments are reproducible, they should state which ones are omitted from the script and why.
        \item At submission time, to preserve anonymity, the authors should release anonymized versions (if applicable).
        \item Providing as much information as possible in supplemental material (appended to the paper) is recommended, but including URLs to data and code is permitted.
    \end{itemize}

\item {\bf Experimental setting/details}
    \item[] Question: Does the paper specify all the training and test details (e.g., data splits, hyperparameters, how they were chosen, type of optimizer) necessary to understand the results?
    \item[] Answer: \answerYes{} 
    \item[] Justification: Yes. We include details in experiment section.
    \item[] Guidelines:
    \begin{itemize}
        \item The answer \answerNA{} means that the paper does not include experiments.
        \item The experimental setting should be presented in the core of the paper to a level of detail that is necessary to appreciate the results and make sense of them.
        \item The full details can be provided either with the code, in appendix, or as supplemental material.
    \end{itemize}

\item {\bf Experiment statistical significance}
    \item[] Question: Does the paper report error bars suitably and correctly defined or other appropriate information about the statistical significance of the experiments?
    \item[] Answer: \answerYes{} 
    \item[] Justification: Appendix D reports mean ± std over four training seeds on six representative distributions.
    \item[] Guidelines:
    \begin{itemize}
        \item The answer \answerNA{} means that the paper does not include experiments.
        \item The authors should answer \answerYes{} if the results are accompanied by error bars, confidence intervals, or statistical significance tests, at least for the experiments that support the main claims of the paper.
        \item The factors of variability that the error bars are capturing should be clearly stated (for example, train/test split, initialization, random drawing of some parameter, or overall run with given experimental conditions).
        \item The method for calculating the error bars should be explained (closed form formula, call to a library function, bootstrap, etc.)
        \item The assumptions made should be given (e.g., Normally distributed errors).
        \item It should be clear whether the error bar is the standard deviation or the standard error of the mean.
        \item It is OK to report 1-sigma error bars, but one should state it. The authors should preferably report a 2-sigma error bar than state that they have a 96\% CI, if the hypothesis of Normality of errors is not verified.
        \item For asymmetric distributions, the authors should be careful not to show in tables or figures symmetric error bars that would yield results that are out of range (e.g., negative error rates).
        \item If error bars are reported in tables or plots, the authors should explain in the text how they were calculated and reference the corresponding figures or tables in the text.
    \end{itemize}

\item {\bf Experiments compute resources}
    \item[] Question: For each experiment, does the paper provide sufficient information on the computer resources (type of compute workers, memory, time of execution) needed to reproduce the experiments?
    \item[] Answer: \answerYes{} 
    \item[] Justification: We report this in experiment section and Appendix.
    \item[] Guidelines:
    \begin{itemize}
        \item The answer \answerNA{} means that the paper does not include experiments.
        \item The paper should indicate the type of compute workers CPU or GPU, internal cluster, or cloud provider, including relevant memory and storage.
        \item The paper should provide the amount of compute required for each of the individual experimental runs as well as estimate the total compute. 
        \item The paper should disclose whether the full research project required more compute than the experiments reported in the paper (e.g., preliminary or failed experiments that didn't make it into the paper). 
    \end{itemize}
    
\item {\bf Code of ethics}
    \item[] Question: Does the research conducted in the paper conform, in every respect, with the NeurIPS Code of Ethics \url{https://neurips.cc/public/EthicsGuidelines}?
    \item[] Answer: \answerYes{} 
    \item[] Justification: The research conforms with the NeurIPS Code of Ethics in all respects. The work is purely methodological, uses only synthetic routing instances.
    \item[] Guidelines:
    \begin{itemize}
        \item The answer \answerNA{} means that the authors have not reviewed the NeurIPS Code of Ethics.
        \item If the authors answer \answerNo, they should explain the special circumstances that require a deviation from the Code of Ethics.
        \item The authors should make sure to preserve anonymity (e.g., if there is a special consideration due to laws or regulations in their jurisdiction).
    \end{itemize}

\item {\bf Broader impacts}
    \item[] Question: Does the paper discuss both potential positive societal impacts and negative societal impacts of the work performed?
    \item[] Answer: \answerYes{} 
    \item[] Justification: We discussed this in Appendix.
    \item[] Guidelines:
    \begin{itemize}
        \item The answer \answerNA{} means that there is no societal impact of the work performed.
        \item If the authors answer \answerNA{} or \answerNo, they should explain why their work has no societal impact or why the paper does not address societal impact.
        \item Examples of negative societal impacts include potential malicious or unintended uses (e.g., disinformation, generating fake profiles, surveillance), fairness considerations (e.g., deployment of technologies that could make decisions that unfairly impact specific groups), privacy considerations, and security considerations.
        \item The conference expects that many papers will be foundational research and not tied to particular applications, let alone deployments. However, if there is a direct path to any negative applications, the authors should point it out. For example, it is legitimate to point out that an improvement in the quality of generative models could be used to generate Deepfakes for disinformation. On the other hand, it is not needed to point out that a generic algorithm for optimizing neural networks could enable people to train models that generate Deepfakes faster.
        \item The authors should consider possible harms that could arise when the technology is being used as intended and functioning correctly, harms that could arise when the technology is being used as intended but gives incorrect results, and harms following from (intentional or unintentional) misuse of the technology.
        \item If there are negative societal impacts, the authors could also discuss possible mitigation strategies (e.g., gated release of models, providing defenses in addition to attacks, mechanisms for monitoring misuse, mechanisms to monitor how a system learns from feedback over time, improving the efficiency and accessibility of ML).
    \end{itemize}
    
\item {\bf Safeguards}
    \item[] Question: Does the paper describe safeguards that have been put in place for responsible release of data or models that have a high risk for misuse (e.g., pre-trained language models, image generators, or scraped datasets)?
    \item[] Answer:  \answerNA{} 
    \item[] Justification: The paper releases neural routing solvers and synthetic VRP benchmark instances, neither of which poses a high risk of misuse.
    \item[] Guidelines:
    \begin{itemize}
        \item The answer \answerNA{} means that the paper poses no such risks.
        \item Released models that have a high risk for misuse or dual-use should be released with necessary safeguards to allow for controlled use of the model, for example by requiring that users adhere to usage guidelines or restrictions to access the model or implementing safety filters. 
        \item Datasets that have been scraped from the Internet could pose safety risks. The authors should describe how they avoided releasing unsafe images.
        \item We recognize that providing effective safeguards is challenging, and many papers do not require this, but we encourage authors to take this into account and make a best faith effort.
    \end{itemize}

\item {\bf Licenses for existing assets}
    \item[] Question: Are the creators or original owners of assets (e.g., code, data, models), used in the paper, properly credited and are the license and terms of use explicitly mentioned and properly respected?
    \item[] Answer: \answerYes{} 
    \item[] Justification: We include this in Appendix.
    \item[] Guidelines:
    \begin{itemize}
        \item The answer \answerNA{} means that the paper does not use existing assets.
        \item The authors should cite the original paper that produced the code package or dataset.
        \item The authors should state which version of the asset is used and, if possible, include a URL.
        \item The name of the license (e.g., CC-BY 4.0) should be included for each asset.
        \item For scraped data from a particular source (e.g., website), the copyright and terms of service of that source should be provided.
        \item If assets are released, the license, copyright information, and terms of use in the package should be provided. For popular datasets, \url{paperswithcode.com/datasets} has curated licenses for some datasets. Their licensing guide can help determine the license of a dataset.
        \item For existing datasets that are re-packaged, both the original license and the license of the derived asset (if it has changed) should be provided.
        \item If this information is not available online, the authors are encouraged to reach out to the asset's creators.
    \end{itemize}

\item {\bf New assets}
    \item[] Question: Are new assets introduced in the paper well documented and is the documentation provided alongside the assets?
    \item[] Answer: \answerYes{} 
    \item[] Justification: The new assets introduced in this paper are documented in the main text and Appendix.
    \item[] Guidelines:
    \begin{itemize}
        \item The answer \answerNA{} means that the paper does not release new assets.
        \item Researchers should communicate the details of the dataset\slash code\slash model as part of their submissions via structured templates. This includes details about training, license, limitations, etc. 
        \item The paper should discuss whether and how consent was obtained from people whose asset is used.
        \item At submission time, remember to anonymize your assets (if applicable). You can either create an anonymized URL or include an anonymized zip file.
    \end{itemize}

\item {\bf Crowdsourcing and research with human subjects}
    \item[] Question: For crowdsourcing experiments and research with human subjects, does the paper include the full text of instructions given to participants and screenshots, if applicable, as well as details about compensation (if any)? 
    \item[] Answer: \answerNA{} 
    \item[] Justification: The paper does not involve crowdsourcing or research with human subjects.
    \item[] Guidelines:
    \begin{itemize}
        \item The answer \answerNA{} means that the paper does not involve crowdsourcing nor research with human subjects.
        \item Including this information in the supplemental material is fine, but if the main contribution of the paper involves human subjects, then as much detail as possible should be included in the main paper. 
        \item According to the NeurIPS Code of Ethics, workers involved in data collection, curation, or other labor should be paid at least the minimum wage in the country of the data collector. 
    \end{itemize}

\item {\bf Institutional review board (IRB) approvals or equivalent for research with human subjects}
    \item[] Question: Does the paper describe potential risks incurred by study participants, whether such risks were disclosed to the subjects, and whether Institutional Review Board (IRB) approvals (or an equivalent approval/review based on the requirements of your country or institution) were obtained?
    \item[] Answer: \answerNA{} 
    \item[] Justification: The paper does not involve research with human subjects.
    \item[] Guidelines:
    \begin{itemize}
        \item The answer \answerNA{} means that the paper does not involve crowdsourcing nor research with human subjects.
        \item Depending on the country in which research is conducted, IRB approval (or equivalent) may be required for any human subjects research. If you obtained IRB approval, you should clearly state this in the paper. 
        \item We recognize that the procedures for this may vary significantly between institutions and locations, and we expect authors to adhere to the NeurIPS Code of Ethics and the guidelines for their institution. 
        \item For initial submissions, do not include any information that would break anonymity (if applicable), such as the institution conducting the review.
    \end{itemize}

\item {\bf Declaration of LLM usage}
    \item[] Question: Does the paper describe the usage of LLMs if it is an important, original, or non-standard component of the core methods in this research? Note that if the LLM is used only for writing, editing, or formatting purposes and does \emph{not} impact the core methodology, scientific rigor, or originality of the research, declaration is not required.
    \item[] Answer: \answerYes{} 
    \item[] Justification: We include this in Appendix.
    \item[] Guidelines:
    \begin{itemize}
        \item The answer \answerNA{} means that the core method development in this research does not involve LLMs as any important, original, or non-standard components.
        \item Please refer to our LLM policy in the NeurIPS handbook for what should or should not be described.
    \end{itemize}

\end{enumerate}

\end{document}